\documentclass[10pt,twocolumn,letterpaper]{article}

\usepackage[pagenumbers]{cvpr} 

\usepackage{graphicx}
\usepackage{amsmath}
\usepackage{amssymb}
\usepackage{booktabs}
\usepackage{sidecap}
\usepackage{enumitem}
\setlist[itemize]{parsep=0pt,topsep=2pt,itemsep=2pt}
\setlist[enumerate]{parsep=0pt,topsep=1pt,itemsep=1pt}
\usepackage{adjustbox}
\usepackage[dvipsnames]{xcolor}
\usepackage{adjustbox}

\definecolor{cvprblue}{rgb}{0.21,0.49,0.74}
\usepackage[pagebackref,breaklinks,colorlinks,citecolor=cvprblue]{hyperref}

\usepackage[capitalize]{cleveref}
\usepackage{standalone}
\newcommand{\secondarytitle}[1]{%
\twocolumn[{
    \begin{center}
        \Large\textbf{#1}\par
        \end{center}
    \vspace{10em} 
}]
}
\crefname{section}{Sec.}{Secs.}
\Crefname{section}{Section}{Sections}
\Crefname{table}{Table}{Tables}
\crefname{table}{Tab.}{Tabs.}

\makeatletter
\renewcommand\paragraph{\@startsection{paragraph}{4}{\z@}%
                                    {1.25ex \@plus1ex \@minus.2ex}%
                                    {-1em}%
                                    {\normalfont\normalsize\bfseries}}
\makeatother

\usepackage{graphicx}
\usepackage{amsmath}
\usepackage{amssymb}

\usepackage[utf8]{inputenc} 
\usepackage[T1]{fontenc}    
\usepackage{hyperref}       
\usepackage{url}            
\usepackage{booktabs}       
\usepackage{amsfonts}       
\usepackage{nicefrac}       
\usepackage{microtype}      
\usepackage[dvipsnames]{xcolor}         

\usepackage{amssymb}
\usepackage{pifont}
\usepackage{fixmetodonotes}
\usepackage{colortbl}
\usepackage{rotating}
\usepackage{scalerel}

\usepackage{multirow}
\usepackage[normalem]{ulem}
\usepackage{cancel}

\usepackage{bm}
\usepackage{adjustbox}
\usepackage{array}
\newcolumntype{R}[2]{%
    >{\adjustbox{angle=#1,lap=1.3\width-(#2)}\bgroup}%
    l%
    <{\egroup}%
}

\usepackage[linesnumbered,ruled,vlined]{algorithm2e}
\usepackage{algpseudocode}
\makeatletter 
\@namedef{ver@everyshi.sty}{}
\makeatother
\usepackage{tikz}
\usetikzlibrary{positioning,shapes}

\definecolor{MyGreen}{RGB}{0, 180, 0}
\definecolor{MyRed}{RGB}{180, 0, 0}
\definecolor{MyBlue}{RGB}{30, 0, 180}
\definecolor{MyGrey}{RGB}{82.75, 82.75, 82.75}

\definecolor{skcar}{RGB}{100,150,245}
\definecolor{skbicycle}{RGB}{255, 200, 0}
\definecolor{skmotorcycle}{RGB}{255, 120, 0}
\definecolor{sktruck}{RGB}{80, 30, 180}
\definecolor{skotherv}{RGB}{0, 0, 255}
\definecolor{skpedestrian}{RGB}{255,30,30}
\definecolor{skdrivable}{RGB}{255, 0,255}
\definecolor{sksidewalk}{RGB}{75, 0,75}
\definecolor{skterrain}{RGB}{150, 240, 80}
\definecolor{skvegetation}{RGB}{0, 175, 0}
\definecolor{skbuilding}{RGB}{255, 200, 0}

\definecolor{tsne_source}{rgb}{0.86, 0.3712, 0.33999999999999997}
\definecolor{tsne_target}{rgb}{0.33999999999999997, 0.8287999999999999, 0.86}

\makeatletter
\renewcommand\paragraph{\@startsection{paragraph}{4}{\z@}%
    {0.7ex \@plus0.5ex \@minus.2ex}%
    {-1em}%
    {\normalfont\normalsize\bfseries}}
\makeatother

\newcommand{\cmark}{{\textcolor{MyGreen}{\ding{51}}}}%
\def \method{SALUDA\xspace}

\def \lidar{lidar\xspace} 

\newcommand{\synth}{{SL}}
\newcommand{\sposssyn}{{SP$_{13}$}}
\newcommand{\spossns}{{SP$_{6}$}}
\newcommand{\sk}{{SK}} 
\newcommand{\ns}{{NS}} 
\newcommand{\skns}{{SK$_{10}$}} 
\newcommand{\sksyn}{{SK$_{19}$}}
\newcommand{\minent}{{MinEnt}}
\newcommand{\cosmix}{{CoSMix}}

\newcommand{\batchnorm}{{batch norm}}
\newcommand{\nstosk}{{\DAsetting{\ns}{\skns}}}
\newcommand{\synthtosk}{{\DAsetting{\synth}{\sksyn}}}
\newcommand{\synthtoposs}{{\DAsetting{\synth}{\sposssyn}}}
\newcommand{\nstoposs}{{\DAsetting{\ns}{\spossns}}}

\newcommand{\tsne}{{t-SNE}}

 \newcommand{\titletext}{\method{}: Surface-based Automotive Lidar Unsupervised Domain Adaptation} 

\newcommand{\perf}[1]{{{#1}}}
\newcommand{\std}[1]{\scaleto{\;\pm \text{#1}}{5pt}}
\newcommand{\second}[1]{\cellcolor{blue!10}{#1}}
\newcommand{\best}[1]{\cellcolor{blue!20}{#1}}
\newcommand{\back}[1]{\it\textcolor{black!30}{#1}}

\newcommand{\mioucell}{\cellcolor{green!0}}

\newcommand{\ssp}[1]{\,{#1}\,}
\newcommand{\sem}{{\mathsf{sem}}}
\newcommand{\occ}{{\mathsf{occ}}}
\newcommand{\src}{{\mathsf{s}}}
\newcommand{\tgt}{{\mathsf{t}}}
\newcommand{\ST}{{\mathsf{ST}}}
\newcommand{\PL}{{\mathsf{PL}}}

\usepackage{stmaryrd}
\usepackage{trimclip}

\makeatletter
\DeclareRobustCommand{\shortto}{%
  \mathrel{\mathpalette\short@to\relax}%
}

\newcommand{\short@to}[2]{%
  \mkern2mu
  \clipbox{{.5\width} 0 0 0}{$\m@th#1\vphantom{+}{\shortrightarrow}$}%
  }
\makeatother

\newcommand{\DAsetting}[2]{{#1}$\rightarrow${#2}}

\def\confName{3DV\xspace}
\def\confYear{2024\xspace}

\title{\titletext}

\author{
Bj\"orn Michele$^{1,3}$
\and
Alexandre Boulch$^{1}$
\and
Gilles Puy$^{1}$
\and
Tuan-Hung Vu$^{1}$
\and
Renaud Marlet$^{1,2}$
\and
Nicolas Courty$^{3}$
\and
\large
\hspace{-3mm}\textsuperscript{1}Valeo.ai, Paris, France  \hspace{1mm} \textsuperscript{2}LIGM, Ecole des Ponts, Univ Gustave Eiffel, CNRS, Marne-la-Vall\'ee, France \\
\textsuperscript{3}CNRS, IRISA, Univ. Bretagne Sud, Vannes, France
}

\begin{document}
\maketitle
\begin{abstract}

Learning models on one labeled dataset that generalize well on another domain is a difficult task, as several shifts might happen between the data domains. This is notably the case for lidar data, for which models can exhibit large performance discrepancies due for instance to different lidar patterns or changes in acquisition conditions. This paper addresses the corresponding Unsupervised Domain Adaptation (UDA) task for semantic segmentation. To mitigate this problem, we introduce an unsupervised auxiliary task of learning an implicit underlying surface representation simultaneously on source and target data. As both domains share the same latent representation, the model is forced to accommodate discrepancies between the two sources of data. This novel strategy differs from classical minimization of statistical divergences or lidar-specific domain adaptation techniques. Our experiments demonstrate that our method achieves a better performance than the current state of the art, both in real-to-real and synthetic-to-real scenarios.

The project repository: \href{https://github.com/valeoai/SALUDA}{github.com/valeoai/SALUDA}
\end{abstract}

\begin{figure}
\centering
\includegraphics[width=\linewidth]{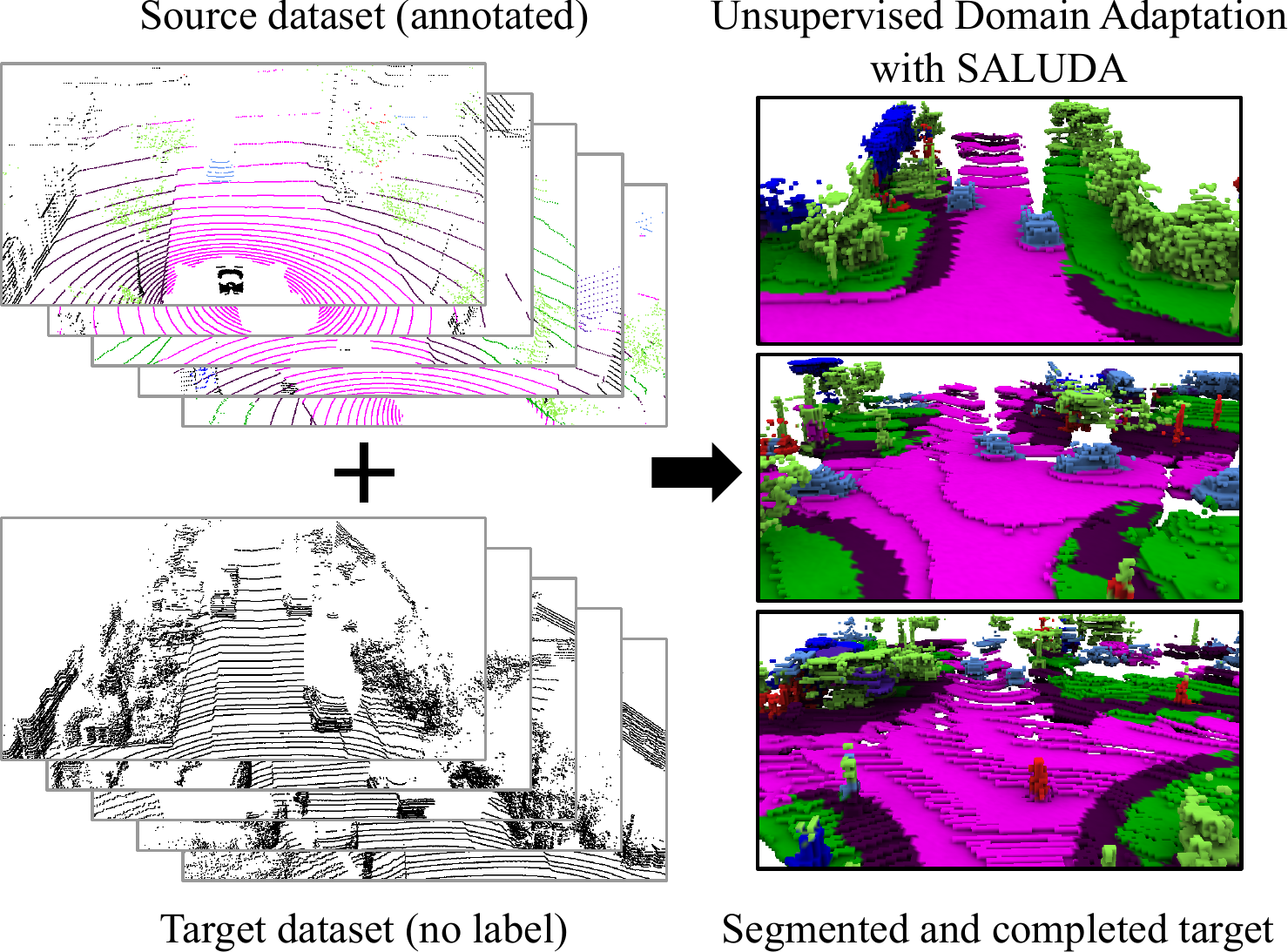}
\vspace*{-6mm}
\caption{\textbf{Unsupervised domain adaptation with \method{}.} It leverages annotated source data, e.g., nuScenes dataset and unlabeled target data, e.g., SemanticKITTI, for semantic segmentation of the target.
The surface, which is a by-product of the approach, is colored according to the semantic predictions.}
\label{fig:teaser}
\vspace*{-3mm}
\end{figure}

\section{Introduction}

Unsupervised domain adaptation (UDA) is an annotation-efficient training technique which permits boosting the performance of a network on a target domain by leveraging labeled data from a source domain and unlabeled data from the target domain~\cite{daume2006domain}.
UDA methods are designed to mitigate the distribution shifts between source and target data, to alleviate the drop in accuracy on target data at test time.

In computer vision, most UDA methods have been designed for image perception tasks~\cite{ganin2016domain, sun2016deep, wang2017deep, long2018conditional, hoffman2018cycada, vu2019advent}, with some recent attempts to adapt these methods to 3D scenarios. Some examples of successful transfer from 2D to 3D involve self-training~\cite{yang2021st3d,yang2021st3d++}, adversarial training~\cite{barrera2021cycle,debortoli2021adversarial,jiang2021lidarnet}, and mixing techniques~\cite{kong2023conda,saltori2022cosmix}.
Nevertheless, the inherent difference between (dense) 2D images and (sparse) 3D point clouds calls for methods specifically designed for point clouds. This is all the more necessary as the 3D domain gaps can be particularly wide due to the type of scenes (indoors vs outdoors, static vs dynamic), to the variety of sensors (depth camera with structured light or time-of-flight, sweeping lidar or
non-repetitive scanning patterns,
photogrammetry, etc.), to the form of scan fusion, if any, and to the sensor characteristics (e.g., number of laser beams, angular resolutions, range, intensity calibration, etc.).

Point-cloud-specific domain adaptation techniques can learn domain invariant feature representations, e.g., by relying on self-supervised tasks such as partial deformation and reconstruction of point clouds \cite{achituve2021self}, or reconstruction of point clouds from 2D projections \cite{fan2022self}.
In \cite{yi2021complete}, densifying the lidar point-clouds and representing them as canonical patterns help mitigate the sensor gaps.

In this paper, we address the important problem of domain adaptation between different automotive lidar sensors (including different sensor locations on the vehicles) for the task of semantic segmentation. We focus on large outdoor scenes captured by relatively coarse lidars as they appear in automotive scenarios. Our work takes inspiration from Complete \& Label~\cite{yi2021complete}, which shows that the underlying surface captured by lidar sensors is a good medium to reduce the domain gap between different sensor settings.

Concretely, we train a single semantic segmentation backbone and two heads (cf.\ Fig.\,\ref{fig:architecture}): one for semantic segmentation and one for implicit surface reconstruction (self-supervised occupancy estimation). The training alternates between source and target data. With (annotated) source input, we train both heads, joining the two losses; with (unannotated) target input, we only train the surface reconstruction head. The obtained model is then used to initialize a student-teacher model that performs self-training. 
At test time, we only use the segmentation head. That is, we train the backbone to know about the input patterns of both domains and to produce point features suitable for both semantic segmentation (on source) and surface reconstruction (on source and target). By favoring the alignment of features when source and target sample a similar surface, although differently, and as we can learn (on source) to map features to semantics, we can then map target data to source-like semantics. Experiments show our method for ``Surface-based Automotive Lidar Unsupervised Domain Adaptation'' (\method) is outperforming the current state of the art (SOTA).

Last but not least, we show that \method's hyperparameters can be selected in a strict UDA protocol, where strictly no target labels can be used. The model selection protocols that we use follow from~\cite{musgrave2022benchmarking}, which indeed questions the common practice in UDA to use a labeled validation set of target data for hyperparameter tuning and model selection.
Such a practice is arguably acceptable when one can afford to annotate a small dataset from the target domain of interest.
However, it is closer to semi-supervision and it does not show  the actual performance that a method can reach in a truly unsupervised setting where no target label is available at all. Therefore, we adopt some of the fully unsupervised validators proposed in~\cite{musgrave2022benchmarking} and compare their behaviors in our setting. As such, we aim at providing UDA practitioners with a complete package with both a new methodology and a model selection guidance.

To summarize, our contributions are the following:
\begin{itemize}
    \item We propose a novel approach based on implicit surface reconstruction to train a semantic segmenter for point clouds that generalizes well across domains. 
    \item Experiments (real-to-real and synthetic-to-real) show that our method outperforms the SOTA.
    \item Further studies (i)~reveal the particular importance of batch norm statistics in 3D, and (ii) illustrate the robustness of {\method} in the strict UDA setting, i.e., with no hyperparameter tuning on labeled target data.
\end{itemize}
\section{Related work}

\subsection{Visual Unsupervised Domain Adaptation}

Unsupervised domain adaptation is a type of transductive transfer learning. It aims at mitigating the domain shift problem between a source domain (for training) and a target domain (for testing). For the source domain, labeled examples are available whereas, for the target domain, only unannotated data are available \cite{wilson2020survey}. The UDA methods used in computer vision can be roughly distinguished into four categories: learning of domain-invariant feature representations, domain mapping, self-supervision, and \batchnorm{} statistics adaptation.
Domain-invariant features can be obtained by explicitly minimizing statistical divergences between source and target feature representations (e.g., \cite{ long2015learning,long2017deep, sun2016deep, wang2017deep, damodaran2018deepjdot, fatras2021jumbot}), or through adversarial training (e.g., \cite{ganin2016domain,tzeng2017adversarial,  long2018conditional}).
Domain mapping approaches try to learn a translation between the source and the target domain \cite{hoffman2018cycada, choi2018stargan}. Self-supervised methods \cite{vu2019advent}, pseudo-labeling \cite{saito2017asymmetric,zou2018unsupervised, zou2019confidence}, or self-ensembling \cite{laine2016temporal, tarvainen2017mean, tranheden2021dacs, hoyer2022daformer} are also popular methods.

Several studies also explored the use of batch norm statistics~\cite{ioffe2015batch} for domain adaptation~\cite{LI2018109, nado2020evaluating,wangtent,mirza2022dua,schneider2020improving}.
Specifically, in \cite{LI2018109}, the running batch norm statistics of the source domain are replaced with the one of the target domain.
\cite{mirza2022dua,schneider2020improving, nado2020evaluating}, focusing on ad-hoc or online adaptation using source statistics as prior so that fewer target examples are needed to adapt \cite{mirza2022dua,schneider2020improving} or calculating the running statistics for each test batch individually \cite{nado2020evaluating}.

\subsection{UDA for 3D data}\label{fig:uda43ddata}

Categories of UDA methods for 3D data can be drawn from the targeted learning task (e.g., detection or segmentation), or from the nature of the 3D information at hand (single object or large outdoor scenes, such as in autonomous driving).

\smallskip\noindent\textbf{UDA for 3D objects.}
Early work targeting single object classification aligns local and global features to mitigate the domain gap \cite{qin2019pointdan}.
More recent work do a joint training of both a classification task and a self-supervised task specifically designed for 3D objects \cite{zhou2018unsupervised, achituve2021self, luo2021learnable, zou2021geometry, shen2022domain, fan2022self, liangpoint}, as it promotes the learning of domain-invariant features. The self-supervised tasks used include the prediction of a transformation applied to the input \cite{zou2021geometry, zhou2018unsupervised, fan2022self} or its inversion, e.g., reconstructing from a 2D projection of the 3D object \cite{fan2022self} or from a deformed input \cite{achituve2021self}, or even learning the deformation \cite{luo2021learnable}.
The self-supervised task can also be to implicitly learn the shape \cite{shen2022domain}  or to learn geometrical properties such as point normals, positions, or densities \cite{liangpoint}.

\smallskip\noindent\textbf{UDA for \lidar\ scenes.} A first class of methods directly  inherits from 2D methods, e.g., using an adversarial approach with a point cloud represented as a projected bird-eye-view image~\cite{zhao2021epointda, barrera2021cycle, debortoli2021adversarial, jiang2021lidarnet} or with a spherical projection~\cite{Li_2023_CVPR}.
A second group of methods is designed specifically to handle point cloud representations for detection \cite{yang2021st3d,yang2021st3d++,Luo_2021_ICCV, saltori2020sf, you2022exploiting, wang2020train, zhang2021srdan, xu2021spg, peng2023cl3d}.
For example, the difference in vehicle size is a common gap when source and target originate from different countries. Therefore, adapting the size of the cars between domains has proved very efficient \cite{wang2020train,yang2021st3d,yang2021st3d++}.
Adaptation can also leverage tempo-spatial sequential relations of automotive \lidar\ scans \cite{saltori2020sf,you2022exploiting}, aligning objects of similar sizes and distances \cite{zhang2021srdan}, or hallucinating points in the object bounding boxes to facilitate detection \cite{xu2021spg}.
Following recent developments of mixing approaches in domain adaptation, \cite{saltori2022cosmix, kong2023conda} mix source and target \lidar\ scenes. This is done in the range image representation \cite{kong2023conda} or in a point cloud \cite{saltori2022cosmix}. Recently, domain generalization for \lidar scenes raised some attention \cite{Xiao_2023_CVPR,Kim_2023_CVPR,Ryu_2023_CVPR,saltori2023walking, sanchez2023domain}. In domain generalization, in contrast to the UDA setting, there is no access to target data at all (until inference).
Finally, multi-modality can also be used, e.g., \lidar and images, to cross the domain gap \cite{peng2021sparse, LIU2021211, jaritz2020xmuda, jaritz2022cross, tsai2022see, fei2022adas}.

\smallskip\noindent\textbf{Cross-sensor domain adaptation.}
A \lidar-specific domain gap is induced by different \lidar\ patterns originating from different \lidar\ sensors \cite{rist2019cross, yi2021complete, rochan2022unsupervised, alonso2020domain, langer2020domain}. In the automotive \lidar, this domain gap is especially studied for Velodyne \lidar\ sensors with a different number of beams, angle of beams, and number of measurements per beam \cite{yi2021complete,wei2022lidar}. 

One way is to adapt the source input points to have a more similar input than the target data. This can be done by resampling and removing beams to make the source and target \lidar\ pattern more similar \cite{wei2022lidar}, a successful approach for a high-beam sensor to a low-beam sensor adaptation, but not applicable the other way around.

Another direction is to use a completion task as regularization for gap bridging.
In \cite{rochan2022unsupervised}, a self-supervised image completion on range images is used as an auxiliary task. 

Closer to us, \cite{yi2021complete} learns to complete input voxels to represent the underlying surface, which acts as a ``canonical domain'' and is used as pivot for domain adaptation. A separate reference dataset with rich annotations is first used to learn dense voxel completion from sparse voxels. Then, the specific sensors of the source and target domains are simulated on the completed reference dataset to learn completion on source and target data as well. Last, semantic segmentation is learned on (interpolated) dense completed source data, inferred on dense completed target data, and projected on sparse uncompleted input target data. While the approach is general and effective, enough frames have to be aggregated for surface reconstruction to succeed, relying also on the identification of moving objects based on tracking annotations in the dataset. Rather than rely on explicit surface reconstruction (voxels and meshes), our approach stays at an implicit level (occupancy estimation); we do not learn to complete source and target data, and only implicitly adapt the segmentation network to accommodate different types of \lidar patterns.
Besides, frame aggregation and tracking information are not needed. Last, at inference time, \cite{yi2021complete} pipelines three networks and a reprojection, with a high computation burden, whereas we only use a standard semantic segmenter.

\subsection{Surface reconstruction with implicit functions}
Implicit neural representations for surface reconstruction of 3D objects from point clouds have shown impressive results  \cite{park2019deepsdf,mescheder2019occupancy,chen2019learning}.  The implicit function models the (signed or unsigned) distance of a query point to the surface, or its volumetric occupancy (empty or full, meaning inside or outside the object). A neural network learns to approximate this function during training. To represent a large or complex scene with different objects, spatially-distributed latent representations are used, where latent vectors encode the surface information of their spatial neighborhood \cite{convonet, rist2020scssnet, rist2021semantic,Boulch_2022_CVPR}. These latent representations can be equally spaced in a 3D grid \cite{convonet, rist2020scssnet, rist2021semantic} or attached to input points \cite{Boulch_2022_CVPR, ALSO}.

Learning implicit functions for surface reconstruction has been proved to be an effective self-supervised pretext task for point-cloud representation learning.
ALSO~\cite{ALSO} successfully uses implicit neural representations for downstream 3D tasks like semantic segmentation. However, it only focuses on in-domain data, i.e., pre-training and downstream fine-tuning on the same dataset. It leaves open the question of how well such methods can learn features that generalize across multiple domains undergoing significant discrepancies.

\section{Method}

\begin{figure*}
\centering
\includegraphics[width=\linewidth]{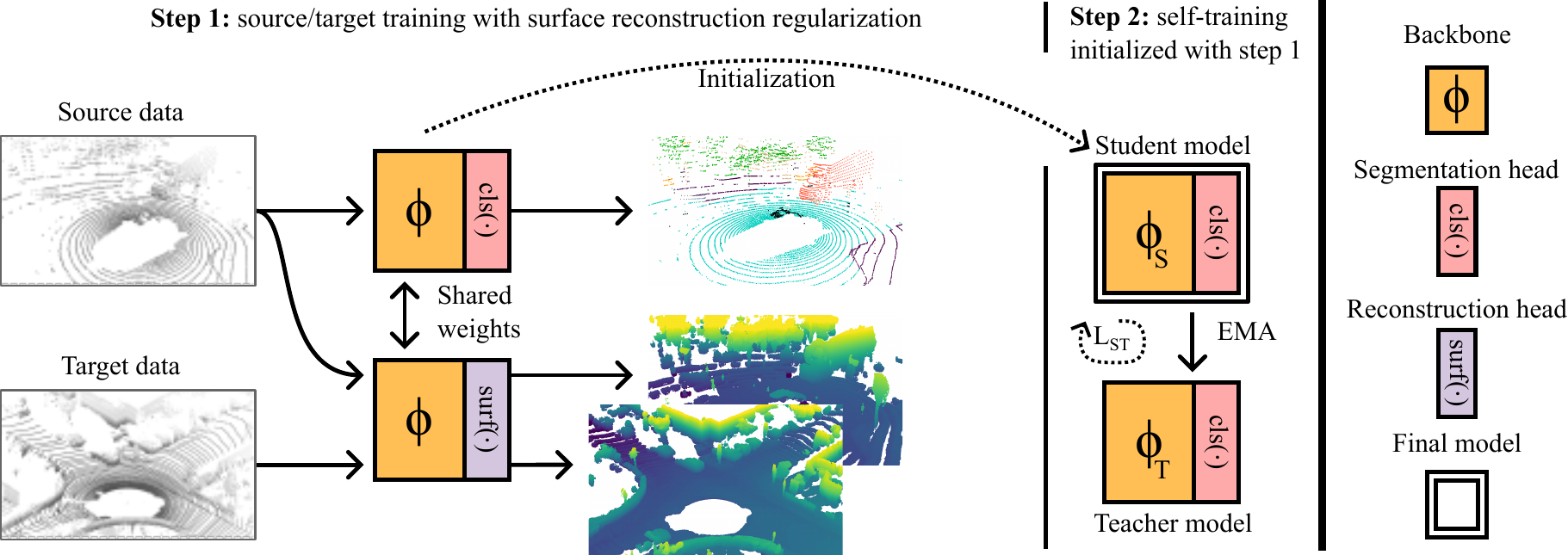}
\caption{\textbf{Overview of {\method} (training stage).} Step 1, the backbone $\phi(\cdot)$ is trained  alternating between source and target point clouds. With (annotated) source data, it produces point-wise latent vectors that are used both by the segmentation head ${\rm cls}(\cdot)$ to classify each point and yield semantic segments, and by the surface reconstruction head ${\rm surf}(\cdot)$ to estimate occupancy. With (unannotated) target data, the latent vectors are only fed to the surface reconstruction head. Conversely, at test time, only the semantic segmentation head is used. Step 2, the obtained weights are used as an initialization for teacher/student self-training. It is done with true labels for source data and pseudo-labels for target data. The teacher is an exponential moving average (EMA) of the student. The self-training loss $\mathcal L_\ST$ is defined in Section~\ref{sec:traininglosses}.}
\label{fig:architecture}
\vspace*{-1mm}
\end{figure*}

We consider a dataset in source (resp.\ target) domain with labeled (resp.\ unlabeled) point clouds $P^\src$ (resp.\ $P^\tgt$). We want to train a high-performing semantic segmenter. This paper explores a novel approach based on surface reconstruction. We assume that if we can create good point features for a downstream task from a labeled source dataset, and if we can create, from an unlabeled target dataset, similar features for points located at similar locations on similar objects (thus semantically similar), then we can directly transfer the learned downstream task from source to target. Effectively, to align source and target features, we require that they reconstruct a similar underlying surface. It is the main intuition behind the SALUDA pipeline, which is overviewed in Fig.\,\ref{fig:architecture}.

\subsection{Architecture}

A common trait between point clouds acquired by different \lidar{s} is the underlying geometry of sampled scenes. Based on this observation, we propose to use implicit surface representation as an auxiliary task to learn semantic features that generalize well across domains. 

The network architecture of \method is displayed on Fig.\,\ref{fig:architecture}. It consists of a single backbone followed by two heads: one for semantic segmentation, and one for implicit surface reconstruction. 
The two tasks are jointly trained, encouraging the backbone to create features that are good both for surface reconstruction and semantic segmentation. At inference time, the surface representation head is discarded; the network reduces to a standard semantic segmenter.

Given a point cloud $P$ (source or target), the backbone $\phi(\cdot)$ infers for each point $p\ssp\in P$ a $d$-dimensional latent vector $z_p \ssp\in \mathbb{R}^d$, that is fed into both the semantic and surface heads.

The semantic segmentation head ${\rm cls}(\cdot)$ is one linear layer with softmax activation, producing classwise probabilities. 
 
For the surface reconstruction head ${\rm surf}(\cdot)$, we follow the architecture design in~\cite{ALSO}.
Given an arbitrary query point $q$ in space where to estimate occupancy (full or empty), we consider a ball $B_q$ centered at $q$. (In our experiments, we use a 1-meter radius ball.) For each input point $p$ falling into $B_q$, we concatenate the latent vector $z_p$ with the position of $p$ relative to~$q$, i.e., $p \ssp- q$. The resulting matrix $\widetilde{Z}_q $ (one row per point) is processed by a row-wise MLP, followed by a weighted average pooling layer (with learned weights) and a final sigmoid, yielding a scalar $\tilde o_q$ 
representing the probability that query point $q$ falls in an object.

Though borrowing the simple, yet effective design of~\cite{ALSO} for the auxiliary surface branch, we emphasize that our task is different and more challenging. 
While in~\cite{ALSO} the surface reconstruction task is used for pre-training single-domain features, we use this task to align cross-domain features.
Furthermore, in~\cite{ALSO}, the training of downstream tasks is done separately from the pre-training phase. In contrast, we propose a novel framework where we jointly train segmentation and reconstruction for source data while achieving source-target alignment through reconstruction.
As shown later in Sec.~\ref{sec:comparison}, pre-training with surface reconstruction, even on both domains, does not guarantee well-generalized features.
Our proposed framework is particularly designed to mitigate domain gaps, resulting in the best results on target data.

\subsection{Learning cross-domain surface reconstruction}
\label{sec:pseudolabelssurface}

\begin{SCfigure}
\centering
\raisebox{3.5mm}{\includegraphics[width=0.48\linewidth]{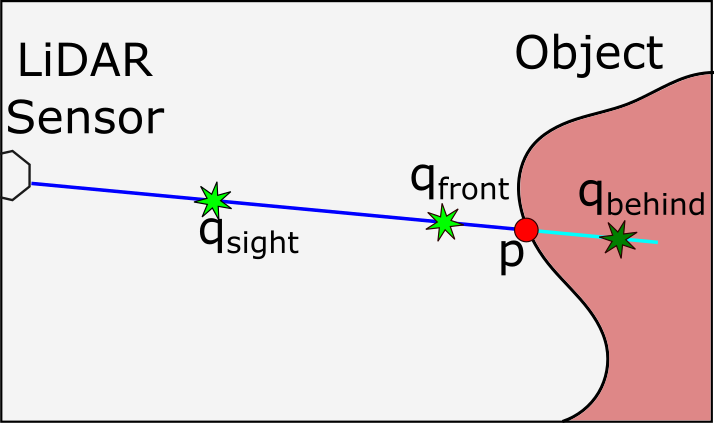}}
\caption{\textbf{Visibility query point sampling:} $q_{\mathsf{sight}}$ and $q_{\mathsf{front}}$ are placed on the line of sight between sensor and observed point $p$ and pseudo-labeled as empty; $q_{\mathsf{behind}}$ is placed just ``after'' $p$ and pseudo-labeled as full. }
\label{fig:aux_sampling}
\vspace*{-2mm}
\end{SCfigure}

To learn the auxiliary surface reconstruction branch, we follow~\cite{sulzer2022deep,ALSO} and formulate a self-supervised training objective. We exploit the line of sight between the sensor and each observed point: the space on this line segment is assumed empty, while it is full immediately behind the observed point. We use it to create visibility-based query points and assign them pseudo-labels full or empty, whether they are likely to fall inside or outside an object (see Fig.\,\ref{fig:aux_sampling}).

Surface reconstruction learning is done on both source and target domains.
For each input point $p$, in either $P^\src$ or $P^\tgt$, we create three query points along the line of sight passing through the center~$c$ of the \lidar sensor and the observed point~$p$. We place the query points $q_{\mathsf{front}}$ and $q_{\mathsf{behind}}$ in front of and behind $p$ with respect to~$c$, at a distance uniformly drawn in $[0,\delta]$ for some $\delta\ssp>0$. We also uniformly draw a third query point $q_{\mathsf{sight}}$ in the interval between $c$ and $q_{\mathsf{front}}$.

Queries $q_{\mathsf{sight}}$ and $q_{\mathsf{front}}$ are pseudo-labeled as empty (outside any object), and $q_{\mathsf{behind}}$ is pseudo-labeled as full (inside an object). While this empty pseudo-labeling is intrinsically correct for $q_{\mathsf{sight}}$ and $q_{\mathsf{front}}$, up to acquisition noise, $q_{\mathsf{behind}}$ may not actually be occupied in case the observed object is less than $\delta$ thick, or if $p$ is close to the outline of the silhouette of the object. In our experiments, we use $\delta \ssp= 10$\,cm, which makes the hypothesis largely valid in outdoor scenes.

\subsection{Self-training for UDA.}
Self-training (ST) using pseudo-labels~\cite{zou2019confidence} has been proven a reliable approach for unsupervised domain adaptation.
As exemplified in the literature~\cite{corbiere2021confidence,tranheden2021dacs}, ST is orthogonal to other lines of DA approaches, helping further boost target performance.
We adopt here the common teacher-student learning scheme previously employed in~\cite{tranheden2021dacs, hoyer2022daformer, saltori2022cosmix}. 
In this scheme, two network instances are maintained during training. While the student is trained with an additional supervision from the teacher's pseudo-labels on target data, the teacher has its weights updated as the exponential moving average (EMA) of the student's weights during the course of training.
In SALUDA, we employ a separate self-training step, i.e., Step~2 depicted in~Fig.~\ref{fig:architecture}. The student model is inititialized from model weights obtained from Step~1.

\subsection{Training losses}\label{sec:traininglosses}
The semantic and surface heads both take 3D points $p \ssp\in P$ with associated latent vectors $z_p$ as input,
producing respectively corresponding probabilities of class and occupancy labels. There is one loss for each head.
The semantic loss $\mathcal L_\sem$ is applied to source point clouds $P^\src$, for which we have semantic labels $Y^\src$. It is the usual cross entropy between ${\rm cls}(z_p^\src)_{p \in P^\src}$ and $(Y_p^\src)_{p\in P^\src}$, averaged over all points in $P^\src$.
The occupancy loss $\mathcal L_\occ$ is used with  both source and target point clouds $P^\src, P^\tgt$. It is the binary cross entropy between the occupancy probability~$\tilde o_q$ of query point $q$ and its pseudo-label~$o_q$, averaged over all visibility query points. The training loss for Step 1 balances both terms: $\mathcal L = \mathcal L_\sem \ssp+ \lambda \mathcal L_\occ$, where $\lambda \ssp> 0$ is an hyperparameter. Controlling the $\lambda$ value helps prevent the model from overfitting to the reconstruction task, which harms segmentation performance. 
We do not aim at the most faithful and detailed surface but at the most meaningful object shape, which can be inaccurate as long as it is enough to bridge the geometric gap.
We detail our parameter selection strategy in Sec.~\ref{sec:param_selection}.

In the second step of self-training, we train the student model with $\mathcal L_{\ST}= \mathcal L_{\sem} + \mathcal L_{\PL}$, where the semantic loss $\mathcal L_{\sem}$ on source is the same cross entropy as in Step 1, and $\mathcal L_{\PL}$ is the cross entropy on target using pseudo-labels (PL) obtained from the EMA teacher model. Once trained, only the student model is kept for testing.

\section{Experiments}

\subsection{Baselines}

\paragraph{General domain adaptation baselines.\!\!\!} Following the lidar UDA survey~\cite{triess2021survey}, we consider the following baselines, applied to semantic segmentation: \emph{\minent{}}~\cite{vu2019advent}, \emph{Coral}~\cite{sun2016deep}, and \emph{LogCoral}~\cite{wang2017deep}.
Additionally, we report the performance of the vanilla teacher-student self-training scheme \emph{ST} in~\cite{hoyer2022daformer}, done upon the source-only model.
We also adapt to 3D \emph{AdaBN}~\cite{LI2018109} and \emph{DUA}~\cite{mirza2022dua}, which use \batchnorm{} adaptation with images. Last, \emph{Mixed BN} is defined in Sect.\,\ref{sec:comparison}.

\paragraph{3D-specific approaches.\!\!\!}~A fair comparison to \emph{Complete and Label (C\&L)}~\cite{yi2021complete} is unfortunately impossible because our \emph{Source only} results, i.e., without any domain adaptation, when computed in the same setting, are already much better than the best results reported in~\cite{yi2021complete} and because the code of~\cite{yi2021complete} is not publicly available to reproduce experiments.

The recent 3D UDA approach \emph{CoSMix}~\cite{saltori2022cosmix} proposes to intertwine a source/target mixing strategy with a teacher-student self-training scheme.
To perform mixing, segments from source (resp.\ target) are extracted using (pseudo-)labels and pasted into target (resp.\ source) data.

\subsection{Datasets}
\label{sec:dataset}

\paragraph{nuScenes (NS)~\cite{lidarseg_nuscenes}\!\!\!}
contains 40k lidar frames, sampled from 1000 driving sequences in Boston and Singapore, with a 32-beam rotating lidar. They are annotated with 32 labels (23 foreground classes and 9 background classes).

\paragraph{SynLiDAR (SL)~\cite{xiao2022transfer}\!\!\!}
is a synthetic dataset designed for domain adaptation containing 13 sequences generated in a varied world designed by 3D experts using Unreal Engine~4. A 64-beam rotating lidar is simulated to acquire the points, that are ground-truth annotated with 32 classes.

\paragraph{SemanticKITTI (SK)~\cite{behley2019iccv, geiger2012cvpr}\!\!\!}
contains 22 sequences captured in Karlsruhe with a 64-beam rotating lidar and labeled with 19 classes (single-scan setting). SK is used as one of the target datasets in our experiments. As we do not have access to test-set labels, we use the validation set for evaluation.

\paragraph{SemanticPOSS (SP)~\cite{pan2020semanticposs}\!\!\!}
includes 2988 annotated frames with 14 semantic classes. It is recorded with a 40-beam rotating lidar at the Peking University with a focus on a large appearance of dynamic instances such as bikes or cars. SP is used as a target dataset in our experiments.

\begin{table}[t!]
    \small
    \setlength{\tabcolsep}{3pt}
    \centering
        \begin{tabular}{lr|c|c|c}
            \toprule

             \multicolumn{2}{l|}{Method 
             } & \nstosk  & \synthtosk & \nstoposs \\

            \midrule
            \rowcolor{black!20}
            Source only & &35.9\std{3.2}  & 21.6\std{0.2}    & 62.5\std{0.2}  \\
            \midrule
            \rowcolor{black!10}
            C\&L\,$^\dagger$ & \cite{yi2021complete}& 33.7\,\,\,\,\,\,\,\,\,\, & - & - \\

            \midrule
            AdaBN& \cite{LI2018109}& 40.1\std{0.4}  & 25.6\std{0.2}  & 62.5\std{0.0}  \\ 
            DUA& \cite{mirza2022dua} &  42.9\std{0.7} & 26.4\std{0.4} & 62.3\std{0.1} \\ 
            MixedBN& \llap{(ours)} & \underline{43.3}\std{0.6}  &  27.0\std{0.6}  &  62.4\std{0.1} \\
            MinEnt & \cite{vu2019advent}& \back{43.3\std{0.6}}  &  \back{27.0\std{0.6}}  & 62.6\std{0.1} \\
            Coral & \cite{sun2016deep}  & \back{43.3\std{0.6}}  & 27.3\std{0.3}  & 63.0\std{0.2}   \\
            LogCoral&\cite{wang2017deep} & \back{43.3\std{0.6}}  & \back{27.0\std{0.6}}  &  62.5\std{0.1} \\
            ST & \cite{hoyer2022daformer} & 37.3\std{2.9}  & 26.7\std{0.4} & \underline{65.5}\std{0.2}  \\
            
            {\cosmix}{$^\ddagger$}
            & \cite{saltori2022cosmix} & 38.3\std{2.8}  & \underline{28.0}\std{1.4}   & 65.2\std{0.2} \\
            \method{} & \llap{(ours)} & \textbf{46.2}\std{0.6}  & \textbf{31.2}\std{0.2}     &  \textbf{65.8}\std{0.3}    \\
            
            \bottomrule
        \end{tabular}

                \vspace*{-0.9mm}
    \caption{\textbf{Classwise mIoU\% of 
    sem. segmentation on target} (avg.\ of 3 runs and std dev). All methods except C\&L are evaluated in the same 10-cm voxel size setting. For each approach, hyperparameter grid-search was done to ensure the best possible performance.\\
    \mbox{}\quad $^\dagger$\,from \cite{yi2021complete}, with voxel size 20\,cm, as no code is available.\\
    \mbox{}\quad $^\ddagger$\,different from \cite{saltori2022cosmix}, as retrained (using original code) with voxel size 10\,cm rather than 5\,cm, and evaluated with official metric.}
    
\label{tab:experiments_alls}
\end{table}

\subsection{Experimental setup}
\label{sec:exp_setup}
We address two types of domain shifts: real-to-real (\nstosk, \nstoposs) and 
synthetic-to-real (\synthtosk, \synthtoposs).

\textbf{Common classes aggregation.}~
The UDA setting is such that the source and target domain share the same set of semantic classes. We therefore select and aggregate common classes in the considered datasets. For the~\nstosk~scenario, we follow the class mappings of~\cite{yi2021complete}, with $10$ classes plus one ignore class.  
For the~\synthtosk~and~\synthtoposs~scenarios, we consider respectively $19$ and $13$ classes (and the ignored class), as in \cite{saltori2022cosmix}. For \nstoposs{}, we aggregate into 6 overlapping classes. We denote by SK$_{10}$, SK$_{19}$, SP$_{6}$ and SP$_{13}$ the corresponding versions of SK, respectively SP.

\textbf{Network architecture.}~
As backbone $\phi(\cdot)$, we use a sparse voxel Minkowski U-Net architecture~\cite{choy20194d}, commonly used for automotive \lidar semantic segmentation. 

\textbf{Metrics.}~
We measure the performance with the classwise intersection over union (IoU) and the global mean IoU (mIoU) as done in the official \sk\ benchmark \cite{behley2019iccv}, i.e., computed globally over the whole evaluation dataset. 
(IoUs in \cite{saltori2022cosmix} are not standard: computed per frame then averaged.)

\textbf{Input.}~
Like in~\cite{yi2021complete, saltori2022cosmix}, \lidar intensity is not used as an input feature. It is difficult to  synthesize in simulated datasets and, for real datasets, its calibration may vary a lot from one sensor to another.

\textbf{Training setup.}~
All methods, except CoSMix, are trained using AdamW with a base learning of 0.001 and a weight decay of 0.01, a cosine annealing scheduler for 600k iterations and a batch size of 4. We alternate between source and target batches at every iteration. For \nstosk, we do 10 epochs for the self-training; for \synthtosk{} and \synthtoposs{}, we do 3 epochs of self-training like in \cite{saltori2022cosmix}.
For \cosmix{} warm-up and refinement, we use the official code and parameters.

\textbf{Averaged scores.}~ 
All scores presented in the tables are averaged over three runs. For a fair comparison, we retrained three \cosmix{} models using the official code.

\textbf{Voxel size.}
We use a voxel size of 10\,cm in the main experiments (\cref{tab:experiments_alls}), i.e., \nstosk{}, \synthtosk{}, \nstoposs. We additionally do  \synthtosk{} and \synthtoposs{} with 5-cm voxels for a specific comparison to \cite{saltori2022cosmix} (\cref{tab:experiments_comp_cosmix}).

\begin{figure*}

    \setlength{\tabcolsep}{1pt}
    \centering
    \begin{tabular}{ccccc}
        \includegraphics[width=0.196\linewidth, trim={0 0 0 3cm},clip]{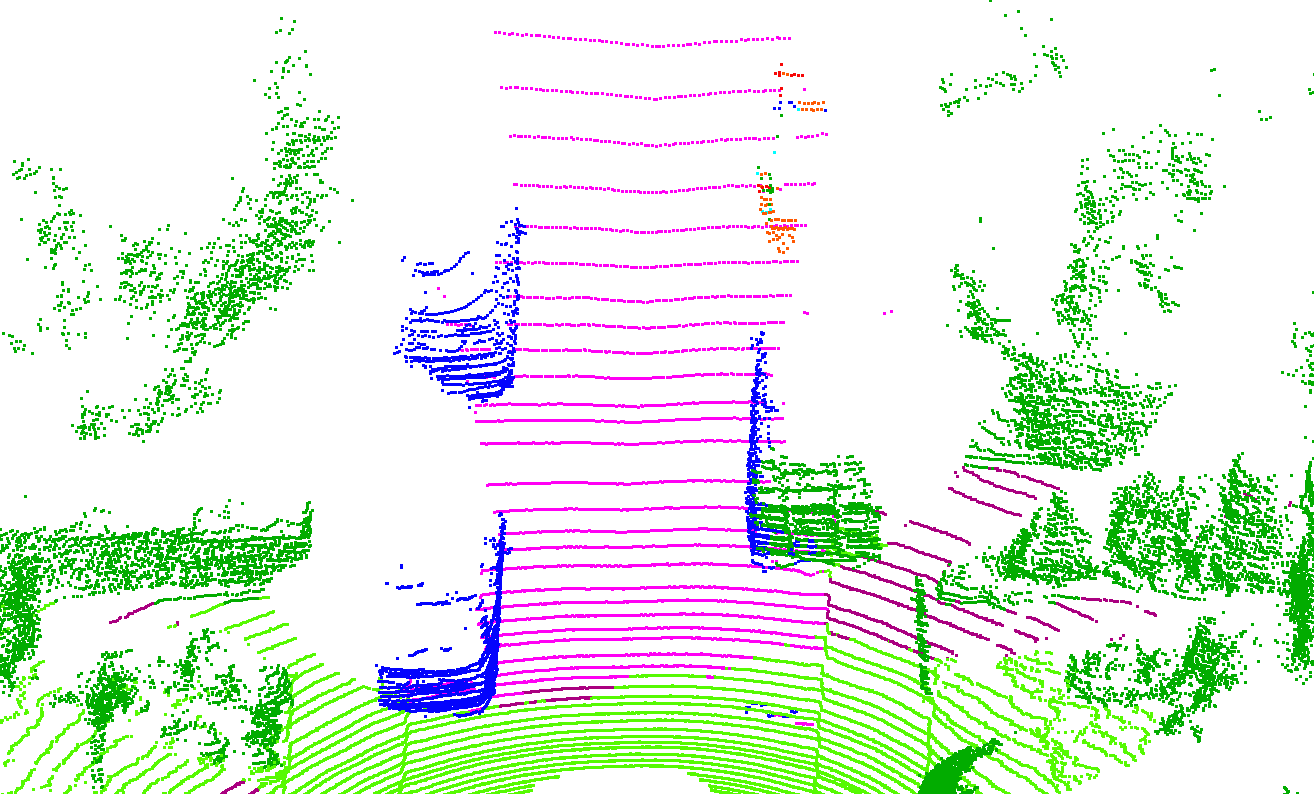}&
        \includegraphics[width=0.196\linewidth, trim={0 0 0 3cm},clip]{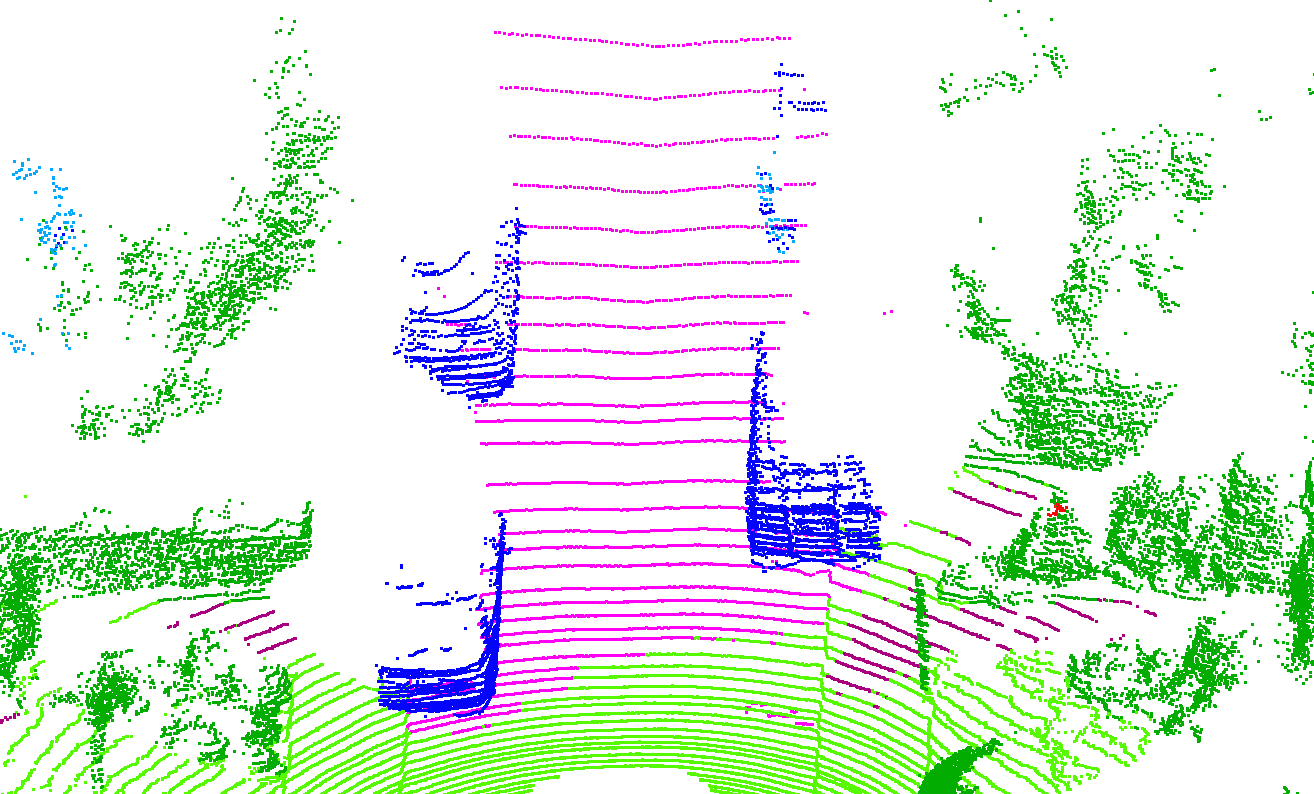}&
        \includegraphics[width=0.196\linewidth, trim={0 0 0 3cm},clip]{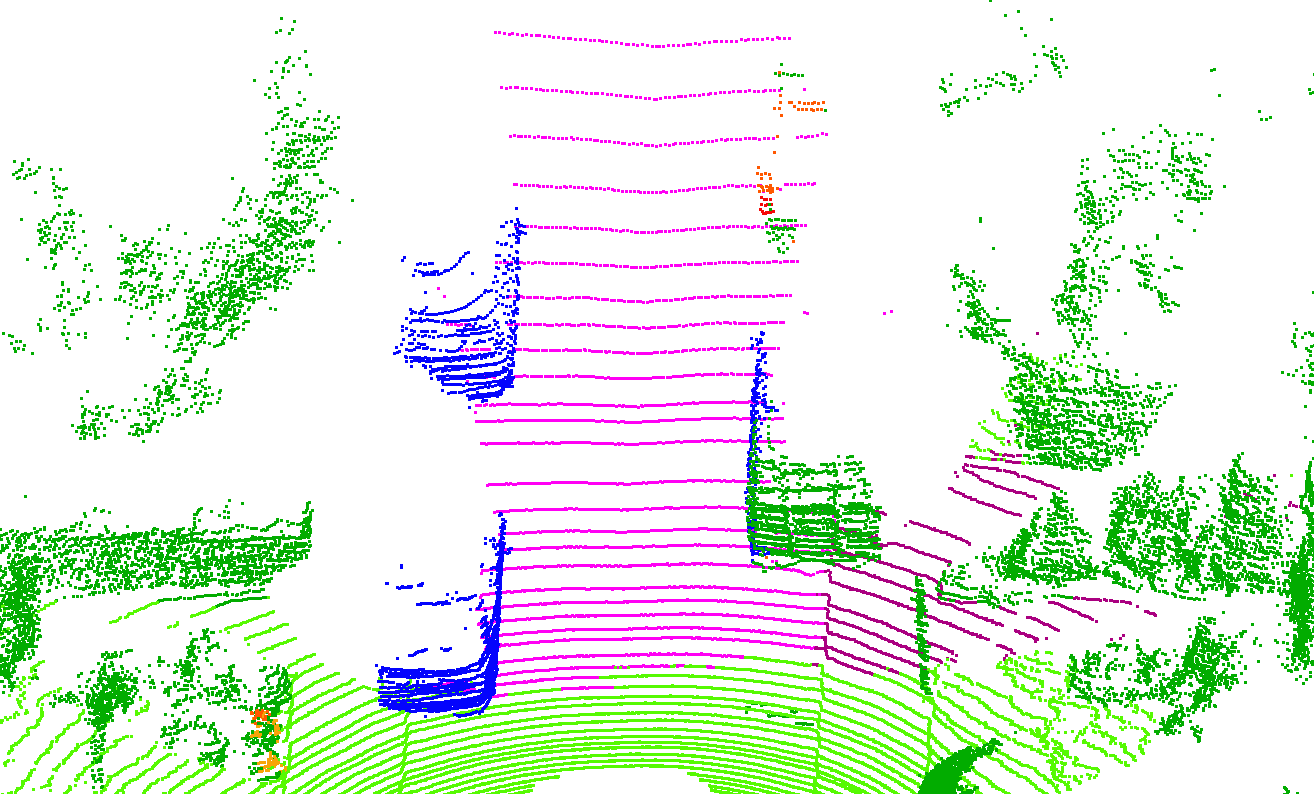}&
    
        \includegraphics[width=0.196\linewidth, trim={0 0 0 3cm},clip]{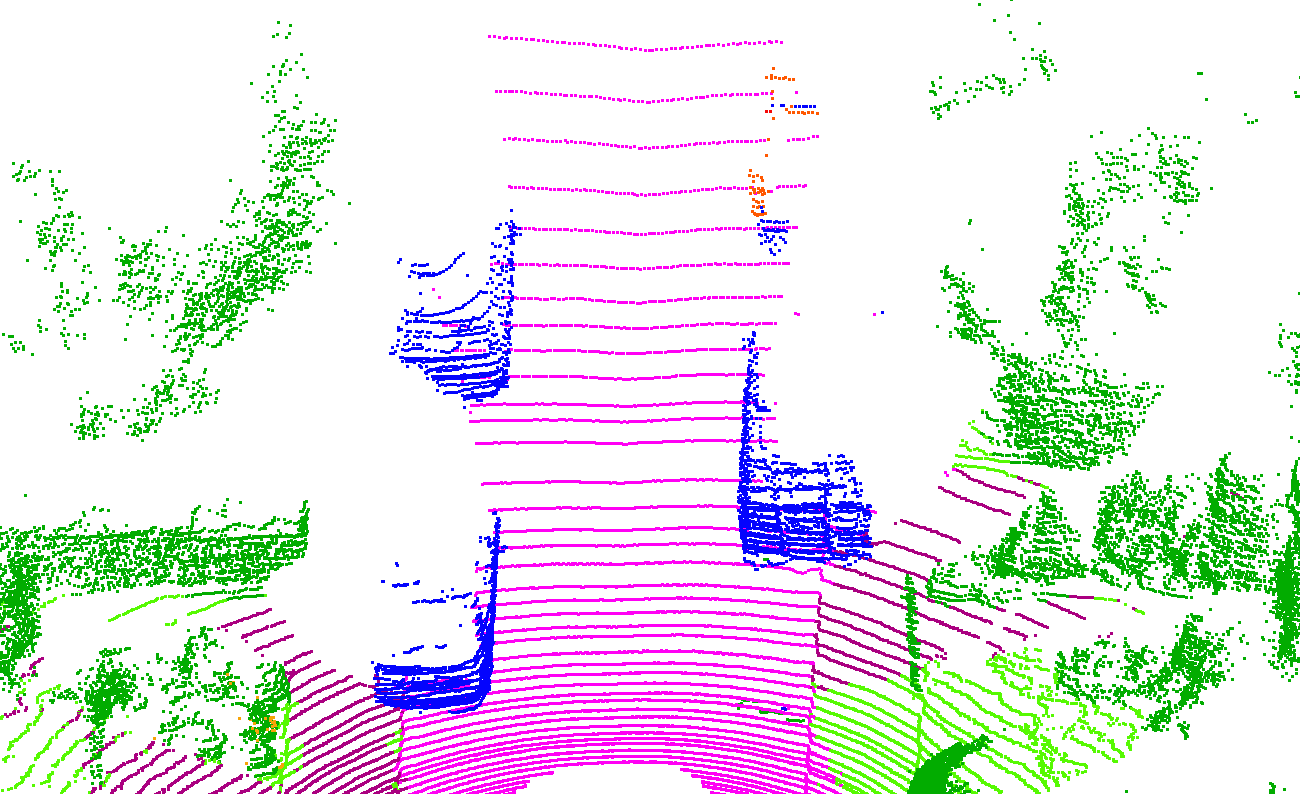}
        &
        \includegraphics[width=0.196\linewidth, trim={0 0 0 3cm},clip]{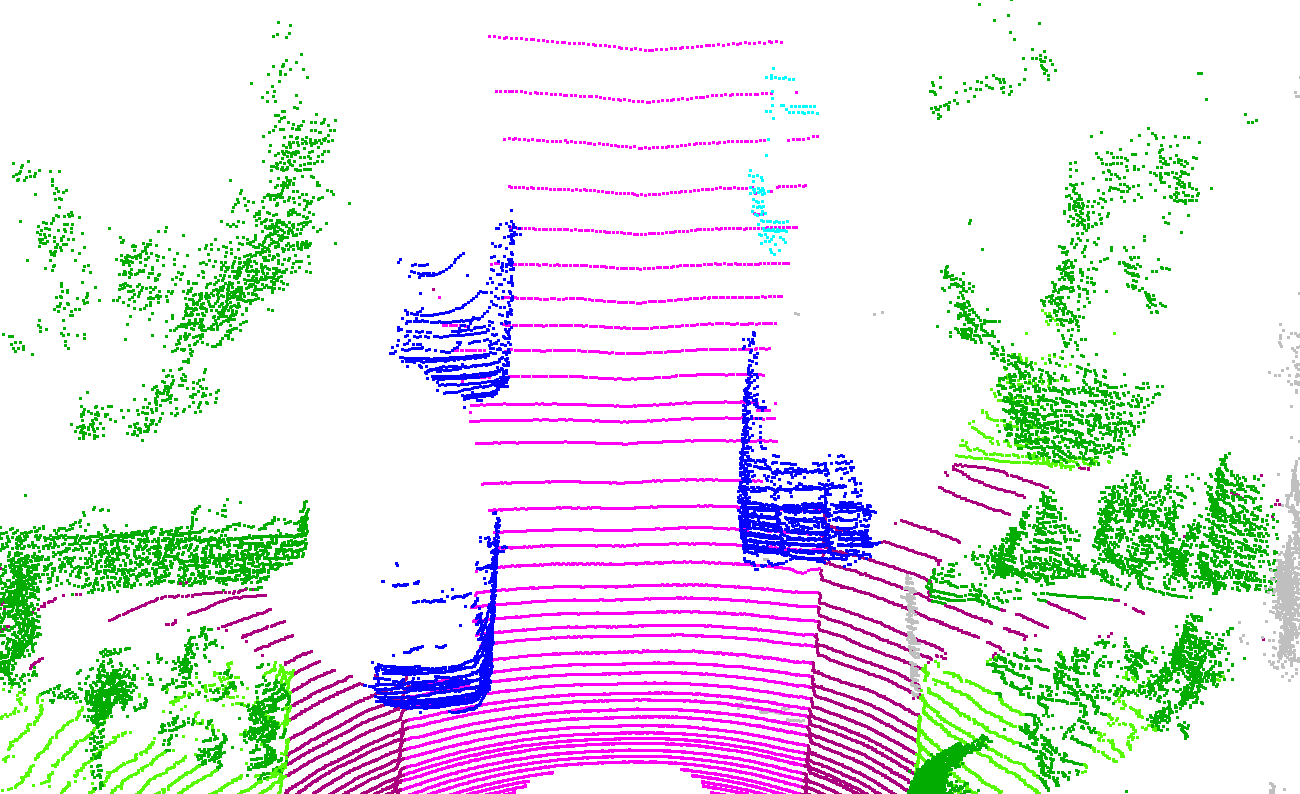}\\
        \includegraphics[width=0.196\linewidth]{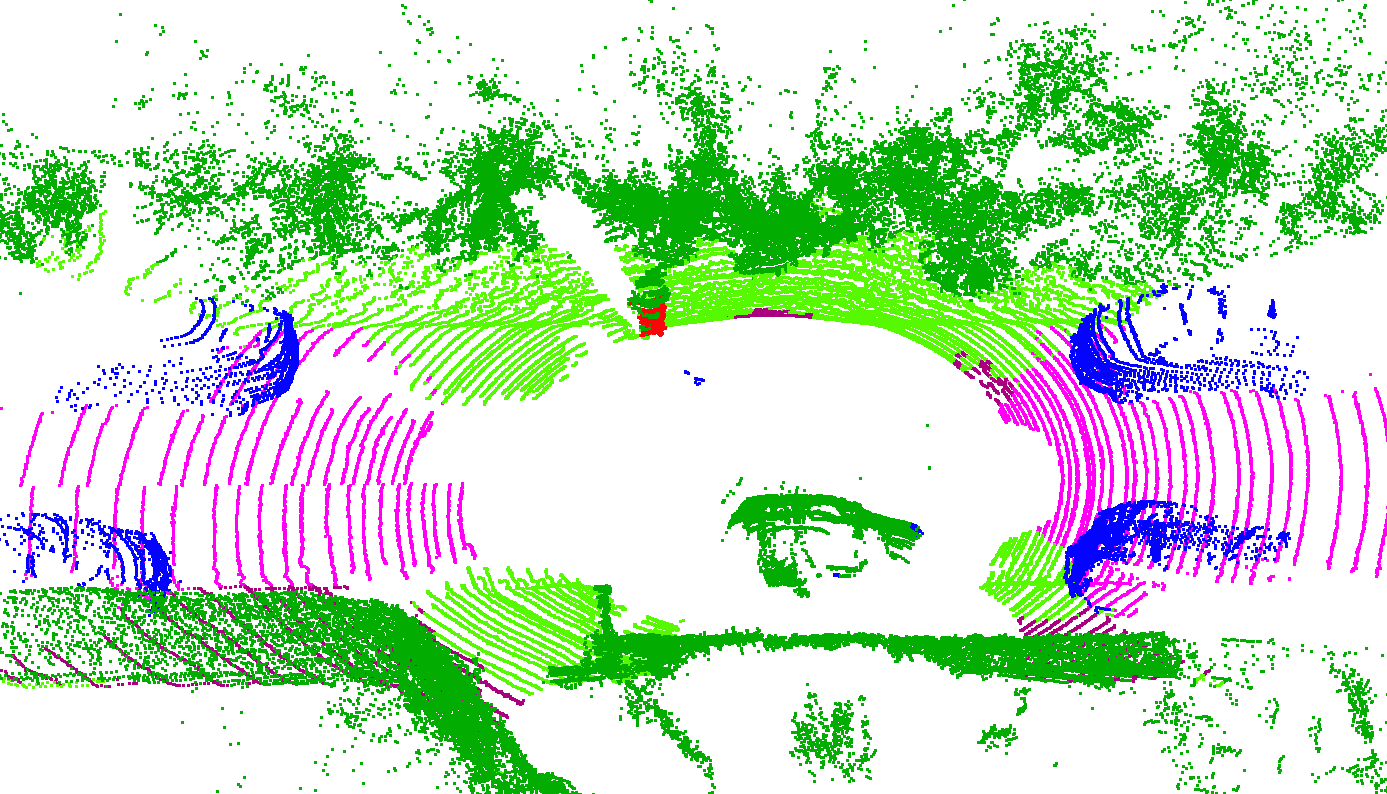}&
        \includegraphics[width=0.196\linewidth]{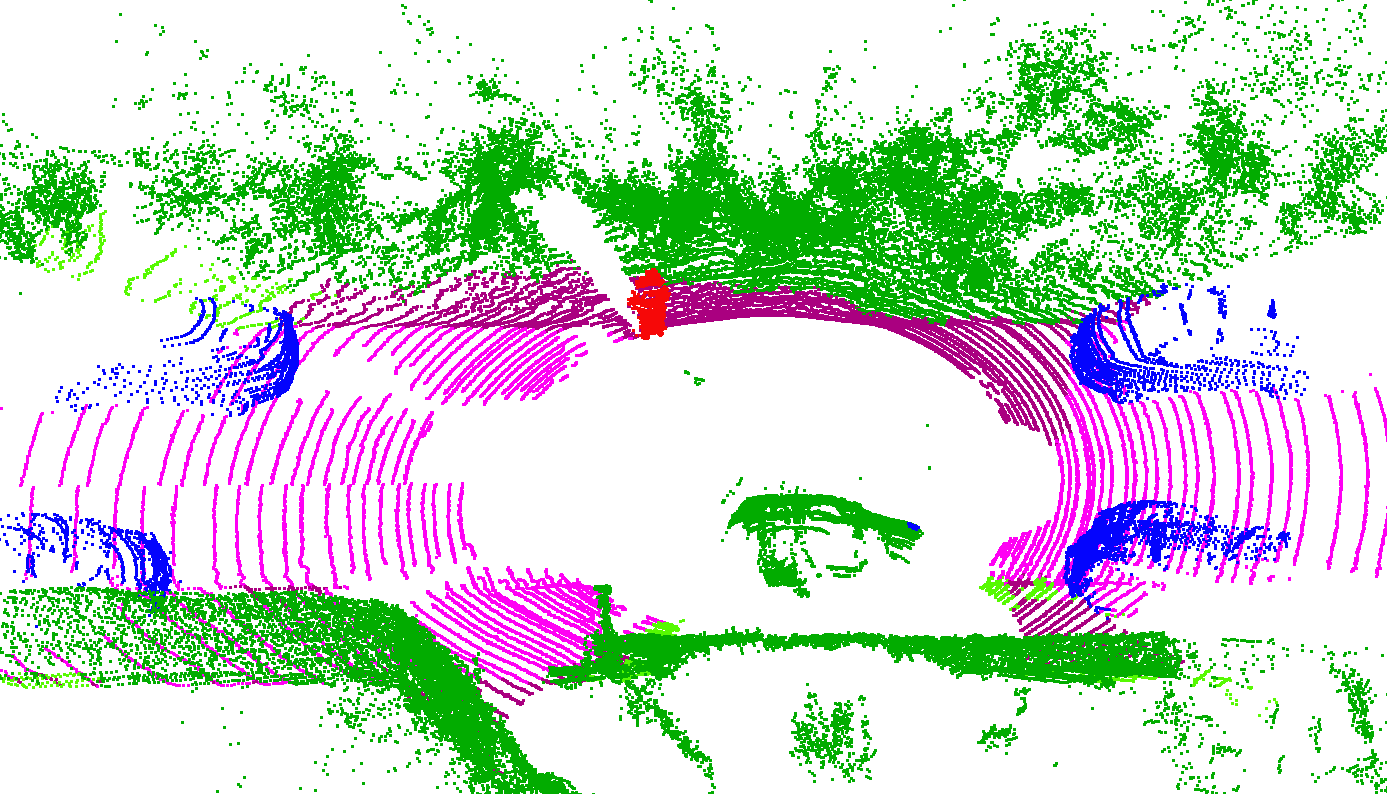}&
        \includegraphics[width=0.196\linewidth]{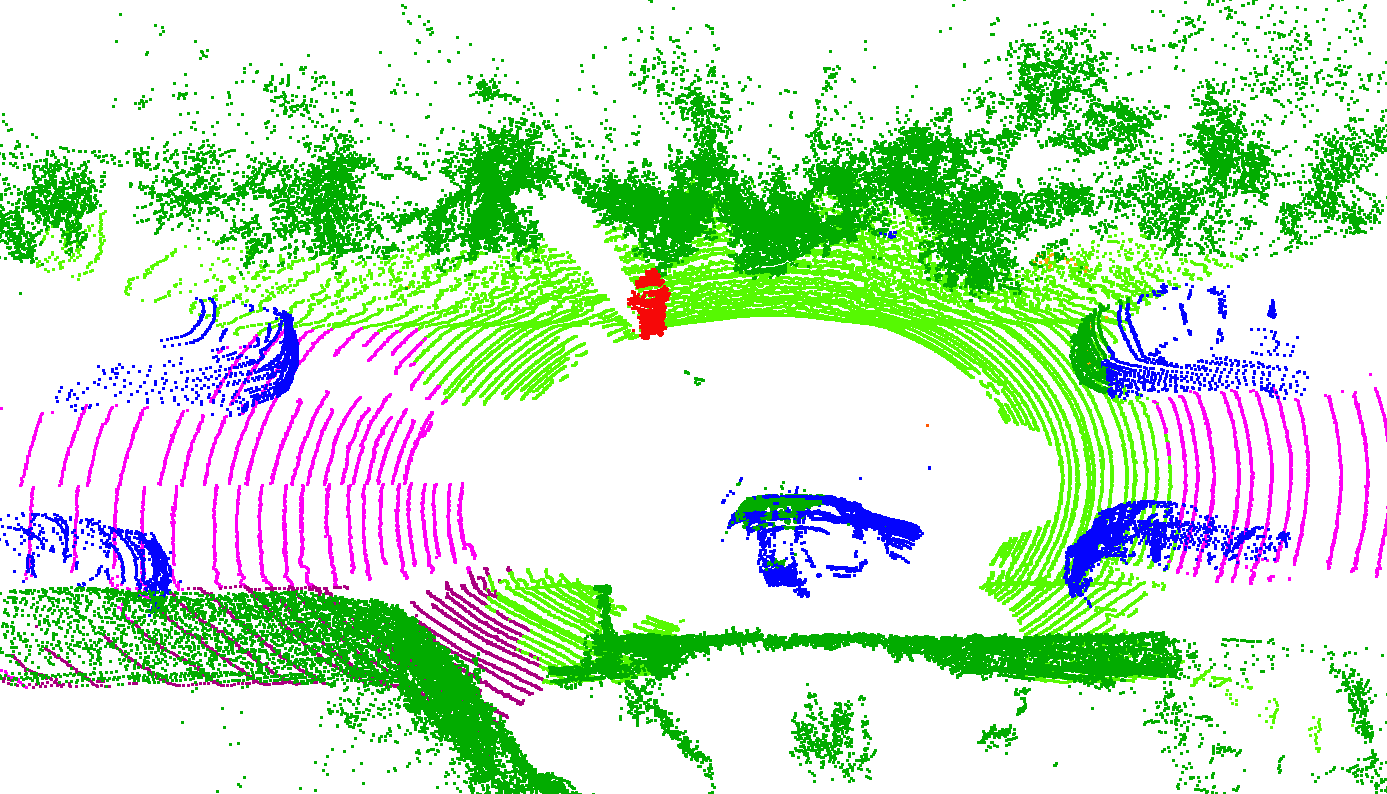}&
      
       \includegraphics[width=0.196\linewidth]{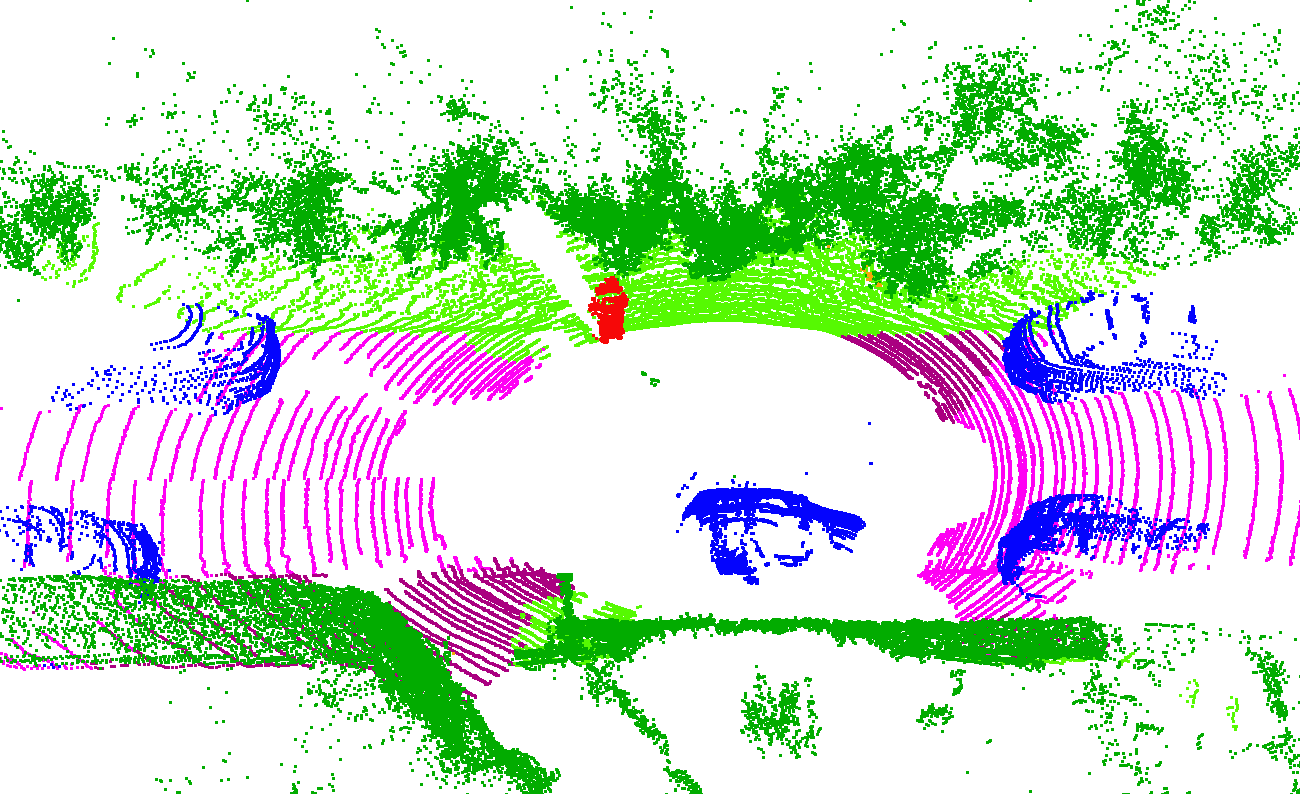}&
       
        \includegraphics[width=0.196\linewidth]{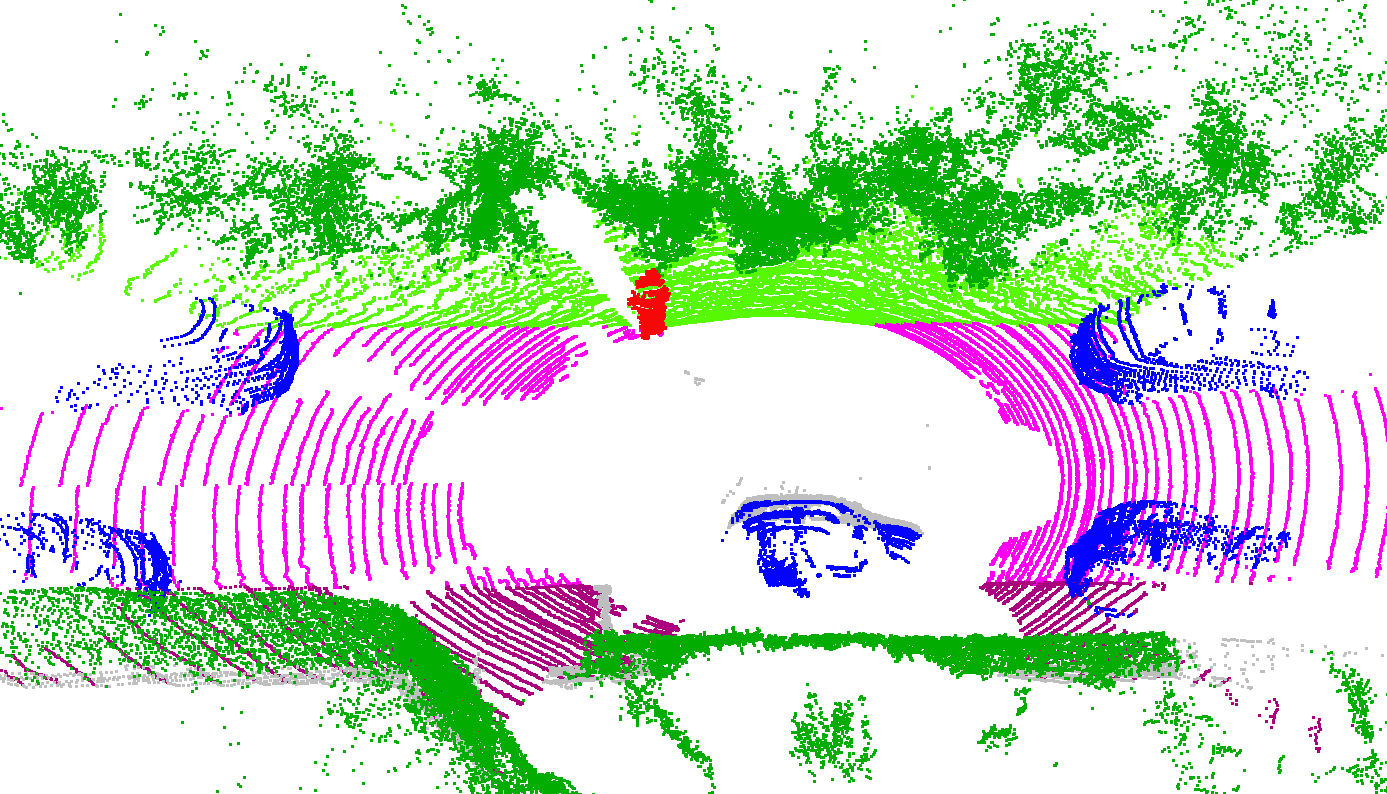}\\
        
        Source-only &
        Mixed BN & CoSMix~\cite{saltori2022cosmix} & \method{} & Ground truth\\
    \end{tabular}
    \caption{\textbf{Visualization of semantic segmentation results.} Obtained with \method{}, source-only, 
    Mixed BN and CoSMix~\cite{saltori2022cosmix} in the setting \DAsetting{NS}{\skns}, along with the ground-truth segmentation. Classes:\colorbox{skcar}{\color{white}car\vphantom{$X_p$}}, \colorbox{skdrivable}{\color{white}drivable surf.\vphantom{$X_p$}},
    \colorbox{skpedestrian}{\color{white}pedestrian\vphantom{$X_p$}},
    \colorbox{sksidewalk}{\color{white}sidewalk\vphantom{$X_p$}}, \colorbox{skterrain}{\color{white}terrain\vphantom{$X_p$}}, \colorbox{skvegetation}{\color{white}vegetation}\vphantom{$X_p$}.}
    \label{fig:qual_res}
    
\end{figure*}

\subsection{Comparison to baselines and SOTA}
\label{sec:comparison}

\cref{tab:experiments_alls} presents the performance of {\method} compared to the baselines. 
For MinEnt, Coral, and LogCoral, we performed a comprehensive hyperparameter search to balance the classification and regularization loss.
To our surprise, we found that a regularization weight of 0 often produced the best results for all three methods. In other words, in our context, the distribution regularization of these baselines is detrimental or, at best, useless.

When using a $0$-weight regularization, the model parameters are updated with gradients computed on source-only data, but the batch normalization (BN) statistics are computed on both source and target data as we still alternate between source and target batches, even with $0$-weight regularization. We refer to this model as the \emph{Mixed BN} baseline.

These results show that \batchnorm{} statistics play a crucial role in 3D domain adaptation. It is confirmed by looking at the performance of AdaBN~\cite{LI2018109} and DUA~\cite{mirza2022dua} (both based on \batchnorm\ adaptation), which is similar to Mixed BN. In particular, with a careful tuning of its two hyperparameters, DUA reaches a performance close to our Mixed BN.

The SOTA approach~\cosmix{} achieves competitive results on synthetic-to-real setups (\synthtosk{} and \synthtoposs{}, cf.~\cref{tab:experiments_alls} and \ref{tab:experiments_comp_cosmix}), while failing to improve over the other baselines in \nstosk{}. As \cosmix{} is a self-training approach enhanced with a mixing strategy, its performance is extremely sensitive to the quality of the warm-up phase, i.e., source-only training. In \nstosk, the source-only training does not perform well enough to provide good initial pseudo labels, presumably because of the large gap in the number of lidar beams (\DAsetting{32}{64}).
We observe the same shortcoming in \nstosk{} for the vanilla ST model, where self-training is employed without the mixing strategy of~\cosmix{}.
In~\method{}, our proposed cross-domain surface reconstruction learning helps to align the source and target representations, resulting in an already well-adapted model after Step~1.
Indeed, the~\method{} w/o ST obtains $44.9\%$ in \nstosk\ (cf.\ Tab.\,\ref{tab:experiments_ablations_data}), outperforming all baselines.
It is further boosted to $46.2\%$ with the self-training Step~2.

Similarly, in \synthtosk, \method{} outperforms \cosmix{} and obtains better results than all other approaches.

In~\nstoposs, the na\"ive source-only model surprisingly achieves a very good result on target, maybe due to the small sensor gap, and most baseline DA models struggle to improve upon it.
In this setting, self-training proves to be the most efficient strategy.
Compared to the vanilla ST,~\cosmix{} suffers from a slight performance drop.
Though~\method{} w/o ST achieves $63.1\%$ (cf.\ supp.\ mat.)
and outperforms the source-only model ($62.5\%$). 
Although, such an improvement does slightly pertain after self-training ($65.8\%$ \vs $65.5\%$) we conjecture that, for settings with small sensor gaps and a strong source-only model like \nstoposs, vanilla ST could be the simple yet sufficient effective DA approach.
Further analyses should be conducted to confirm this assumption.

In~\nstosk, \method{} achieves SOTA results on 6/10 classes, outperforming the source-only model with a large margin on \emph{Car}, \emph{Driveable surface}, \emph{Sidewalk} and \emph{Terrain} (cf.\ supp.\ mat.). 
The latter three classes are often characterized by large surfaces, which could indicate that \method's geometric regularization is particularly beneficial in a specific sensor domain gap for such classes.
In the setting \synthtosk, our method achieves SOTA results on 8/19 classes and is ranked in the top two on 10/19 classes (cf.\ supp.\ mat.). 
However, classes such as \emph{Truck}, \emph{Other vehicle}, \emph{Motorcyclist}, and \emph{Other ground} are particularly challenging to classify correctly. The IoU of the source-only model is less than 4\% on these classes, and no adaptation method goes beyond 8\%.

In \cref{fig:qual_res}, we visualize some qualitative results of \method{} and baselines, showing that our method provides the closest segmentation to the ground truth.

\paragraph{Reported results for \cosmix{}.\!\!}
In our experiments, we universally opt for a voxel size of 10\,cm for all models, including the baselines, the reproduced experiments with 
\cosmix{}~\cite{saltori2022cosmix} (with the authors' code) and~\method{}. In contrast, \cosmix{}~\cite{saltori2022cosmix} reports results with a voxel size of 5\,cm.
We also follow a more standard practice in mIoU computation: our IoU scores are computed over the whole dataset, whereas in~\cite{saltori2022cosmix} those are calculated per scene and then averaged.
All this explains the discrepancy between the~\cosmix{} results in Tab.~\ref{tab:experiments_alls} and the results reported in~\cite{saltori2022cosmix}.

In \cref{tab:experiments_comp_cosmix}, we evaluate the performance of \method{} using the same voxel size of 5\,cm as in~\cite{saltori2022cosmix}, in two settings~\synthtosk{} and~\synthtoposs{}. In \synthtosk{} as well as the ``new'' setting~\synthtoposs{}, \cosmix{} is trained with the official code. We report again the average performance over 3 runs. Our~\method{} obtains better results in both settings. Detailed per-class numbers are given in the Appendix.

\begin{table}[t!]
    \small
    \setlength{\tabcolsep}{3pt}
    \centering
        \begin{tabular}{l@{}r|c|c}
            \toprule

             Method & &  \synthtosk & \synthtoposs \\        
            \midrule
            CoSMix$^\dagger$ & \cite{saltori2022cosmix} &  {29.6}\std{0.8} &  {40.8}\std{0.7}\\ 
            \method{}~~~~~ & \llap{(ours)} & \textbf{30.2}\std{0.4}  & \textbf{42.9}\std{0.7} \\
            
            \bottomrule
        \end{tabular}
        \vspace*{-0.9mm}
   
    \caption{\textbf{Classwise mIoU\% of sem. segmentation on target} (avg.\ of 3 runs and std dev). Both methods are evaluated in the same 5-cm voxel size setting. For each approach, hyperparameter grid-search was done to ensure the best possible performance.\\
    \mbox{}\quad $^\dagger$\,different from \cite{saltori2022cosmix}, as retrained (using original code) and evaluated with official metric.}
\label{tab:experiments_comp_cosmix}
\vspace*{-2mm}
\end{table}

\paragraph{Comparison to geometric based pre-training.\!\!}
While we use the unsupervised surface reconstruction as a shared auxiliary task, on source and target data, one could also use it as a pre-training task similar to  \cite{ALSO}. The motivation for geometric-based pre-training is to learn semantically meaningful features. One could hypothesize that such a pre-training could also lead to more robustness across different sensors. We therefore tested a source-only model in the \nstosk{} setting, which is pre-trained beforehand. The pre-trained models perform significantly worse than the \method{} model on the target domain: pre-training on NS, SK and NS+SK yields respectively 38.7, 39.5 and 34.7 mIoU\%.

\begin{table}[t!]
    \small
    \centering
        \begin{tabular}{l|cc|cc}
            \toprule

              Validator & \multicolumn{2}{c|}{\DAsetting{\ns}{\skns}}  & \multicolumn{2}{c}{\DAsetting{\synth}{\sksyn}}\\
                                & w/o ST & w/ ST & w/o ST & w/ ST \\
            
            \midrule
            \rowcolor{black!10}
            Oracle & {44.9} & {46.2} & {27.6} & {31.2} \\ 
            \midrule
            Entropy & \textbf{44.8} & \textbf{46.2} &  \textbf{27.6} & \textbf{31.2} \\
            IM      & {44.0} & 45.3 &  26.6 &  30.0 \\
            SrcVal  & {43.3}\rlap{$^\dagger$} &  43.7 &  27.0\rlap{$^\dagger$} &  29.7 \\
            \bottomrule
            \multicolumn{5}{l}{$^{\dagger^{\phantom{X}}}$: selection of no regularization = Mixed BN.}
        \end{tabular}
    \caption{\textbf{Oracle vs validator performance.} The validator selection is applied on \method{} w/o ST (first column per setting) and then self-training is done on the selected model (second column per setting). We measure mIoU\% averaged over 3 runs.}
\label{tab:experiments_comparison_sota_direct_im}
\end{table}

\subsection{Pure UDA hyperparameter selection}
\label{sec:param_selection}

Hyperparameter tuning in UDA is tricky as, in principle, no target label is available to measure accuracy.
Previous work relax the unsupervised constraint in UDA and allow using a 
target validation set with labels to select the hyperparameter that works best on the target test set.
This practice, however, does not truly reflect the real performance in a pure unsupervised setting.
In~\cite{musgrave2022benchmarking}, the authors propose a set of unsupervised \emph{validators} that can replace the use of a target validation set.
Extensive studies in~\cite{musgrave2022benchmarking} indicate strong correlations between selections made by 
proposed validators and the choice based on target validation performance.
Although~\cite{musgrave2022benchmarking} only considers the image classification task, we argue for the possible adoption of this validation protocol for point cloud segmentation, and apply it to \method{}.

To that end, we sample 7 values for the regularization parameter $\lambda$, in the range $[0, 1]$.
To take into account the variation in performance,
we train two models starting from different random seeds for every hyperparameter, resulting in 14 runs. Then, we select the hyperparameter with best average validator score, train a third model, and use the so-trained models for initialization for the self-training.

In \cref{tab:experiments_comparison_sota_direct_im}, we report the scores of \method{} before and after self-training with purely UDA hyperparameter selection. We compare three validators from~\cite{musgrave2022benchmarking}: source validation (SrcVal), target entropy (Entropy), and information maximization (IM).
We select these validators due to their reported good performance, but also due to their ease of application without any further training need. This could be a motivation to compare also in this strict setting in further UDA work.

All validators select meaningful models. However, using performance on SrcVal as a metric for hyperparameter selection leads to the choice of Mixed BN. 
This corresponds to the model with minimal deviation from source-only training, which indeed gives the best performances on the source validation set.
Entropy and IM make more relevant choices of model. 
Entropy even chooses models that are identical or very close to optimal/oracle parameter set.

\begin{table}
\small
\setlength{\tabcolsep}{2pt}
\centering
\begin{tabular}{c|cc|cc|cc|c|c}
\toprule
\multicolumn{1}{l}{NS}&Source&Target &\multicolumn{2}{c|}{$\mathcal{L}_\occ$ applied to} &  \multicolumn{2}{c|}{Surf. head} & ST &mIoU\\
\multicolumn{1}{l}{$\shortto$\skns\!\!\!}&data& data &Source & Target & POCO & ALSO & & in \% \\
\midrule
(a)&\cmark& & & & & && 35.9 \\
(b)&\cmark& &\cmark &  & & \cmark & & 34.3 \\
(c)&\cmark& \cmark & & & & & &43.3 \\
(d)&\cmark&\cmark& & \cmark & & \cmark &  & 39.2 \\
(e)&\cmark&\cmark& \cmark & & & \cmark &  & 41.9 \\

(f)&\cmark&\cmark&\cmark & \cmark & \cmark &  & & 44.2 \\
\rowcolor{blue!10}
(g)&\cmark&\cmark&\cmark & \cmark &       & \cmark &  & \underline{44.9} \\
\rowcolor{blue!20}
(h)&\cmark&\cmark&\cmark & \cmark &       & \cmark & \cmark & \textbf{46.2} \\
\bottomrule
\end{tabular}

\caption{\textbf{Ablation study.} (c) is Mixed BN. (h) is full \method{}.}
\label{tab:experiments_ablations_data}
\end{table}

\subsection{Ablation study}
\label{sec:ablation_study}

We study the influence of the reconstruction head and the use of the target data.
Results are summarized
in~\cref{tab:experiments_ablations_data}.

Starting from source-only~(a), we study if using only source data with surface regularization~(b) can construct a latent space structured enough to reduce the domain gap between source and target domains. It is not the case; the performance is even lower than the naive baseline. Next, we leverage target data without applying the reconstruction loss~(c). It corresponds to our Mixed BN.
Training with surface regularization either only on source data~(d) or target data~(e) does not improve over Mixed BN. As intended, this additional regularization is helpful when used on both datasets (f, g).
$\mathcal{L}_\occ$ operates a soft alignment between the two latent spaces, letting optimization select which part of the space should be dedicated to source or target reconstruction. Finally, we ablate the ball-based reconstruction head~(g), borrowed from ALSO \cite{ALSO}, and replace it with the POCO head~\cite{Boulch_2022_CVPR}, which relies on nearest neighbors~(f).
While the POCO head targets accurate reconstruction, distributing geometrical details on input points, the ALSO head (and loss) departs from accurate reconstruction and encourages each point in a ball to be able to reconstruct alone all the surface in this ball, thus favoring feature sharing within objects and setting the stage for semantic segmentation. A final improvement is obtained with a self-training (h).

\subsection{Per-distance results}

The largest performance drop of the source-only model occurs close to the lidar sensor (cf. Tab.~5 supp. mat.). Due to the higher point density in this area, the network might learn finer patterns, leading to better results in the supervised setting, but less robust to a domain change. Nevertheless, \method{} is the most successful in the two closer bands. The geometric regularization is mostly helpful where point density is sufficient for good surface estimation.

\section{Conclusion}
This paper has introduced \method{}, a new approach to mitigate the domain shift problem occurring for the semantic segmentation of automotive \lidar{} point clouds. Our proposal, which stresses the importance of mixing the batch norm statistics, is based on a geometric regularization that imposes the latent point representations to also perform well in an unsupervised surface reconstruction task on both domains. We show that \method{} achieves SOTA performance on challenging datasets, including in a pure UDA setting, which would be a more realistic comparison for future works.

\footnotesize{\noindent\paragraph{Acknowledgements:}We acknowledge the support of the French Agence Nationale de la Recherche (ANR), under grant ANR-21-CE23-0032 (project MultiTrans). This work was performed using HPC resources from GENCI–IDRIS (Grants 2022-AD011012883R1, 2023-AD011012883R2, 2022-AD011013839).}

{
    \small
    \bibliographystyle{ieeenat_fullname}
    \bibliography{egbib}

\begin{thebibliography}{91}
\providecommand{\natexlab}[1]{#1}
\providecommand{\url}[1]{\texttt{#1}}
\expandafter\ifx\csname urlstyle\endcsname\relax
  \providecommand{\doi}[1]{doi: #1}\else
  \providecommand{\doi}{doi: \begingroup \urlstyle{rm}\Url}\fi

\bibitem[Achituve et~al.(2021)Achituve, Maron, and Chechik]{achituve2021self}
Idan Achituve, Haggai Maron, and Gal Chechik.
\newblock Self-supervised learning for domain adaptation on point clouds.
\newblock In \emph{WACV}, 2021.

\bibitem[Alonso et~al.(2020)Alonso, Montesano, Murillo, et~al.]{alonso2020domain}
Inigo Alonso, Luis~Riazuelo Montesano, Ana~C Murillo, et~al.
\newblock Domain adaptation in lidar semantic segmentation by aligning class distributions.
\newblock In \emph{ICINCO}, 2020.

\bibitem[Barrera et~al.(2021)Barrera, Beltr{\'a}n, Guindel, Iglesias, and Garc{\'\i}a]{barrera2021cycle}
Alejandro Barrera, Jorge Beltr{\'a}n, Carlos Guindel, Jose~Antonio Iglesias, and Fernando Garc{\'\i}a.
\newblock Cycle and semantic consistent adversarial domain adaptation for reducing simulation-to-real domain shift in lidar bird's eye view.
\newblock In \emph{ITSC}, 2021.

\bibitem[Behley et~al.(2019)Behley, Garbade, Milioto, Quenzel, Behnke, Stachniss, and Gall]{behley2019iccv}
J. Behley, M. Garbade, A. Milioto, J. Quenzel, S. Behnke, C. Stachniss, and J. Gall.
\newblock {SemanticKITTI: A Dataset for Semantic Scene Understanding of LiDAR Sequences}.
\newblock In \emph{ICCV}, 2019.

\bibitem[Boulch and Marlet(2022)]{Boulch_2022_CVPR}
Alexandre Boulch and Renaud Marlet.
\newblock Poco: Point convolution for surface reconstruction.
\newblock In \emph{CVPR}, 2022.

\bibitem[Boulch et~al.(2023)Boulch, Sautier, Michele, Puy, and Marlet]{ALSO}
Alexandre Boulch, Corentin Sautier, Björn Michele, Gilles Puy, and Renaud Marlet.
\newblock Also: Automotive lidar self-supervision by occupancy estimation.
\newblock In \emph{CVPR}, 2023.

\bibitem[Chen and Zhang(2019)]{chen2019learning}
Zhiqin Chen and Hao Zhang.
\newblock Learning implicit fields for generative shape modeling.
\newblock In \emph{CVPR}, 2019.

\bibitem[Choi et~al.(2018)Choi, Choi, Kim, Ha, Kim, and Choo]{choi2018stargan}
Yunjey Choi, Minje Choi, Munyoung Kim, Jung-Woo Ha, Sunghun Kim, and Jaegul Choo.
\newblock Stargan: Unified generative adversarial networks for multi-domain image-to-image translation.
\newblock In \emph{CVPR}, 2018.

\bibitem[Choy et~al.(2019)Choy, Gwak, and Savarese]{choy20194d}
Christopher Choy, JunYoung Gwak, and Silvio Savarese.
\newblock 4d spatio-temporal convnets: Minkowski convolutional neural networks.
\newblock In \emph{CVPR}, 2019.

\bibitem[Corbiere et~al.(2021)Corbiere, Thome, Saporta, Vu, Cord, and Perez]{corbiere2021confidence}
Charles Corbiere, Nicolas Thome, Antoine Saporta, Tuan-Hung Vu, Matthieu Cord, and Patrick Perez.
\newblock Confidence estimation via auxiliary models.
\newblock \emph{IEEE Transactions on Pattern Analysis and Machine Intelligence}, 2021.

\bibitem[Damodaran et~al.(2018)Damodaran, Kellenberger, Flamary, Tuia, and Courty]{damodaran2018deepjdot}
Bharath~Bhushan Damodaran, Benjamin Kellenberger, R{\'e}mi Flamary, Devis Tuia, and Nicolas Courty.
\newblock Deepjdot: Deep joint distribution optimal transport for unsupervised domain adaptation.
\newblock In \emph{ECCV}, 2018.

\bibitem[Daume~III and Marcu(2006)]{daume2006domain}
Hal Daume~III and Daniel Marcu.
\newblock Domain adaptation for statistical classifiers.
\newblock \emph{JAIR}, 2006.

\bibitem[DeBortoli et~al.(2021)DeBortoli, Fuxin, Kapoor, and Hollinger]{debortoli2021adversarial}
Robert DeBortoli, Li Fuxin, Ashish Kapoor, and Geoffrey~A Hollinger.
\newblock Adversarial training on point clouds for sim-to-real 3d object detection.
\newblock \emph{RA-L}, 2021.

\bibitem[Fan et~al.(2022)Fan, Chang, Zhang, Cheng, Sun, and Kankanhalli]{fan2022self}
Hehe Fan, Xiaojun Chang, Wanyue Zhang, Yi Cheng, Ying Sun, and Mohan Kankanhalli.
\newblock Self-supervised global-local structure modeling for point cloud domain adaptation with reliable voted pseudo labels.
\newblock In \emph{CVPR}, 2022.

\bibitem[Fatras et~al.(2021)Fatras, S\'ejourn\'e, Courty, and Flamary]{fatras2021jumbot}
Kilian Fatras, Thibault S\'ejourn\'e, Nicolas Courty, and R\'emi Flamary.
\newblock Unbalanced minibatch optimal transport; applications to domain adaptation.
\newblock In \emph{ICML}, 2021.

\bibitem[Fei et~al.(2022)Fei, Huang, Yuan, Shi, Zhang, Chen, Dou, and Qiao]{fei2022adas}
Ben Fei, Siyuan Huang, Jiakang Yuan, Botian Shi, Bo Zhang, Tao Chen, Min Dou, and Yu Qiao.
\newblock Adas: A simple active-and-adaptive baseline for cross-domain 3d semantic segmentation.
\newblock \emph{arXiv preprint arXiv:2212.10390}, 2022.

\bibitem[Fong et~al.(2021)Fong, Mohan, Hurtado, Zhou, Caesar, Beijbom, and Valada]{lidarseg_nuscenes}
Whye~Kit Fong, Rohit Mohan, Juana~Valeria Hurtado, Lubing Zhou, Holger Caesar, Oscar Beijbom, and Abhinav Valada.
\newblock Panoptic nuscenes: A large-scale benchmark for lidar panoptic segmentation and tracking.
\newblock \emph{RA-L}, 2021.

\bibitem[Ganin et~al.(2016)Ganin, Ustinova, Ajakan, Germain, Larochelle, Laviolette, Marchand, and Lempitsky]{ganin2016domain}
Yaroslav Ganin, Evgeniya Ustinova, Hana Ajakan, Pascal Germain, Hugo Larochelle, Fran{\c{c}}ois Laviolette, Mario Marchand, and Victor Lempitsky.
\newblock Domain-adversarial training of neural networks.
\newblock \emph{JMLR}, 2016.

\bibitem[Geiger et~al.(2012)Geiger, Lenz, and Urtasun]{geiger2012cvpr}
A. Geiger, P. Lenz, and R. Urtasun.
\newblock {Are we ready for Autonomous Driving? The KITTI Vision Benchmark Suite}.
\newblock In \emph{CVPR}, 2012.

\bibitem[Hoffman et~al.(2018)Hoffman, Tzeng, Park, Zhu, Isola, Saenko, Efros, and Darrell]{hoffman2018cycada}
Judy Hoffman, Eric Tzeng, Taesung Park, Jun-Yan Zhu, Phillip Isola, Kate Saenko, Alexei Efros, and Trevor Darrell.
\newblock Cycada: Cycle-consistent adversarial domain adaptation.
\newblock In \emph{ICLR}, 2018.

\bibitem[Hoyer et~al.(2022)Hoyer, Dai, and Van~Gool]{hoyer2022daformer}
Lukas Hoyer, Dengxin Dai, and Luc Van~Gool.
\newblock Daformer: Improving network architectures and training strategies for domain-adaptive semantic segmentation.
\newblock In \emph{CVPR}, 2022.

\bibitem[Ioffe and Szegedy(2015)]{ioffe2015batch}
Sergey Ioffe and Christian Szegedy.
\newblock Batch normalization: Accelerating deep network training by reducing internal covariate shift.
\newblock In \emph{International conference on machine learning}. PMLR, 2015.

\bibitem[Jaritz et~al.(2020)Jaritz, Vu, Charette, Wirbel, and P{\'e}rez]{jaritz2020xmuda}
Maximilian Jaritz, Tuan-Hung Vu, Raoul~de Charette, Emilie Wirbel, and Patrick P{\'e}rez.
\newblock xmuda: Cross-modal unsupervised domain adaptation for 3d semantic segmentation.
\newblock In \emph{CVPR}, 2020.

\bibitem[Jaritz et~al.(2022)Jaritz, Vu, de~Charette, Wirbel, and P{\'e}rez]{jaritz2022cross}
Maximilian Jaritz, Tuan-Hung Vu, Raoul de Charette, Emilie Wirbel, and Patrick P{\'e}rez.
\newblock Cross-modal learning for domain adaptation in {3D} semantic segmentation.
\newblock In \emph{T-PAMI}, 2022.

\bibitem[Jiang and Saripalli(2021)]{jiang2021lidarnet}
Peng Jiang and Srikanth Saripalli.
\newblock Lidarnet: A boundary-aware domain adaptation model for point cloud semantic segmentation.
\newblock In \emph{ICRA}, 2021.

\bibitem[Kim et~al.(2023)Kim, Kang, Oh, and Yoon]{Kim_2023_CVPR}
Hyeonseong Kim, Yoonsu Kang, Changgyoon Oh, and Kuk-Jin Yoon.
\newblock Single domain generalization for lidar semantic segmentation.
\newblock In \emph{CVPR}, 2023.

\bibitem[Kong et~al.(2023)Kong, Quader, and Liong]{kong2023conda}
Lingdong Kong, Niamul Quader, and Venice~Erin Liong.
\newblock Conda: Unsupervised domain adaptation for lidar segmentation via regularized domain concatenation.
\newblock In \emph{ICRA}, 2023.

\bibitem[Laine and Aila(2017)]{laine2016temporal}
Samuli Laine and Timo Aila.
\newblock Temporal ensembling for semi-supervised learning.
\newblock In \emph{ICLR}, 2017.

\bibitem[Langer et~al.(2020)Langer, Milioto, Haag, Behley, and Stachniss]{langer2020domain}
Ferdinand Langer, Andres Milioto, Alexandre Haag, Jens Behley, and Cyrill Stachniss.
\newblock Domain transfer for semantic segmentation of lidar data using deep neural networks.
\newblock In \emph{IROS}, 2020.

\bibitem[Li et~al.(2023)Li, Kang, Wang, Wei, and Yang]{Li_2023_CVPR}
Guangrui Li, Guoliang Kang, Xiaohan Wang, Yunchao Wei, and Yi Yang.
\newblock Adversarially masking synthetic to mimic real: Adaptive noise injection for point cloud segmentation adaptation.
\newblock In \emph{CVPR}, 2023.

\bibitem[Li et~al.(2018)Li, Wang, Shi, Hou, and Liu]{LI2018109}
Yanghao Li, Naiyan Wang, Jianping Shi, Xiaodi Hou, and Jiaying Liu.
\newblock Adaptive batch normalization for practical domain adaptation.
\newblock \emph{PR}, 80, 2018.

\bibitem[Liang et~al.(2022)Liang, Fan, Fan, Wang, Chen, Cheng, and Wang]{liangpoint}
Hanxue Liang, Hehe Fan, Zhiwen Fan, Yi Wang, Tianlong Chen, Yu Cheng, and Zhangyang Wang.
\newblock Point cloud domain adaptation via masked local 3d structure prediction.
\newblock In \emph{ECCV}, 2022.

\bibitem[Liu et~al.(2021)Liu, Luo, Cai, Yu, Ke, Junior, Gonçalves, and Li]{LIU2021211}
Wei Liu, Zhiming Luo, Yuanzheng Cai, Ying Yu, Yang Ke, José~Marcato Junior, Wesley~Nunes Gonçalves, and Jonathan Li.
\newblock Adversarial unsupervised domain adaptation for 3d semantic segmentation with multi-modal learning.
\newblock \emph{ISPRS}, 2021.

\bibitem[Long et~al.(2015)Long, Cao, Wang, and Jordan]{long2015learning}
Mingsheng Long, Yue Cao, Jianmin Wang, and Michael Jordan.
\newblock Learning transferable features with deep adaptation networks.
\newblock In \emph{ICML}, 2015.

\bibitem[Long et~al.(2017)Long, Zhu, Wang, and Jordan]{long2017deep}
Mingsheng Long, Han Zhu, Jianmin Wang, and Michael~I Jordan.
\newblock Deep transfer learning with joint adaptation networks.
\newblock In \emph{ICML}, 2017.

\bibitem[Long et~al.(2018)Long, Cao, Wang, and Jordan]{long2018conditional}
Mingsheng Long, Zhangjie Cao, Jianmin Wang, and Michael~I Jordan.
\newblock Conditional adversarial domain adaptation.
\newblock In \emph{NeurIPS}, 2018.

\bibitem[Loshchilov and Hutter(2019)]{loshchilov2019decoupled}
Ilya Loshchilov and Frank Hutter.
\newblock Decoupled weight decay regularization.
\newblock In \emph{ICLR}, 2019.

\bibitem[Luo et~al.(2021{\natexlab{a}})Luo, Liu, Fu, Wang, and Song]{luo2021learnable}
Xiaoyuan Luo, Shaolei Liu, Kexue Fu, Manning Wang, and Zhijian Song.
\newblock A learnable self-supervised task for unsupervised domain adaptation on point clouds.
\newblock In \emph{WACV}, 2021{\natexlab{a}}.

\bibitem[Luo et~al.(2021{\natexlab{b}})Luo, Cai, Zhou, Zhang, Zhao, Yi, Lu, Li, Zhang, and Liu]{Luo_2021_ICCV}
Zhipeng Luo, Zhongang Cai, Changqing Zhou, Gongjie Zhang, Haiyu Zhao, Shuai Yi, Shijian Lu, Hongsheng Li, Shanghang Zhang, and Ziwei Liu.
\newblock Unsupervised domain adaptive 3d detection with multi-level consistency.
\newblock In \emph{ICCV}, 2021{\natexlab{b}}.

\bibitem[Mescheder et~al.(2019)Mescheder, Oechsle, Niemeyer, Nowozin, and Geiger]{mescheder2019occupancy}
Lars Mescheder, Michael Oechsle, Michael Niemeyer, Sebastian Nowozin, and Andreas Geiger.
\newblock Occupancy networks: Learning 3d reconstruction in function space.
\newblock In \emph{CVPR}, 2019.

\bibitem[Mirza et~al.(2022)Mirza, Micorek, Possegger, and Bischof]{mirza2022dua}
M.~Jehanzeb Mirza, Jakub Micorek, Horst Possegger, and Horst Bischof.
\newblock The norm must go on: Dynamic unsupervised domain adaptation by normalization.
\newblock In \emph{CVPR}, 2022.

\bibitem[Musgrave et~al.(2022)Musgrave, Belongie, and Lim]{musgrave2022benchmarking}
Kevin Musgrave, Serge Belongie, and Ser-Nam Lim.
\newblock Benchmarking validation methods for unsupervised domain adaptation.
\newblock \emph{arXiv preprint arXiv:2208.07360}, 2022.

\bibitem[Nado et~al.(2020)Nado, Padhy, Sculley, D'Amour, Lakshminarayanan, and Snoek]{nado2020evaluating}
Zachary Nado, Shreyas Padhy, D Sculley, Alexander D'Amour, Balaji Lakshminarayanan, and Jasper Snoek.
\newblock Evaluating prediction-time batch normalization for robustness under covariate shift.
\newblock \emph{arXiv preprint arXiv:2006.10963}, 2020.

\bibitem[Pan et~al.(2020)Pan, Gao, Mei, Geng, Li, and Zhao]{pan2020semanticposs}
Yancheng Pan, Biao Gao, Jilin Mei, Sibo Geng, Chengkun Li, and Huijing Zhao.
\newblock Semanticposs: A point cloud dataset with large quantity of dynamic instances.
\newblock In \emph{2020 IEEE Intelligent Vehicles Symposium (IV)}. IEEE, 2020.

\bibitem[Park et~al.(2019)Park, Florence, Straub, Newcombe, and Lovegrove]{park2019deepsdf}
Jeong~Joon Park, Peter Florence, Julian Straub, Richard Newcombe, and Steven Lovegrove.
\newblock Deepsdf: Learning continuous signed distance functions for shape representation.
\newblock In \emph{CVPR}, 2019.

\bibitem[Paszke et~al.(2019)Paszke, Gross, Massa, Lerer, Bradbury, Chanan, Killeen, Lin, Gimelshein, Antiga, et~al.]{paszke2019pytorch}
Adam Paszke, Sam Gross, Francisco Massa, Adam Lerer, James Bradbury, Gregory Chanan, Trevor Killeen, Zeming Lin, Natalia Gimelshein, Luca Antiga, et~al.
\newblock Pytorch: An imperative style, high-performance deep learning library.
\newblock \emph{NeurIPS}, 2019.

\bibitem[Peng et~al.(2021)Peng, Lei, Li, Zhang, and Guo]{peng2021sparse}
Duo Peng, Yinjie Lei, Wen Li, Pingping Zhang, and Yulan Guo.
\newblock Sparse-to-dense feature matching: Intra and inter domain cross-modal learning in domain adaptation for 3d semantic segmentation.
\newblock In \emph{ICCV}, 2021.

\bibitem[Peng et~al.(2020)Peng, Niemeyer, Mescheder, Pollefeys, and Geiger]{convonet}
Songyou Peng, Michael Niemeyer, Lars Mescheder, Marc Pollefeys, and Andreas Geiger.
\newblock Convolutional occupancy networks.
\newblock In \emph{ECCV}, 2020.

\bibitem[Peng et~al.(2023)Peng, Zhu, and Ma]{peng2023cl3d}
Xidong Peng, Xinge Zhu, and Yuexin Ma.
\newblock Cl3d: Unsupervised domain adaptation for cross-lidar 3d detection.
\newblock In \emph{AAAI}, 2023.

\bibitem[Qin et~al.(2019)Qin, You, Wang, Kuo, and Fu]{qin2019pointdan}
Can Qin, Haoxuan You, Lichen Wang, C-C~Jay Kuo, and Yun Fu.
\newblock Pointdan: A multi-scale 3d domain adaption network for point cloud representation.
\newblock In \emph{NeurIPS}, 2019.

\bibitem[Rist et~al.(2019)Rist, Enzweiler, and Gavrila]{rist2019cross}
Christoph~B Rist, Markus Enzweiler, and Dariu~M Gavrila.
\newblock Cross-sensor deep domain adaptation for lidar detection and segmentation.
\newblock In \emph{IV}, 2019.

\bibitem[Rist et~al.(2020)Rist, Schmidt, Enzweiler, and Gavrila]{rist2020scssnet}
Christoph~B. Rist, David Schmidt, Markus Enzweiler, and Dariu~M. Gavrila.
\newblock Scssnet: Learning spatially-conditioned scene segmentation on lidar point clouds.
\newblock In \emph{IV}, 2020.

\bibitem[Rist et~al.(2021)Rist, Emmerichs, Enzweiler, and Gavrila]{rist2021semantic}
Christoph~B Rist, David Emmerichs, Markus Enzweiler, and Dariu~M Gavrila.
\newblock Semantic scene completion using local deep implicit functions on lidar data.
\newblock \emph{IEEE T-PAMI}, 2021.

\bibitem[Rochan et~al.(2022)Rochan, Aich, Corral-Soto, Nabatchian, and Liu]{rochan2022unsupervised}
Mrigank Rochan, Shubhra Aich, Eduardo~R Corral-Soto, Amir Nabatchian, and Bingbing Liu.
\newblock Unsupervised domain adaptation in lidar semantic segmentation with self-supervision and gated adapters.
\newblock In \emph{ICRA}, 2022.

\bibitem[Ryu et~al.(2023)Ryu, Hwang, and Park]{Ryu_2023_CVPR}
Kwonyoung Ryu, Soonmin Hwang, and Jaesik Park.
\newblock Instant domain augmentation for lidar semantic segmentation.
\newblock In \emph{CVPR}, 2023.

\bibitem[Saito et~al.(2017)Saito, Ushiku, and Harada]{saito2017asymmetric}
Kuniaki Saito, Yoshitaka Ushiku, and Tatsuya Harada.
\newblock Asymmetric tri-training for unsupervised domain adaptation.
\newblock In \emph{ICML}, 2017.

\bibitem[Saltori et~al.(2020)Saltori, Lathuili{\'e}re, Sebe, Ricci, and Galasso]{saltori2020sf}
Cristiano Saltori, St{\'e}phane Lathuili{\'e}re, Nicu Sebe, Elisa Ricci, and Fabio Galasso.
\newblock Sf-uda 3d: Source-free unsupervised domain adaptation for lidar-based 3d object detection.
\newblock In \emph{3DV}, 2020.

\bibitem[Saltori et~al.(2022)Saltori, Galasso, Fiameni, Sebe, Ricci, and Poiesi]{saltori2022cosmix}
Cristiano Saltori, Fabio Galasso, Giuseppe Fiameni, Nicu Sebe, Elisa Ricci, and Fabio Poiesi.
\newblock Cosmix: Compositional semantic mix for domain adaptation in 3d lidar segmentation.
\newblock In \emph{ECCV}, 2022.

\bibitem[Saltori et~al.(2023)Saltori, Ošep, Ricci, and Leal-Taixé]{saltori2023walking}
Cristiano Saltori, Aljoša Ošep, Elisa Ricci, and Laura Leal-Taixé.
\newblock Walking your lidog: A journey through multiple domains for lidar semantic segmentation.
\newblock In \emph{ICCV}, 2023.

\bibitem[Sanchez et~al.(2023)Sanchez, Deschaud, and Goulette]{sanchez2023domain}
Jules Sanchez, Jean-Emmanuel Deschaud, and Fran{\c{c}}ois Goulette.
\newblock Domain generalization of 3d semantic segmentation in autonomous driving.
\newblock In \emph{ICCV}, 2023.

\bibitem[Schneider et~al.(2020)Schneider, Rusak, Eck, Bringmann, Brendel, and Bethge]{schneider2020improving}
Steffen Schneider, Evgenia Rusak, Luisa Eck, Oliver Bringmann, Wieland Brendel, and Matthias Bethge.
\newblock Improving robustness against common corruptions by covariate shift adaptation.
\newblock In \emph{NeurIPS}, 2020.

\bibitem[Shen et~al.(2022)Shen, Yang, Yan, Wang, Zheng, and Guibas]{shen2022domain}
Yuefan Shen, Yanchao Yang, Mi Yan, He Wang, Youyi Zheng, and Leonidas~J Guibas.
\newblock Domain adaptation on point clouds via geometry-aware implicits.
\newblock In \emph{CVPR}, 2022.

\bibitem[Sulzer et~al.(2022)Sulzer, Landrieu, Boulch, Marlet, and Vallet]{sulzer2022deep}
Raphael Sulzer, Loic Landrieu, Alexandre Boulch, Renaud Marlet, and Bruno Vallet.
\newblock Deep surface reconstruction from point clouds with visibility information.
\newblock In \emph{ICPR}, 2022.

\bibitem[Sun and Saenko(2016)]{sun2016deep}
Baochen Sun and Kate Saenko.
\newblock Deep coral: Correlation alignment for deep domain adaptation.
\newblock In \emph{ECCV}, 2016.

\bibitem[Tang et~al.(2020)Tang, Liu, Zhao, Lin, Lin, Wang, and Han]{tang2020searching}
Haotian Tang, Zhijian Liu, Shengyu Zhao, Yujun Lin, Ji Lin, Hanrui Wang, and Song Han.
\newblock {Searching Efficient 3D Architectures with Sparse Point-Voxel Convolution}.
\newblock In \emph{ECCV}, 2020.

\bibitem[Tang et~al.(2022)Tang, Liu, Li, Lin, and Han]{tang2022torchsparse}
Haotian Tang, Zhijian Liu, Xiuyu Li, Yujun Lin, and Song Han.
\newblock {TorchSparse: Efficient Point Cloud Inference Engine}.
\newblock In \emph{MLSys}, 2022.

\bibitem[Tarvainen and Valpola(2017)]{tarvainen2017mean}
Antti Tarvainen and Harri Valpola.
\newblock Mean teachers are better role models: Weight-averaged consistency targets improve semi-supervised deep learning results.
\newblock In \emph{NeurIPS}, 2017.

\bibitem[Tranheden et~al.(2021)Tranheden, Olsson, Pinto, and Svensson]{tranheden2021dacs}
Wilhelm Tranheden, Viktor Olsson, Juliano Pinto, and Lennart Svensson.
\newblock Dacs: Domain adaptation via cross-domain mixed sampling.
\newblock In \emph{WACV}, 2021.

\bibitem[Triess et~al.(2021)Triess, Dreissig, Rist, and Z{\"o}llner]{triess2021survey}
Larissa~T Triess, Mariella Dreissig, Christoph~B Rist, and J~Marius Z{\"o}llner.
\newblock A survey on deep domain adaptation for lidar perception.
\newblock In \emph{IV Workshops}, 2021.

\bibitem[Tsai et~al.(2022)Tsai, Berrio, Shan, Worrall, and Nebot]{tsai2022see}
Darren Tsai, Julie~Stephany Berrio, Mao Shan, Stewart Worrall, and Eduardo Nebot.
\newblock See eye to eye: A lidar-agnostic 3d detection framework for unsupervised multi-target domain adaptation.
\newblock \emph{RA-L}, 2022.

\bibitem[Tzeng et~al.(2017)Tzeng, Hoffman, Saenko, and Darrell]{tzeng2017adversarial}
Eric Tzeng, Judy Hoffman, Kate Saenko, and Trevor Darrell.
\newblock Adversarial discriminative domain adaptation.
\newblock In \emph{CVPR}, 2017.

\bibitem[Van~der Maaten and Hinton(2008)]{van2008visualizing}
Laurens Van~der Maaten and Geoffrey Hinton.
\newblock Visualizing data using t-sne.
\newblock \emph{Journal of machine learning research}, 9\penalty0 (11), 2008.

\bibitem[Vu et~al.(2019)Vu, Jain, Bucher, Cord, and P{\'e}rez]{vu2019advent}
Tuan-Hung Vu, Himalaya Jain, Maxime Bucher, Matthieu Cord, and Patrick P{\'e}rez.
\newblock Advent: Adversarial entropy minimization for domain adaptation in semantic segmentation.
\newblock In \emph{CVPR}, 2019.

\bibitem[Wang et~al.(2021)Wang, Shelhamer, Liu, Olshausen, and Darrell]{wangtent}
Dequan Wang, Evan Shelhamer, Shaoteng Liu, Bruno Olshausen, and Trevor Darrell.
\newblock Tent: Fully test-time adaptation by entropy minimization.
\newblock In \emph{ICLR}, 2021.

\bibitem[Wang et~al.(2017)Wang, Li, Dai, and Van~Gool]{wang2017deep}
Yifei Wang, Wen Li, Dengxin Dai, and Luc Van~Gool.
\newblock Deep domain adaptation by geodesic distance minimization.
\newblock In \emph{CVPRW}, 2017.

\bibitem[Wang et~al.(2020)Wang, Chen, You, Li, Hariharan, Campbell, Weinberger, and Chao]{wang2020train}
Yan Wang, Xiangyu Chen, Yurong You, Li~Erran Li, Bharath Hariharan, Mark Campbell, Kilian~Q Weinberger, and Wei-Lun Chao.
\newblock Train in germany, test in the usa: Making 3d object detectors generalize.
\newblock In \emph{CVPR}, 2020.

\bibitem[Wei et~al.(2022)Wei, Wei, Rao, Li, Zhou, and Lu]{wei2022lidar}
Yi Wei, Zibu Wei, Yongming Rao, Jiaxin Li, Jie Zhou, and Jiwen Lu.
\newblock Lidar distillation: Bridging the beam-induced domain gap for 3d object detection.
\newblock In \emph{ECCV}, 2022.

\bibitem[Wilson and Cook(2020)]{wilson2020survey}
Garrett Wilson and Diane~J Cook.
\newblock A survey of unsupervised deep domain adaptation.
\newblock \emph{ACM TIST}, 2020.

\bibitem[Xiao et~al.(2022)Xiao, Huang, Guan, Zhan, and Lu]{xiao2022transfer}
Aoran Xiao, Jiaxing Huang, Dayan Guan, Fangneng Zhan, and Shijian Lu.
\newblock Transfer learning from synthetic to real lidar point cloud for semantic segmentation.
\newblock In \emph{AAAI}, 2022.

\bibitem[Xiao et~al.(2023)Xiao, Huang, Xuan, Ren, Liu, Guan, El~Saddik, Lu, and Xing]{Xiao_2023_CVPR}
Aoran Xiao, Jiaxing Huang, Weihao Xuan, Ruijie Ren, Kangcheng Liu, Dayan Guan, Abdulmotaleb El~Saddik, Shijian Lu, and Eric~P. Xing.
\newblock 3d semantic segmentation in the wild: Learning generalized models for adverse-condition point clouds.
\newblock In \emph{CVPR}, 2023.

\bibitem[Xu et~al.(2021)Xu, Zhou, Wang, Qi, and Anguelov]{xu2021spg}
Qiangeng Xu, Yin Zhou, Weiyue Wang, Charles~R Qi, and Dragomir Anguelov.
\newblock Spg: Unsupervised domain adaptation for 3d object detection via semantic point generation.
\newblock In \emph{ICCV}, 2021.

\bibitem[Yang et~al.(2021)Yang, Shi, Wang, Li, and Qi]{yang2021st3d}
Jihan Yang, Shaoshuai Shi, Zhe Wang, Hongsheng Li, and Xiaojuan Qi.
\newblock St3d: Self-training for unsupervised domain adaptation on 3d object detection.
\newblock In \emph{CVPR}, 2021.

\bibitem[Yang et~al.(2022)Yang, Shi, Wang, Li, and Qi]{yang2021st3d++}
Jihan Yang, Shaoshuai Shi, Zhe Wang, Hongsheng Li, and Xiaojuan Qi.
\newblock St3d++: Denoised self-training for unsupervised domain adaptation on 3d object detection.
\newblock \emph{T-PAMI}, 2022.

\bibitem[Yi et~al.(2021)Yi, Gong, and Funkhouser]{yi2021complete}
Li Yi, Boqing Gong, and Thomas Funkhouser.
\newblock {Complete \& Label}: A domain adaptation approach to semantic segmentation of lidar point clouds.
\newblock In \emph{CVPR}, 2021.

\bibitem[You et~al.(2022)You, Diaz-Ruiz, Wang, Chao, Hariharan, Campbell, and Weinbergert]{you2022exploiting}
Yurong You, Carlos~Andres Diaz-Ruiz, Yan Wang, Wei-Lun Chao, Bharath Hariharan, Mark Campbell, and Kilian~Q Weinbergert.
\newblock Exploiting playbacks in unsupervised domain adaptation for 3d object detection in self-driving cars.
\newblock In \emph{ICRA}, 2022.

\bibitem[Zhang et~al.(2021)Zhang, Li, and Xu]{zhang2021srdan}
Weichen Zhang, Wen Li, and Dong Xu.
\newblock Srdan: Scale-aware and range-aware domain adaptation network for cross-dataset 3d object detection.
\newblock In \emph{CVPR}, 2021.

\bibitem[Zhao et~al.(2021)Zhao, Wang, Li, Wu, Gao, Xu, Darrell, and Keutzer]{zhao2021epointda}
Sicheng Zhao, Yezhen Wang, Bo Li, Bichen Wu, Yang Gao, Pengfei Xu, Trevor Darrell, and Kurt Keutzer.
\newblock epointda: An end-to-end simulation-to-real domain adaptation framework for lidar point cloud segmentation.
\newblock In \emph{AAAI}, 2021.

\bibitem[Zhou et~al.(2018)Zhou, Karpur, Gan, Luo, and Huang]{zhou2018unsupervised}
Xingyi Zhou, Arjun Karpur, Chuang Gan, Linjie Luo, and Qixing Huang.
\newblock Unsupervised domain adaptation for 3d keypoint estimation via view consistency.
\newblock In \emph{ECCV}, 2018.

\bibitem[Zou et~al.(2021)Zou, Tang, Chen, and Jia]{zou2021geometry}
Longkun Zou, Hui Tang, Ke Chen, and Kui Jia.
\newblock Geometry-aware self-training for unsupervised domain adaptation on object point clouds.
\newblock In \emph{CVPR}, 2021.

\bibitem[Zou et~al.(2018)Zou, Yu, Kumar, and Wang]{zou2018unsupervised}
Yang Zou, Zhiding Yu, BVK Kumar, and Jinsong Wang.
\newblock Unsupervised domain adaptation for semantic segmentation via class-balanced self-training.
\newblock In \emph{ECCV}, 2018.

\bibitem[Zou et~al.(2019)Zou, Yu, Liu, Kumar, and Wang]{zou2019confidence}
Yang Zou, Zhiding Yu, Xiaofeng Liu, BVK Kumar, and Jinsong Wang.
\newblock Confidence regularized self-training.
\newblock In \emph{ICCV}, 2019.

\end{thebibliography}
}

\clearpage
\secondarytitle{\titletext\\
\textit{--- Supplementary Material ---}}
\normalsize
\appendix
\setcounter{table}{4}
\setcounter{figure}{4}

\section*{Appendix}
We provide in this document: more ablations (Section~\ref{sup:sec:more_ablations}), a discussion regarding the UDA validators we used (Section~\ref{sup:sec:clearification_validators}), more implementation details (Section~\ref{sup:sec:implementation_details}),
class-wise results (Section~\ref{sup:sec:classwise_results}) and frequency-weighted mIoU results (Section~\ref{sup:sec:fwiou_results}), as well as further
qualitative analyses (Section~\ref{sup:sec:qualitative_results}). 

\section{More ablations}
\label{sup:sec:more_ablations}
\subsection{Ablation per distance}
In Table~\ref{tab:experiments_comparison_ns_sk_distance} we report the detailed results for different distances, that are discussed in Section~4.7 in the main paper.

\begin{table}[h!]
    \small
    \setlength{\tabcolsep}{2.5pt}
    \centering
    \begin{tabular}{ll|ccccc}
        \toprule
        &Distance (m) & 0$\,\shortto$7.5 & 7.5$\,\shortto$15 & 15$\,\shortto$30 & 30$\,\shortto$50 \\
        &Proportion of points & 45.3\% & 34.4\% & 15.7\% & 4.7\%  \\
        \midrule
        \rowcolor{black!10}
        \cellcolor{white}&Supervised on \skns & 82.3 & 69.8 &63.7 & 51.1\\
        & Source-only  & 33.8	& 44.8 & \textbf{47.1} &	32.1 \\
        &SALUDA & \textbf{47.9} & \textbf{49.2} & 47.0 & \textbf{33.3} \\
        \multirow{-4}{*}{\rotatebox[origin=c]{90}{NS$\shortto$\skns}}&\qquad \footnotesize gain wrt src-only    & \footnotesize\color{ForestGreen}{+24.1} &  \footnotesize\color{ForestGreen}{+4.4} & \footnotesize\color{BrickRed}{-0.1} & \footnotesize\color{ForestGreen}{+1.2}\\
        \midrule
        \rowcolor{black!10}
        \cellcolor{white}&Supervised on \sksyn &  58.6 & 55.6 & 	54.0 &	41.6 \\
         &Source-only  & 20.5	 &22.4 &	25.1 &	20.6 \\
        &SALUDA  & \textbf{31.4} & \textbf{31.5} & \textbf{31.5} & \textbf{24.7}\\
        \multirow{-4}{*}{\rotatebox[origin=c]{90}{\synth$\shortto$\sksyn}}&\qquad \footnotesize gain wrt src-only    & \footnotesize\color{ForestGreen}{+10.9} &  \footnotesize\color{ForestGreen}{+9.1} & \footnotesize\color{ForestGreen}{+6.4} & \footnotesize\color{ForestGreen}{+4.1} \\
        \bottomrule
    \end{tabular}
    \caption{\textbf{Per-distance performance}. mIoU\% averaged over 3 runs.}
\label{tab:experiments_comparison_ns_sk_distance}
\end{table}

\subsection{Ablation different reconstruction head}
 ALSO and POCO heads are compared in their best setting in Table~4 in the main paper. It confirms that the ALSO head outperforms POCO's (+4.6\,mIoU\,pp) on segmentation~\cite{ALSO}\,(supp.). In fact, the POCO head was designed for local geometric accuracy, ALSO’s for object-level semantics.

\begin{table}[h]
    \vspace{-3mm}
    \small
    \setlength{\tabcolsep}{2pt}
    \centering
        \begin{tabular}{l|ccc}
            \toprule

             Setting (\%\,mIoU) & \nstosk  & \synthtosk & \synthtoposs \\
            
            \midrule
            
            \method{} w/o ST-POCO head & 44.2 &  27.0 & 35.8  \\
            \method{} w/o ST-ALSO head & \textbf{44.9}  &  \textbf{27.6}  & \textbf{37.2}  \\
            \bottomrule
        \end{tabular}
        \vspace*{0.5mm}
    \caption{\textbf{Comparison of different reconstruction heads.}}
\label{tab:experiments_rec_heads}
\vspace{-4mm}
\end{table}
 
 We can see in Table~\ref{tab:experiments_rec_heads} that the ALSO head is better in various domain adaptation settings. We hypothesize too that its neighborhood of fixed radius makes it more robust to varying point cloud densities.

\section{UDA validators used in this work}
\label{sup:sec:clearification_validators}

Among all validators presented in the \href{https://arxiv.org/pdf/2208.07360v1.pdf}{first version} of~\cite{musgrave2022benchmarking}, we originally selected source validation (SrcVal), target entropy (Entropy) and information maximization (IM).
One of the advantages of these
three
validators, which are ranked high among other validators,
is that they
do not require training; still, they can help to choose hyperparameters bringing favorable target performance.
In the \href{https://arxiv.org/pdf/2208.07360.pdf}{last version} of~\cite{musgrave2022benchmarking}, the IM validator has been removed.
For completeness, we still report results obtained with IM.

\section{Implementation details}
\label{sup:sec:implementation_details}

\subsection{Training details}

\paragraph{\method{} and baselines (besides \cosmix{}).}

For training the baseline methods as well as \method{}, we use a batch size of 4. 
We show during training 300K batches of source data and 300K batches of target data (if used by the method), and alternate between batches of source and target data.

The learning rate is initialized to 10$^{-3}$ and is gradually decayed to $0$ using a cosine annealing scheduler. 
We use AdamW with default parameters as optimizer~\cite{loshchilov2019decoupled}. For data augmentation on point cloud, we perform
random rotation around the $z$-axis (from $-180 ^{\circ}$ to  $180^{\circ}$) and random flipping on the other axes.

The pseudo code of our training procedure is presented in Algorithm~\ref{appendix:alg:training}.

The optimal parameter choice for the different methods, when using target validation for selection, (referred to as ``Oracle'' in Table 3 of the submission), are reported in Table~\ref{tab:appendix_hyperparameter}.

\begin{table}[!h!]
    \small
    \newcommand*\rotext{\multicolumn{1}{R{45}{1em}}}
    \setlength{\tabcolsep}{3.5pt}
    \centering
    \begin{tabular}{l|c|c|c|c|c|c}
    \toprule  
     Method &  \rotext{\method} &\rotext{DUA} &  \rotext{MinEnt} & \rotext{Coral} & \rotext{LogCoral} & \rotext{CoSMix} \\
     Parameter & $\lambda$ & $\omega$, $\zeta$  & $\lambda$  & $\lambda$ &  $\lambda$ & conf. \\
    \midrule
           
    \nstosk{} & $10^{-5}$   & 0.89, 0.0 & 0 & 0 &  0 & 0.9\\
    \nstoposs{} & $10^{-5}$ & 0.54, 0.0 & $10^{-1}$ & $10^{-5}$ & $10^{-2}$ & 0.9\\
    \synthtosk{} (10-cm)  & $10^{-3}$ & 0.89, 0.0 &  0 & $10^{-5}$ & 0 & 0.9\\
    \midrule
    \synthtosk{} (5-cm) & $10^{-3}$ & -- & -- & --& --& 0.9\\
    \synthtoposs &  $10^{-5}$ &  -- &  -- & -- & -- & 0.85 \\

    \bottomrule
    \end{tabular}
    \vspace*{-0.9mm}
    \caption{Optimal hyperparameter choice (referred to as ``Oracle'' in Table 3 of the submission).}
\label{tab:appendix_hyperparameter}
\end{table}

\paragraph{\cosmix.}
We used the official implementation of \cosmix{} and followed the training setting described in \cite{saltori2022cosmix}.
\cosmix\ uses a teacher-student architecture during training.
In all our experiments, the student outperformed the exponential moving average (EMA) teacher. We therefore only reported the student results.

\subsection{Architecture}

For fair comparisons, we trained all models using a voxel size of 10 cm and the same Minkowski U-Net architecture, where the feature dimensions at each layer are 32, 32, 64, 128, 256, 256, 128, 96, 96, followed by a single linear layer with dimension 128.

All methods are implemented using PyTorch~\cite{paszke2019pytorch}.
The sparse Minkowski U-Net is implemented using MinkowskiEngine~\cite{choy20194d} for \cosmix, and TorchSparse~\cite{tang2022torchsparse,tang2020searching} for \method{} and the other baselines. 
TorchSparse is a fork from MinkowskiEngine, differing only in the indexation library.

\subsection{Hardware configuration}

In \method{} and \minent{}, as source and target batches are alternated during training, we manage to train those methods on a single NVIDIA GeForce RTX 2080 Ti 11GB.
\cosmix\ is trained with two NVIDIA GeForce RTX 2080 Ti to ensure the same batch size recommended in the original paper.
The loss computation of Coral and LogCoral requires having both source and target data in one iteration batch; to train with these methods, we use a bigger GPU with more memory, i.e., a NVIDIA Tesla V100 32GB.

\paragraph{Training time.}
The training time for \nstosk{} on one NVIDIA RTX 2080 Ti is about 5 days for SALUDA w/o ST, 4 for LogCoral, 3 for MinEnt and Coral, 2 for CoSMix and 2 for ST. As SALUDA does not reconstruct any surface explicitly, contrary to~\cite{yi2021complete}, it is as fast as the baseline methods at inference time.

\subsection{Datasets.}

For nuScenes, SemanticKITTI and SemanticPOSS, we use the official train/val splits.
For SynLidar, we use the recommended sub-split from \cite{xiao2022transfer}.
For \nstoposs{} we use the 6 overlapping classes of \emph{Person}, \emph{Bike}, \emph{Car}, \emph{Ground}, \emph{Vegetation} and  \emph{Man-made} like in \cite{sanchez2023domain}.

\RestyleAlgo{ruled}
\RestyleAlgo{linesnumbered}
\begin{algorithm}
    \small
    \caption{Training of \method{}}
    \label{appendix:alg:training}
    \SetKwInOut{Input}{Input}
    \SetKwInOut{Output}{output}
    \SetKwInOut{Parameter}{Param}
    \SetKwFunction{SQ}{sample-query}
    \SetKwFunction{CE}{CE}
    \SetKwFunction{BP}{backpropagate}
    \SetKwFunction{ST}{ST}

    \renewcommand{\CommentSty}[1]{\textnormal{\ttfamily\color{green!50!black}#1}}
    \DontPrintSemicolon
    \Input{~Source data: $\mathcal{D}^{s}$\\
           ~Target data: $\mathcal{D}^{t}$\\
           ~Hyperparameter: $\lambda>0$ 
        } 
    \Parameter{ Number of iterations: $\mathit{number}_{\mathit{iter}}$
    }
    \tcp{Init.\ backbone}
     $\phi(\cdot) \gets init$\;
     \tcp{Init.\ semantic segmentation head}
    $\mathrm{cls}(\cdot) \gets init $\;
    \tcp{Init.\ surface head}
    $\mathrm{surf}(\cdot) \gets init$\; 
    \While{$i \leq \mathit{number}_{\mathit{iter}}$}{
        \eIf{$i$ is odd}{
        \tcp{Source training}
            $P^{s}_{i}, Y^{s}_{i} \gets \mathcal{D}^{s}$  \\
            $Z^s_i \gets \phi(P^s_i)$ \\
            $\hat{Y}^s_i \gets \mathrm{cls}(Z^s_i)$ \\
            \tcp{occ is the variable indicating the occupancy of a query point}
            $\widetilde{Z^s_{i}}, occ^{s}_i \gets \SQ(P^s_i, Z^s_i)$ \\
            $\hat{occ}^s_i \gets \mathrm{surf}(\widetilde{Z}^s_i)$ \\
            $L \gets \CE(Y_{\mathit{src}_i}, \hat{Y}) + \lambda \CE(occ^s_i, \hat{occ}^s_i)$ \\
            \BP($L$)\ \\
            }
        {
            \tcp{Target training}
            $P^t_{i} \gets \mathcal{D}^t$\\
            $Z^t_i \gets \phi(P^t_i)$ \\
            $\widetilde{Z^t_{i}}, occ^{t}_i \gets \SQ(P^t_i, Z^t_i)$ \\
            $\hat{occ}^t_i \gets \mathrm{surf}(\widetilde{Z}^t_i)$ \\
            $L \gets \lambda \CE(occ^t_i, \hat{occ}^t_i)$ \\
            \BP($L$)\\
            
        }
        
    }
   
    \tcp{Init.\ Student model}
    $\phi_{s}(\cdot), \mathrm{cls}_{s}(\cdot) \gets \phi(\cdot), \mathrm{cls}(\cdot)$\;
    $\mathrm{model}_{s}(\cdot) \gets \mathrm{cls}_{s}(\phi_{s}(\cdot))$\;
    \tcp{Init.\ Teacher model}
    $\phi_{t}(\cdot), \mathrm{cls}_{t}(\cdot) \gets \phi(\cdot), \mathrm{cls}(\cdot)$\;
    $\mathrm{model}_{t}(\cdot) \gets \mathrm{cls}_{t}(\phi_{t}(\cdot))$\;
    
    \tcp{Self-training}
    $\mathrm{model}_{final}(\cdot)  \gets \ST(\mathrm{model}_{s}(\cdot),\mathrm{model}_{t}(\cdot), \mathcal{D}^{s}, \mathcal{D}^{t})$\;
    
\end{algorithm}

\section{Detailed quantitative results}

\subsection{Classwise results}

\label{sup:sec:classwise_results}
\begin{table*}[t!]
\small
\centering
\newcommand*\rotext{\multicolumn{1}{R{45}{1em}}}
\setlength{\tabcolsep}{3.5pt}
\begin{tabular}{lr|cccccccccc|c@{\hskip3pt}l}
\toprule
 \rlap{\raisebox{12mm}{\DAsetting{\ns}{\skns}}}%
 \rlap{\raisebox{6mm}{~~~(\%\,IoU)}}%
 \rlap{\raisebox{2mm}{~~~10-cm voxel size}}%
 && \rotext{Car} &	\rotext{Bicycle}	& \rotext{Motorcycle} &\rotext{Truck} &	\rotext{Other vehicle}	 & \rotext{Pedestrian}	 & \rotext{Driveable surf.} &	\rotext{Sidewalk} &	\rotext{Terrain} &\rotext{Vegetation} & \rotext{\%\,mIoU}\\ \midrule
\rowcolor{black!10}
C\&L (20\,cm)~\: & \cite{yi2021complete} &- &- &- &- &- &- &- &- &- &- & 33.7 & \scaleto{\text{(paper)}}{5pt}\\ 
\midrule
Source-only  && 73.7 &8.0&	17.8&	12.0&	\second{7.4}&	\second{49.4}&	50.2&	27.0&	31.6&	82.1 & \mioucell \perf{35.9} & \std{3.2} \\ 
\midrule

AdaBN & \cite{LI2018109} & 84.1 &	\best{16.5} &	24.0 &	7.6 &	3.5 &	19.2 &	{76.0} &	35.6 &	51.0 &	83.1 & \mioucell\perf{40.1} & \std{0.4} \\
DUA & \cite{mirza2022dua} & 85.6 &	13.3	& 28.3 &	{13.3} &	6.0 &	37.3 &	75.9 &	33.8 &	48.0& {87.4} & \mioucell\perf{42.9} & \std{0.7}\\
Mixed BN & \llap{(ours)} & 87.4 &	\second{15.3} &	\second{30.6} &	10.1 &	5.3	& {38.6}	& 75.1 & 40.1 & {44.2} & 86.1 & {\perf{43.3}} & \std{0.6}\\

\back{MinEnt$^\dagger$} & \cite{vu2019advent} & \back{87.4} &	\back{15.3} &	\back{30.6} &	\back{10.1} &	\back{5.3}	& \back{38.6}	& \back{75.1} & \back{40.1} & \back{44.2} & \back{86.1} & \mioucell\back{\perf{43.3}} & \back{\std{0.6}}\\

\back{Coral$^\dagger$} & \cite{sun2016deep} & \back{87.4} &	\back{15.3} &	\back{30.6} &	\back{10.1} &	\back{5.3}	& \back{38.6}	& \back{75.1} & \back{40.1} & \back{44.2} & \back{86.1} & \mioucell\back{\perf{43.3}}  & \back{\std{0.6}}\\

\back{LogCoral$^\dagger$} & \cite{wang2017deep} & \back{87.4} &	\back{15.3} &	\back{30.6} &	\back{10.1} &	\back{5.3}	& \back{38.6}	& \back{75.1} & \back{40.1} & \back{44.2} & \back{86.1} & \mioucell\back{\perf{43.3}}  & \back{\std{0.6}}\\

ST & \cite{zou2019confidence} &78.4	&	9.5	&	20.2	&	11.9	&	\best{7.5}		& {39.9}	&	54.2	&	33.4	&	34.0	&	84.0 & 37.3& \std{2.9}  \\
CoSMix & \cite{saltori2022cosmix} &77.1	&	10.4& 20.0	&\second{15.2}	&	6.6	&	\best{51.0}&	52.1&	31.8&	34.5&84.8 & 38.3& \std{2.8}\\  

\midrule
\method{} w/o ST & \llap{(ours)} & \second{88.8} &	14.1 &	\best{33.0} & 12.8 &	5.5 &	37.6	 & \second{76.4} & \second{41.5} &	\second{51.8} & \second{87.6} & \second{\perf{44.9}}& \std{0.2}\\ 

SALUDA & \llap{(ours)} & \best{89.8} &		13.2&	26.2&	\best{15.3}	&	7.0	&	37.6&	\best{79.0}&	\best{50.4}&	\best{55.0}&	\best{88.3}& \best{46.2} & \std{0.6}\\
\midrule
\rowcolor{red!10}
\emph{Upper bound} & &  \\
\rowcolor{red!10}
Supervised & & 	95.3 & 	9.1 &	38.5 &	70.8 &	40.0 &	64.8 &	91.3 &	78.9 &	73.1 &	91.3& 65.3 & \std{0.6}\\

\bottomrule

\end{tabular}
\vspace*{-2mm}
\caption{\textbf{Classwise results \DAsetting{\ns}{\skns}.} Average over 3 runs. Color:\colorbox{blue!20}{\color{black}Best}, \colorbox{blue!10}{\color{black}Second} $^\dagger$: best results with no regularization = Mixed BN. 
}
\label{tab:experiments_per_class_ns_sk}
\end{table*}
\begin{table*}[!ht]
\small
\centering
\newcommand*\rotext{\multicolumn{1}{R{45}{1em}}}
\setlength{\tabcolsep}{2.2pt}
\begin{tabular}{lr|ccccccccccccccccccc|c@{\hskip2pt}l}
\toprule
 \rlap{\raisebox{12mm}{\synthtosk}}%
 \rlap{\raisebox{6mm}{~~~(\%\,IoU)}}%
 \rlap{\raisebox{2mm}{~~~10-cm voxel size}}%
 && \rotext{Car} &	\rotext{Bicycle} & \rotext{Motorcycle} & \rotext{Truck} &	\rotext{Other vehicle} &	\rotext{Pedestrian} &	\rotext{Bicyclist} &	\rotext{Motorcyclist} & \rotext{Road} & \rotext{Parking} & \rotext{Sidewalk} &	\rotext{Other ground} &	\rotext{Building} &	\rotext{Fence} & \rotext{Vegetation} &	\rotext{Trunk} & \rotext{Terrain} &	\rotext{Pole} &\rotext{Traffic sign} & \rotext{\%\,mIoU}\\ \midrule

\rowcolor{black!10}

\midrule
 
Source-only && 34.8	& 5.9 &	{14.7} & 2.0	 & 1.4	 & 18.5 &	{48.9} &	{3.1} &	26.1 &	6.3	 & 33.6	& 0.0 &	33.9 &	17.7 &	61.2 &	24.0 &	\second{46.4} &	27.0 &	4.9 & 21.6 & \std{0.2}\\ \midrule

 AdaBN & \cite{LI2018109} & 49.5 &	7.5 &	9.7 &	\second{3.9} &	\second{6.5} &	15.4 &	37.6 &	0.7	& \best{57.3} &	6.7 &	34.4 &	\best{0.4}	& {58.9} &	24.5 &	64.7 &	26.4 &	38.8 &	30.3 &	13.6 & \mioucell\perf{25.6} & \std{0.2}\\
 
DUA & \cite{mirza2022dua} & 44.8 & 7.5 &	11.8 &	2.7 &	5.0 &	19.0 &	41.6 &	1.2 &	51.0 &	7.2 &	35.7 &	{0.2} &	56.4 &	29.4 &	66.4 &	29.2 &	41.3 &	35.6 & 15.0 & \mioucell\perf{26.4} & \std{0.4}  \\
 
 Mixed BN & \llap{(ours)} & 50.4	& \best{9.0}	& 13.5	& 2.1 &	3.6 &	19.9 &	41.8 &	2.7 &	52.8 &	6.7	 & 35.7 &	0.1	& 57.5 &	25.4 &	{69.0} &	28.8 &	41.1	&37.6	& {15.3} & \mioucell \perf{27.0} & \std{0.6}\\

 \back{MinEnt$^\dagger$} & \cite{vu2019advent} & \back{50.4}	& \back{9.0}	& \back{13.5} & \back{2.1} & \back{3.6} & \back{19.9} &	\back{41.8} & \back{2.7} & \back{52.8} & \back{6.7}	 & \back{35.7} & \back{0.1} & \back{57.5} &	\back{25.4} &	\back{69.0} & \back{28.8} &	\back{41.1}	& \back{37.6}	& \back{15.3} & \mioucell \back{\perf{27.0}} & \back{\std{0.6}} \\ 
 
Coral & \cite{sun2016deep} & 47.1 &	7.6 &	13.7 &	 1.9	& 3.8	& {21.0} &	{45.1}&	1.6 &	49.1 &	7.5 &	36.1 &	0.1 &	58.7&	 \best{32.4} &	67.9 &	30.7 &	43.2 &	36.8 &	13.3& \mioucell \perf{27.3} & \std{0.3}\\  

 \back{LogCoral$^\dagger$} & \cite{wang2017deep} & \back{50.4}	& \back{9.0}	& \back{13.5} & \back{2.1} & \back{3.6} & \back{19.9} &	\back{41.8} & \back{2.7} & \back{52.8} & \back{6.7}	 & \back{35.7} & \back{0.1} & \back{57.5} &	\back{25.4} &	\back{69.0} & \back{28.8} &	\back{41.1}	& \back{37.6}	& \back{15.3} &  \mioucell \back{\perf{27.0}} & \back{\std{0.6}} \\ 

 ST & \cite{zou2019confidence} & 46.0&7.2 &	\best{16.6} &	2.3	 &	3.9	 &	\second{22.8} &	\best{54.2} &	\second{3.2} &32.5	&8.5&37.7&0.3&41.7&26.9&64.7&30.4& \best{46.7}&	45.0 &	16.7 & 26.7 & \std{0.4}\\

 CoSMix & \cite{saltori2022cosmix}& \second{63.9} &	5.6 &	11.4 &	\best{5.7} &	\best{7.9}	 & 20.0 &	40.3 &	\best{3.8}	& \second{56.4} &	\best{13.2} &	{37.9} &	0.1 &	42.6 &	\second{29.5} &	66.9 &	27.9 &	29.6 &	\second{46.0} &	\second{22.5} & \best{\perf{28.0}} & \std{{1.4}}\\
 
 \midrule 
 \method{}  w/o ST & \llap{(ours)} & {52.1} &	\second{8.6} &	\second{15.0} &	1.9 &	3.8 & {21.4} &	43.4 &	1.7 &	53.7 &	{7.8} &	{38.0} &	0.1 &	\second{59.2} &	22.3 &	\second{69.1} &	\second{30.8} &	{45.1} &	{38.0} &	12.6 & {\perf{27.6}} & \std{0.5}\\

 \method{} & (ours)&\best{65.4}& 7.5 &13.6&3.2 &5.9&\best{23.9}&43.7&	1.7&52.9&\second{11.6}&\best{39.8}&\second{0.3}&\best{67.8}&28.2&\best{74.2}&\best{37.6}&43.6&\best{47.5}&\best{22.7}&\best{31.2} &\std{0.2} \\
 \midrule 
 \rowcolor{red!10}
\emph{Upper bound} & &  \\
\rowcolor{red!10}
 Supervised &  & 94.5&	11.5&	32.2&	81.1&	33.0&	59.5&	82.7&	0.0&	92.4&	49.7&	78.2&	1.3&	90.9&	57.2&	87.1&	63.5&	72.4&	61.8&	44.9&	57.5  & \std{0.9}\\
 \bottomrule

\end{tabular}
\vspace*{-2mm}
\caption{\textbf{Classwise results \synthtosk.} Average over 3 runs. Color:\colorbox{blue!20}{\color{black}Best}, \colorbox{blue!10}{\color{black}Second}. $^\dagger$: best results with no regularization = Mixed BN. }
\label{tab:experiments_per_class_syn_sk}
\vspace*{-3mm}
\end{table*}
\begin{table*}[!ht]
\small
\centering
\newcommand*\rotext{\multicolumn{1}{R{45}{1em}}}
\setlength{\tabcolsep}{2.2pt}
\begin{tabular}{lr|cccccc|c@{\hskip2pt}l}
\toprule
 \rlap{\raisebox{12mm}{\nstoposs}}%
 \rlap{\raisebox{6mm}{~~~(\%\,IoU)}}%
 \rlap{\raisebox{2mm}{~~~10-cm voxel size}}%
 && \rotext{Person} &	\rotext{Bike} & \rotext{Car} & \rotext{Ground} &	\rotext{Vegetation} &	\rotext{Manmade}& \rotext{\%\,mIoU}\\ \midrule

\midrule
 \rowcolor{black!10}
Source-only && 59.7	&	13.3&65.4 &	80.1&79.9&76.3 &62.5 & \std{0.2}\\ \midrule
 AdaBN & \cite{LI2018109} & 59.7	&	12.7	&	67.2	&	78.5	&	80.7	&	76.3 & 62.5 & \std{0.0} \\
 
DUA & \cite{mirza2022dua} & 59.6	&	12.8	&	64.6	&	80.0	&	80.0	&	76.5& 62.3 & \std{0.1}\\
 
 Mixed BN &\llap{(ours)}&60.0&12.0&64.9&80.4&80.1&	77.2 & 62.4 & \std{0.1} \\

 MinEnt & \cite{vu2019advent} &\second{60.1}&11.5&66.4&80.7&80.1& 77.1 & 62.6& \std{0.1} \\
 
Coral & \cite{sun2016deep} & 59.5	&	11.1	&	69.4	&	80.1	&	80.5	&77.2 & 63.0 & \std{0.2}\\ 

LogCoral & \cite{wang2017deep} & 59.6	&	11.2	&	67.2 &		78.9 &		80.6	&	77.2 & 62.5 & \std{0.1} \\ 

  ST & \cite{zou2019confidence} & 59.4	&	\second{21.6}	&	\best{71.3}	&	80.6	&	\best{81.6}	&	\second{78.6} & \second{65.5} & \std{0.2} \\
 
 CoSMix & \cite{saltori2022cosmix}& \best{60.3} &		\best{24.1}	&	66.4	&	80.4	&	\second{81.4}	&	78.3 & 65.2 & \std{0.2} \\

 \midrule 
 \method{}  w/o ST & \llap{(ours)} & 59.6	&	12.2	&	67.1&		\second{81.0}&		80.5	&	78.1 & 63.1 & \std{0.1} \\

 \method{} & (ours) & 59.0		& 20.5	&	\second{70.6} &		\best{82.6}	&	\second{81.4}	 &	\best{81.0}	& \best{65.8} & \std{0.3}  \\
 \midrule 
 \rowcolor{red!10}
\emph{Upper bound} & &  \\
\rowcolor{red!10}
Supervised& & 67.2 & 	62.5 & 	70.2 &	80.2 &	82.1 &	81.2 & 73.9 & \std{1.3}\\
 \bottomrule

\end{tabular}
\vspace*{-2mm}
\caption{\textbf{Classwise results \nstoposs{}.} Average over 3 runs. Color:\colorbox{blue!20}{\color{black}Best}, \colorbox{blue!10}{\color{black}second}. }
\label{tab:experiments_per_class_ns_sp}
\end{table*}

\begin{table*}[!ht]
\small
\centering
\newcommand*\rotext{\multicolumn{1}{R{45}{1em}}}
\setlength{\tabcolsep}{2.2pt}
\begin{tabular}{lr|ccccccccccccccccccc|c@{\hskip2pt}l}
\toprule
 \rlap{\raisebox{12mm}{\synthtosk}}%
 \rlap{\raisebox{6mm}{~~~(\%\,IoU)}}%
 \rlap{\raisebox{2mm}{~~~5-cm voxel size}}%
 && \rotext{Car} &	\rotext{Bicycle} & \rotext{Motorcycle} & \rotext{Truck} &	\rotext{Other vehicle} &	\rotext{Pedestrian} &	\rotext{Bicyclist} &	\rotext{Motorcyclist} & \rotext{Road} & \rotext{Parking} & \rotext{Sidewalk} &	\rotext{Other ground} &	\rotext{Building} &	\rotext{Fence} & \rotext{Vegetation} &	\rotext{Trunk} & \rotext{Terrain} &	\rotext{Pole} &\rotext{Traffic sign} & \rotext{\%\,mIoU}\\ \midrule

\midrule

CoSMix (5\,cm)~\:~ & \cite{saltori2022cosmix} & \best{80.9}	& 7.3 &	\best{22.7} &	\best{7.1} &	\best{9.9} &	\best{25.0} &	30.9 &	2.8 &	\best{74.6} &	9.6 &	42.5 &	0.2 &	39.9 &	\best{19.8} &	68.2 &	36.1 &	24.1 &	\best{47.8} &	13.4 &\perf{29.6}&\std{0.8}\\ 
 \midrule 
 \method{}  w/o ST & \llap{(ours)} & 58.4	&	\best{8.4}	 &	13.7	&	1.0 &		3.9	&	19.8	&	45.4	&	\best{2.9}	&	29.1	&	7.3		&35.8	&	0.2		&47.7	&	7.2	&	66.9	&	33.6	&	52.7 &		30.7	&	7.5 & 24.9 & \std{0.2} \\

 \method{} & (ours)& 67.0	&	{7.7}	&	14.4	&	1.3	&	5.2		&24.1	&	 \best{52.6}	 &	2.7	 &	52.5	 &	\best{10.5}	 &	\best{44.1}	 &	\best{0.4}		 & \best{51.8}	 &	13.6 &		\best{69.7}	 &	\best{40.5}	 &	\best{56.5}	 &	45.0	&	\best{14.3} & \best{30.2} & \std{0.4} \\
 \bottomrule

\end{tabular}
\vspace*{-2mm}
\caption{\textbf{Classwise results \synthtosk.} Direct comparison to CoSMix in 5-cm voxel size setting. Average over 3 runs. Color:\colorbox{blue!20}{\color{black}Best}
.} 
\label{tab:experiments_per_class_syn_sk_5cm}
\end{table*}
\begin{table*}[!ht]
\small
\centering
\newcommand*\rotext{\multicolumn{1}{R{45}{1em}}}
\setlength{\tabcolsep}{2.2pt}
\begin{tabular}{lr|ccccccccccccc|c@{\hskip2pt}l}
\toprule
 \rlap{\raisebox{12mm}{\synthtoposs}}%
 \rlap{\raisebox{6mm}{~~~(\%\,IoU)}}%
 \rlap{\raisebox{2mm}{~~~5-cm voxel size}}%
 & & \rotext{Person} & \rotext{Rider} &	\rotext{Car}	& \rotext{Trunk}	& \rotext{Plants} &	\rotext{Traffic sign}	& \rotext{Pole} &	\rotext{Garbage can}& \rotext{Building}	& \rotext{Cone}	& \rotext{Fence} & \rotext{Bike}	& \rotext{Ground} & \rotext{\%\,mIoU}\\ \midrule

\rowcolor{black!10}

\midrule

 \midrule 
 CoSMix & \cite{saltori2022cosmix}& 50.9 &	54.5 &	34.9 &	33.6 &	\best{71.1} & \best{19.4} &	35.6 &	26.8 &	65.2 &	\best{30.4} &	24.0	&\best{6.0} & 	78.5& 40.8 & \std{0.7} \\
 \midrule 
 \method{}  w/o ST & \llap{(ours)} & 44.5	&	41.1	&	\best{60.4}	&	30.7	&	70.0	&	7.1	&	22.3	&	22.1	&	\best{66.0}	&	9.5	&	\best{26.8}	&	4.2	&	\best{78.7} & 37.2 & \std{0.3} \\

 \method{} & (ours) &\best{59.9}& \best{54.6}	&	59.2	&	\best{33.7}	&	69.8	&14.9&\best{40.9}&\best{30.8}&64.5&26.2&22.1&2.7&78.0	& \best{42.9} & \std{0.7} \\
 \bottomrule

\end{tabular}
\vspace*{-2mm}
\caption{\textbf{Classwise results \synthtoposs{}.} Direct comparison to CoSMix in 5-cm voxel size setting. Average over 3 runs. Color:\colorbox{blue!20}{\color{black}Best}.}
\label{tab:experiments_per_class_sl_sp}
\end{table*}

In Tables~\ref{tab:experiments_per_class_ns_sk},~\ref{tab:experiments_per_class_syn_sk}~and~\ref{tab:experiments_per_class_ns_sp} we show the per-class results for the baselines and for \method{}, in the 10-cm voxel size setting.
It can be observed that \method{} improves on the majority of the classes and often with a large gap over a simple source-only model.
In Tables~\ref{tab:experiments_per_class_syn_sk_5cm}~and~\ref{tab:experiments_per_class_sl_sp}, we report the per-class results for a 5-cm voxel size, thus in a direct comparison with the setting used in the CoSMix paper.

\subsection{Frequency-weighted mIoU results}
As an additionally metric,  we provide the frequency-weighted mIoU in Table~\ref{tab:experiments_alls_fwIoU} (for the 10-cm voxel size setting) and in  Table~\ref{tab:experiments_alls_fwIoU_5cm} (the 5-cm voxel size setting). With this metric also, which accounts for the different class frequencies, \method{} achieves SOTA results in the \nstosk{}, \synthtosk{} and \nstoposs{} settings. Especially with \nstosk{} and \synthtosk{}, \method{} outperforms the best baseline methods with a comfortable margin of 4.9 points, respectively 5 points. In the comparison with CoSMix on the 5-cm voxel setting, \method{} reaches as well SOTA on \synthtosk{} and is only 0.3 points behind CoSMix on \synthtoposs{}.
\label{sup:sec:fwiou_results}
\begin{table}[!h!]
    \small
    \setlength{\tabcolsep}{1pt}
    \centering
        \begin{tabular}{l|c|c|c|}
            \toprule

            Method & \nstosk  & \synthtosk & \nstoposs \\
            \midrule
            \rowcolor{black!10}
            Src only &  57.2 &  41.3 & 75.6   \\ 
            \midrule
            AdaBN &  68.7 &  51.4 & 75.5  \\ 
            DUA &  69.7 &  50.9 &  75.6 \\ 
            MixedBN &  69.5 &  52.3 & 75.9 \\
            MinEnt &  \back{69.5} &  \back{52.3} & 76.0    \\
            Coral   &  \back{69.5} & 51.8 & 76.1 \\
            LogCoral &  \back{69.5} &  \back{52.3} & 75.7 \\
            ST &  60.5 &  46.2 & \underline{77.5} \\
            \cosmix &   60.1 &  50.4  & 77.3 \\

            \midrule
            \method{} w/o ST &  \underline{71.8} &  \underline{53.6}  &  76.5  \\
            \method{} &  \textbf{74.6} & \textbf{57.3}   & \textbf{78.5}  \\
            
            \bottomrule
        \end{tabular}
        \vspace*{-0.9mm}
    \caption{
    \textbf{Frequency-Weighted IoU\% on target} (avg. of 3 runs). All methods are evaluated in the same 10-cm voxel size setting. For C\&L are no per class results available.}
\label{tab:experiments_alls_fwIoU}
\end{table}

\begin{table}[!h!]
    \small
    \setlength{\tabcolsep}{1pt}
    \centering
    \begin{tabular}{l|c|c}
    \toprule  
     Setting &  \synthtosk & \synthtoposs \\
    \midrule
           
    \cosmix &   \underline{54.6} & \textbf{65.7} \\
           
    \midrule
    \method{} w/o ST  &   47.8 & \underline{65.5}   \\
    \method{} &  \textbf{55.9} &  65.4  \\
            
    \bottomrule
    \end{tabular}
    \vspace*{-0.9mm}
    \caption{
    \textbf{Frequency-Weighted IoU\% on target} (avg. of 3 runs). All methods are evaluated in the same 5-cm voxel size setting.}
\label{tab:experiments_alls_fwIoU_5cm}
\end{table}

\section{Qualitative results}
 \label{sup:sec:qualitative_results}
 
 \subsection{Scenes of \DAsetting{NS}{\skns}}
 
\begin{figure*}
\newcommand{\rotext}[1]{{\begin{turn}{90}{#1}\end{turn}}}
\setlength{\tabcolsep}{1pt}
\centering
\begin{tabular}{ccc}


\begin{tikzpicture}[baseline=-13mm]
\node(a){\includegraphics[width=0.3\linewidth]{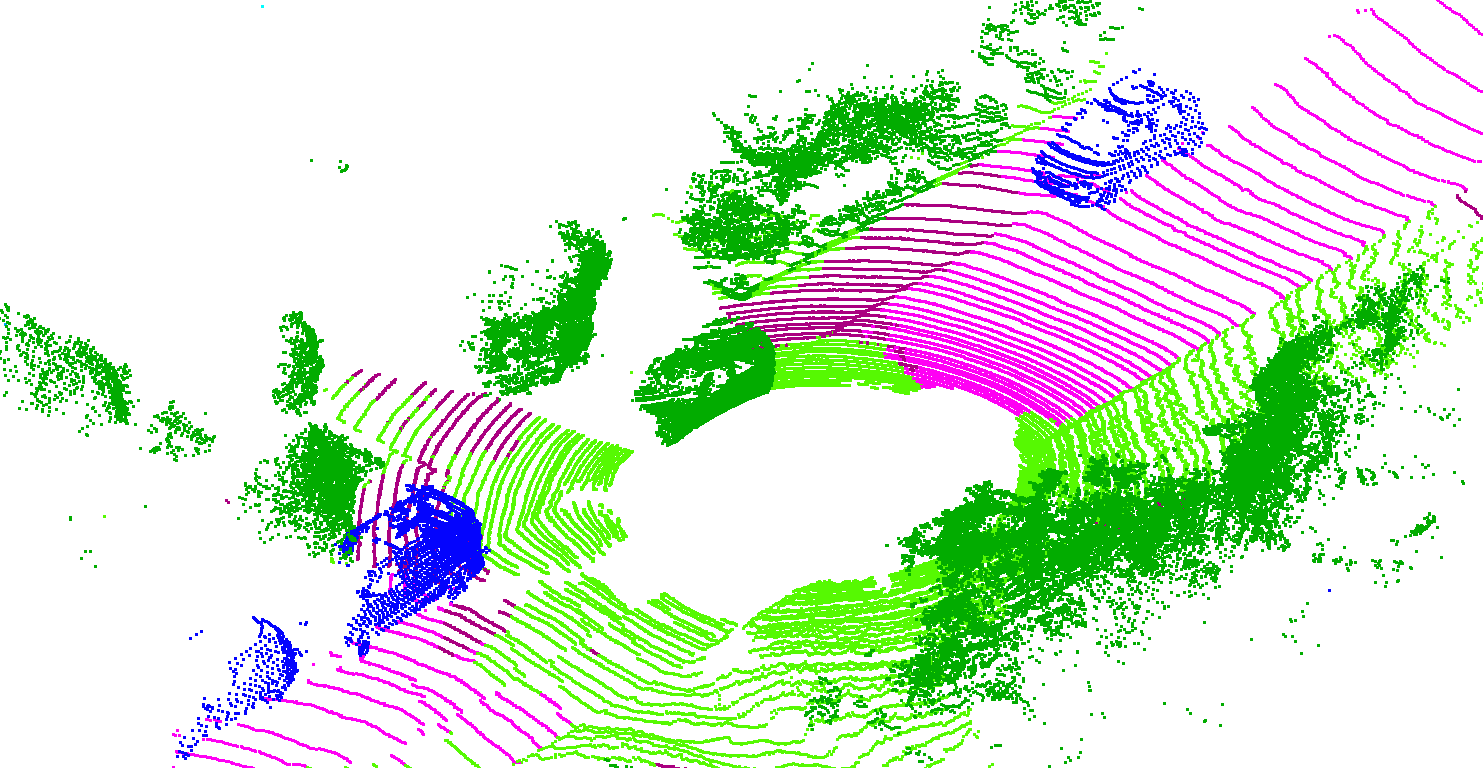}};
\node at(a.center)[draw, red,line width=1pt,ellipse, minimum width=30pt, minimum height=20pt,rotate=-5,xshift=0pt, yshift=0pt]{};
\end{tikzpicture} 
& 
\includegraphics[width=0.3\linewidth]{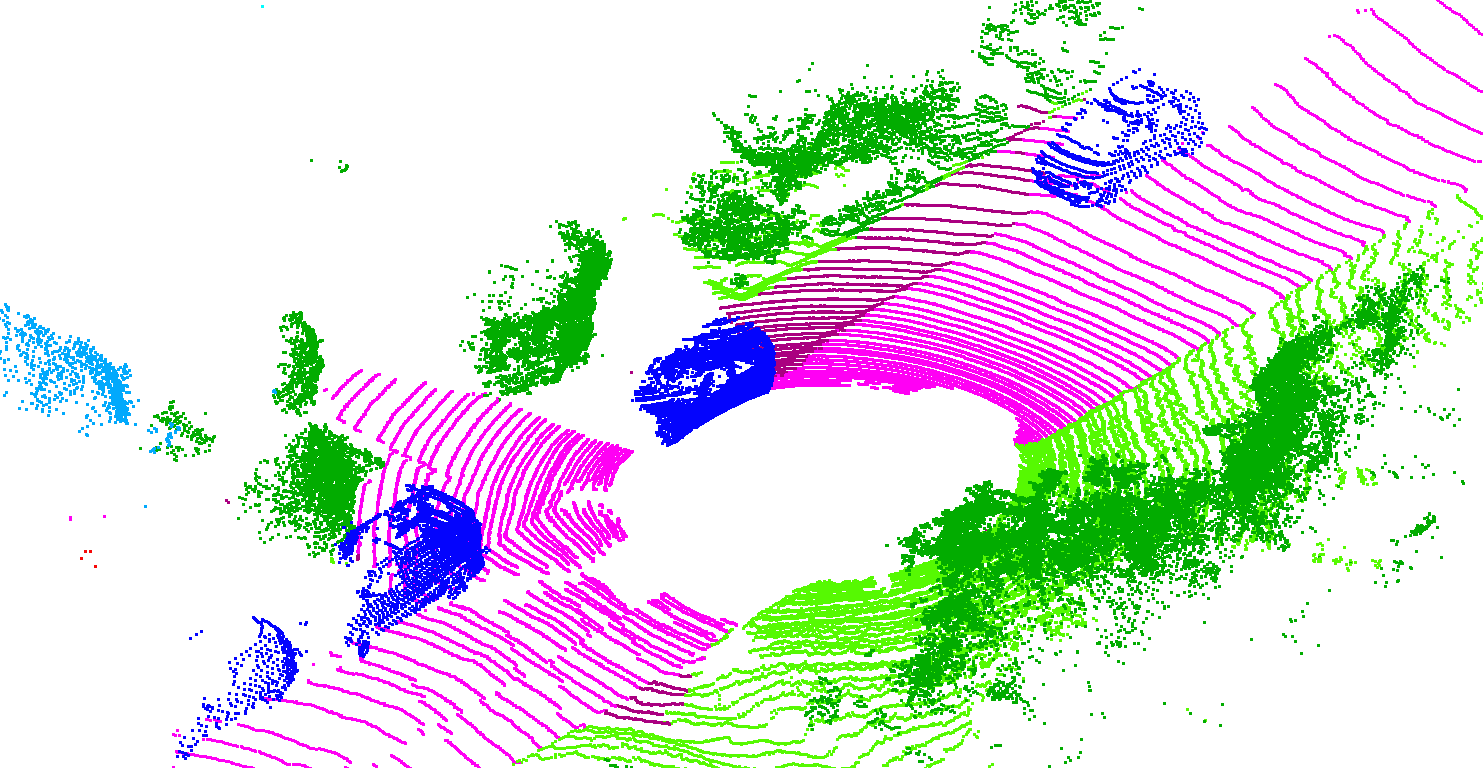}
&
\includegraphics[width=0.3\linewidth]{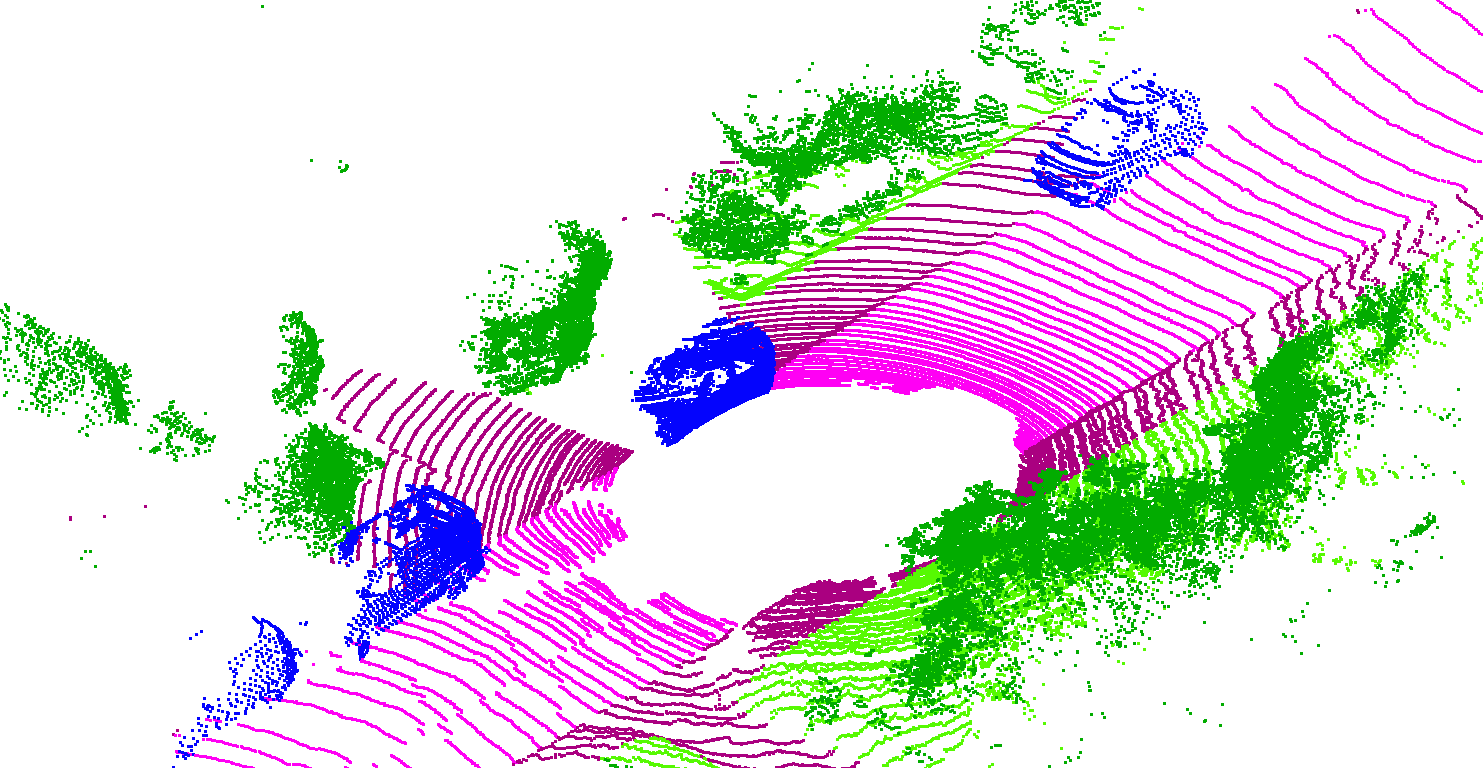} 
\\
Source-only & \method{} & GT\\

\begin{tikzpicture}[baseline=-16mm]
\node(a){\includegraphics[width=0.3\linewidth]{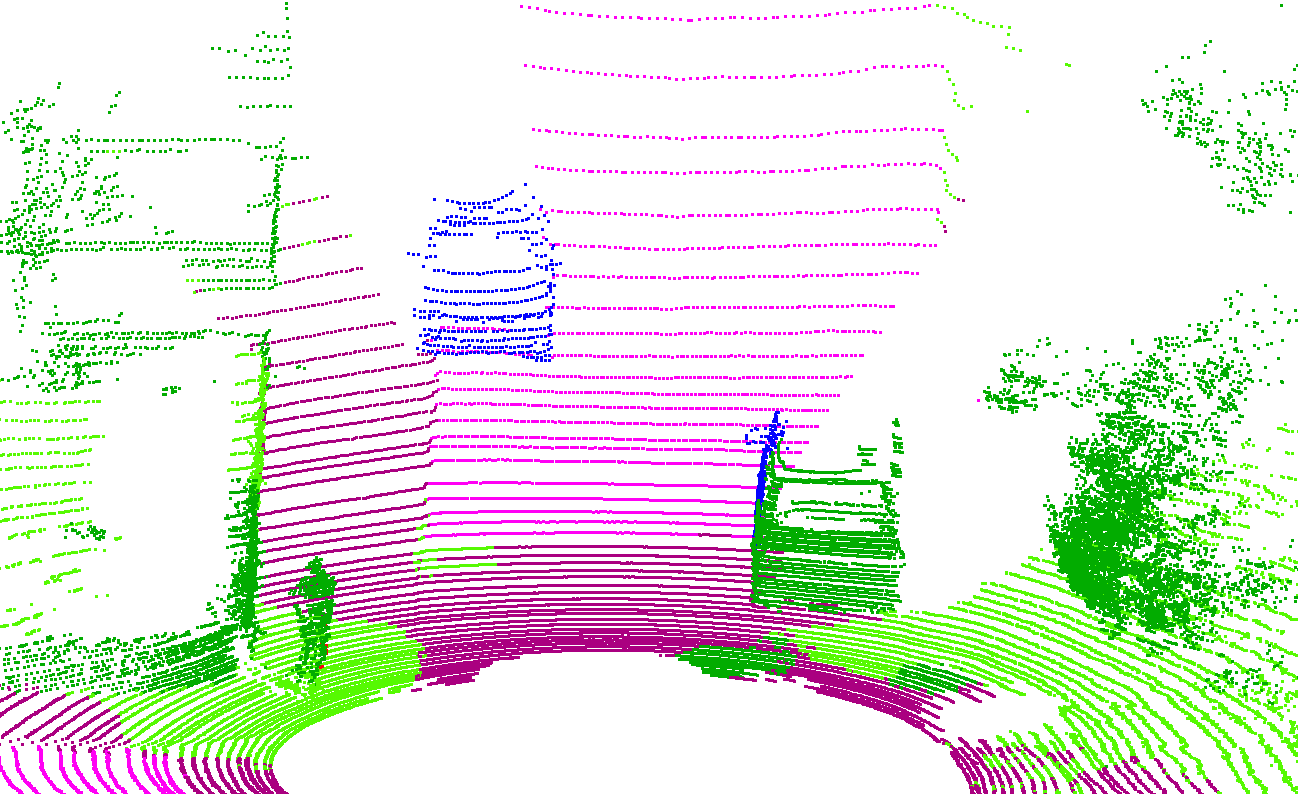}};
\node at(a.center)[draw, red,line width=1pt,ellipse, minimum width=10pt, minimum height=20pt,rotate=-5,xshift=-36pt, yshift=-30pt]{};
\node at(a.center)[draw, red,line width=1pt,ellipse, minimum width=25pt, minimum height=25pt,rotate=-5,xshift=22pt, yshift=-12pt]{};
\end{tikzpicture} & 
\includegraphics[width=0.3\linewidth]{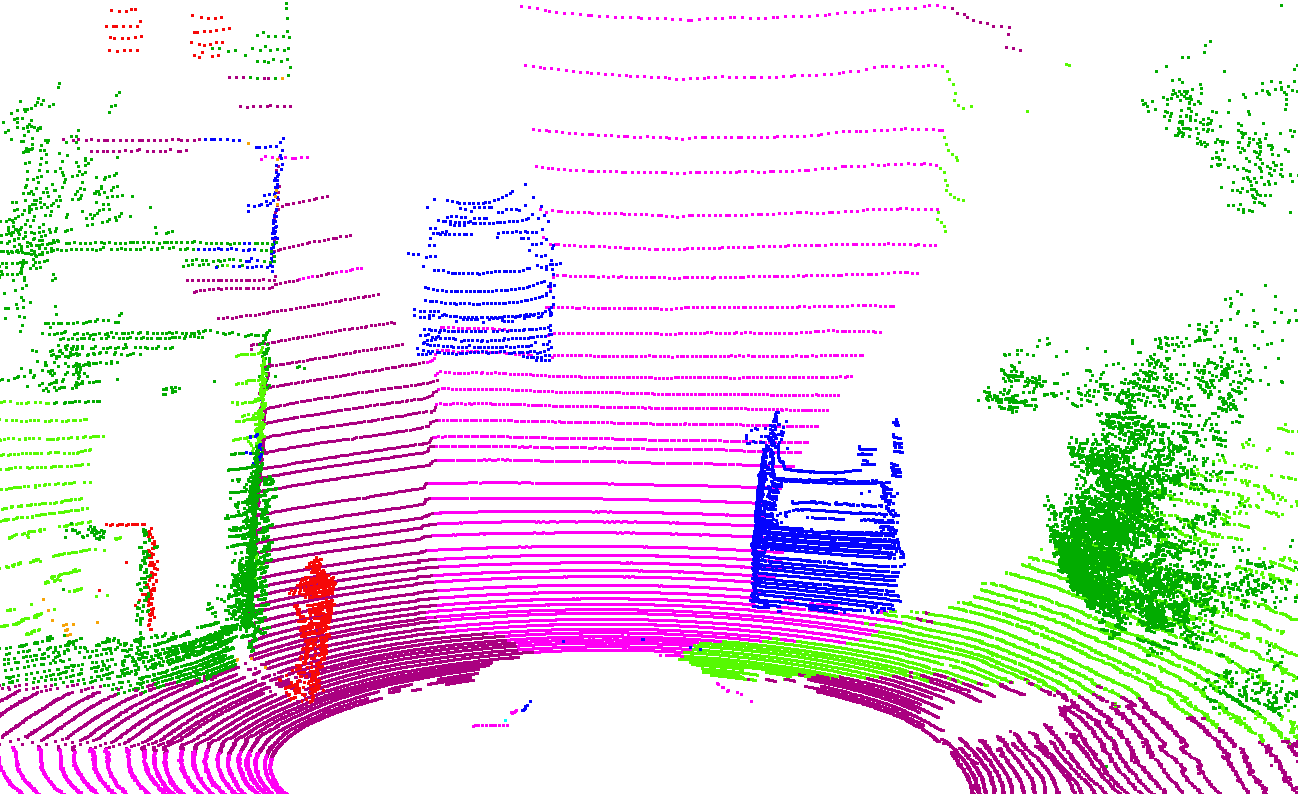}&
\includegraphics[width=0.3\linewidth]{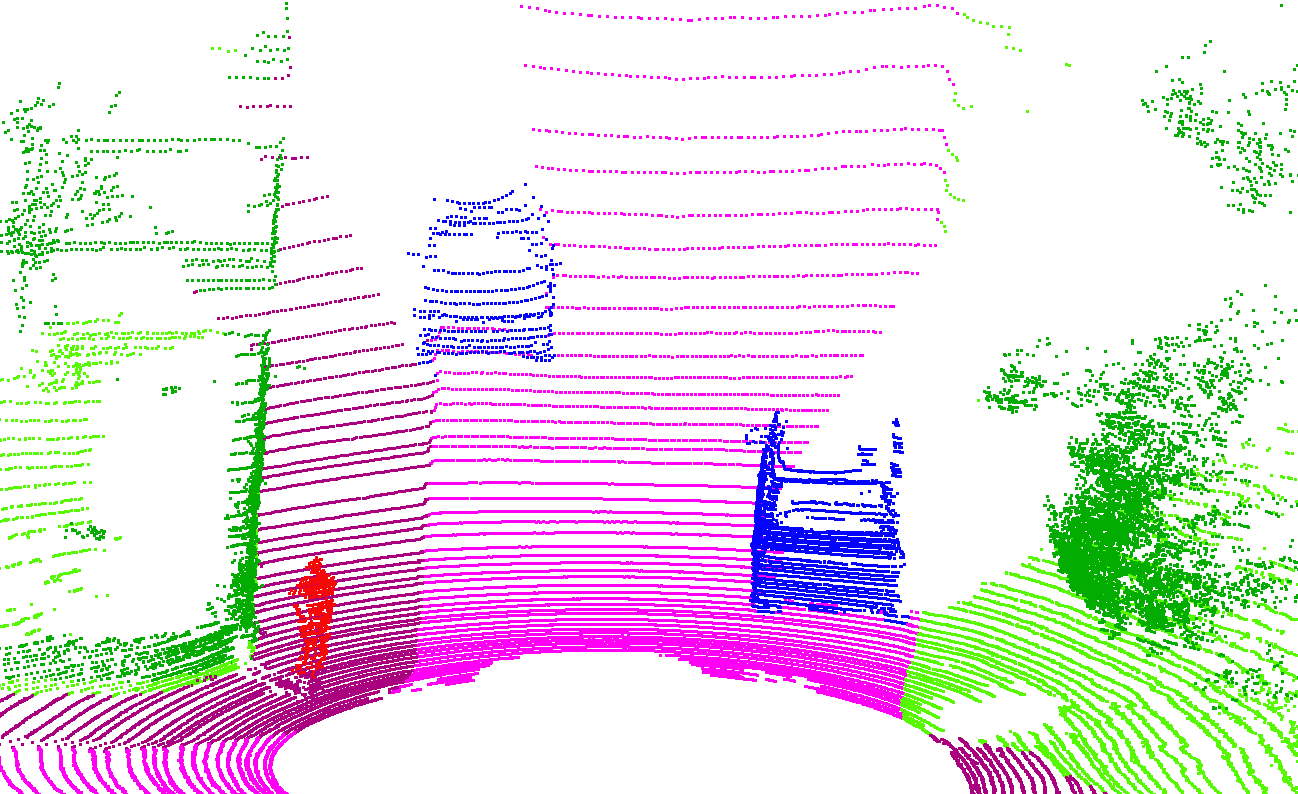}\\
Source-only & \method{} & GT\\

\begin{tikzpicture}
\node(a){\includegraphics[width=0.3\linewidth]{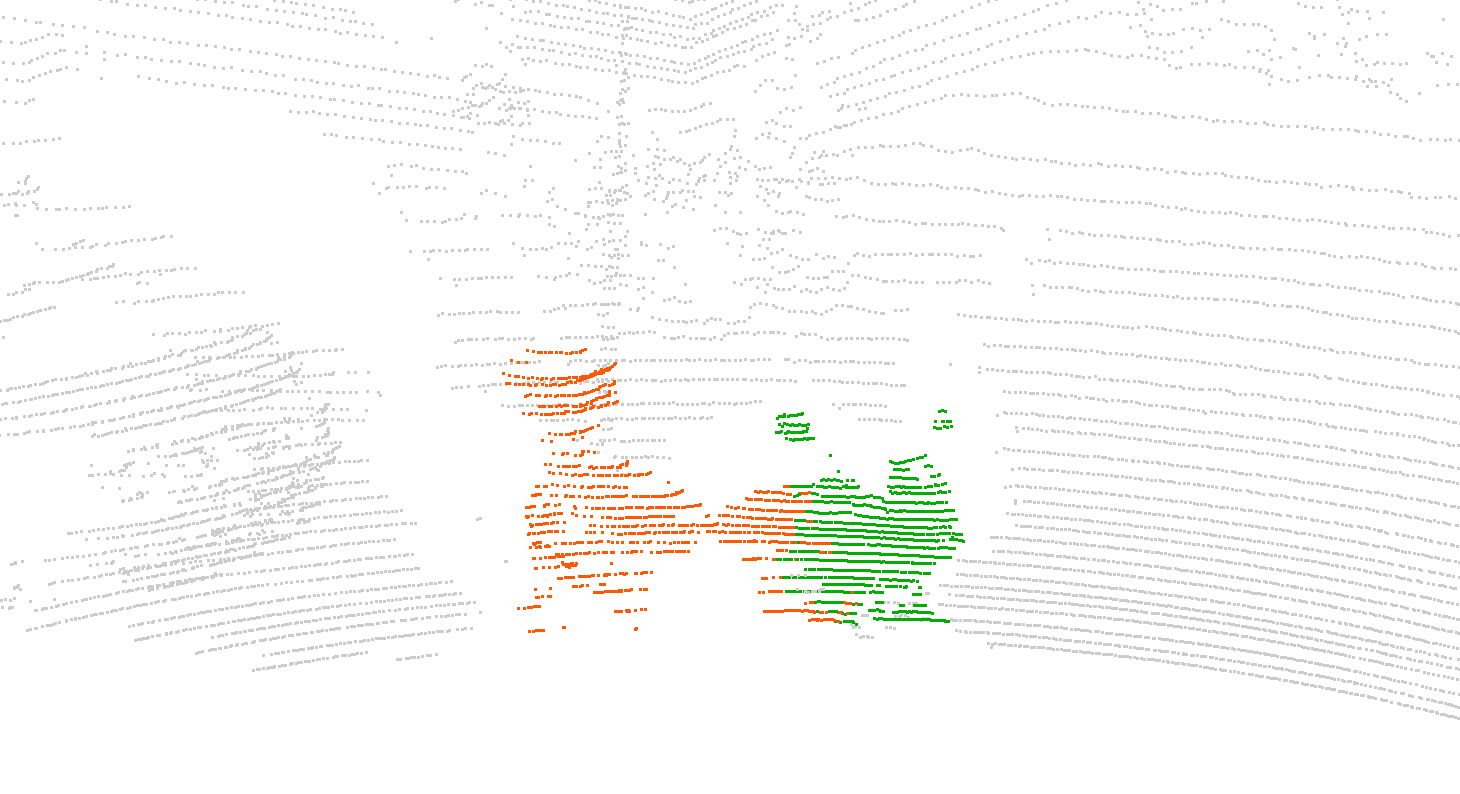}};
\node at(a.center)[draw, red,line width=1pt,ellipse, minimum width=22pt, minimum height=30pt,xshift=15pt, yshift=-15pt,rotate=-5]{};
\end{tikzpicture} & 
\includegraphics[width=0.3\linewidth]{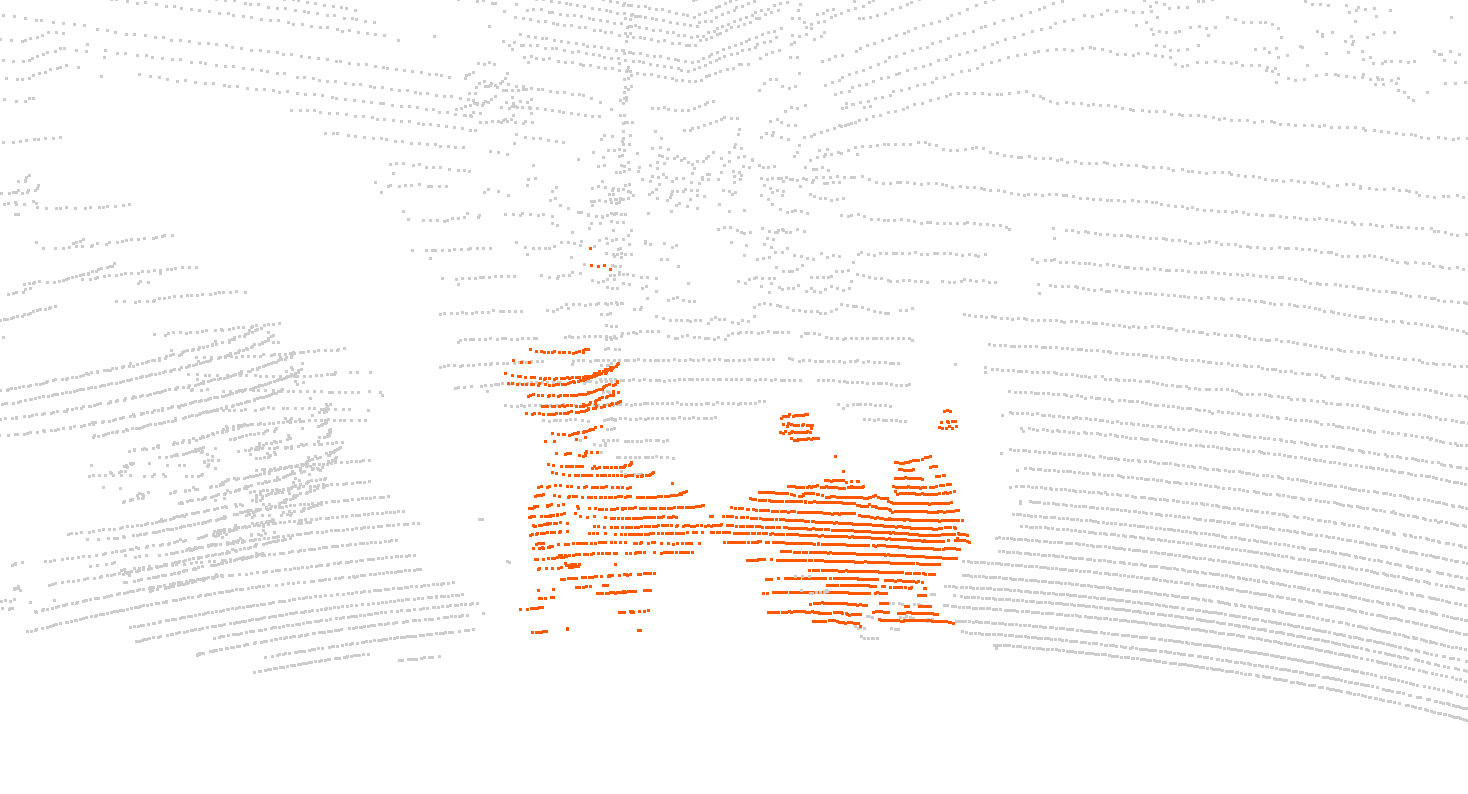}&
\includegraphics[width=0.3\linewidth]{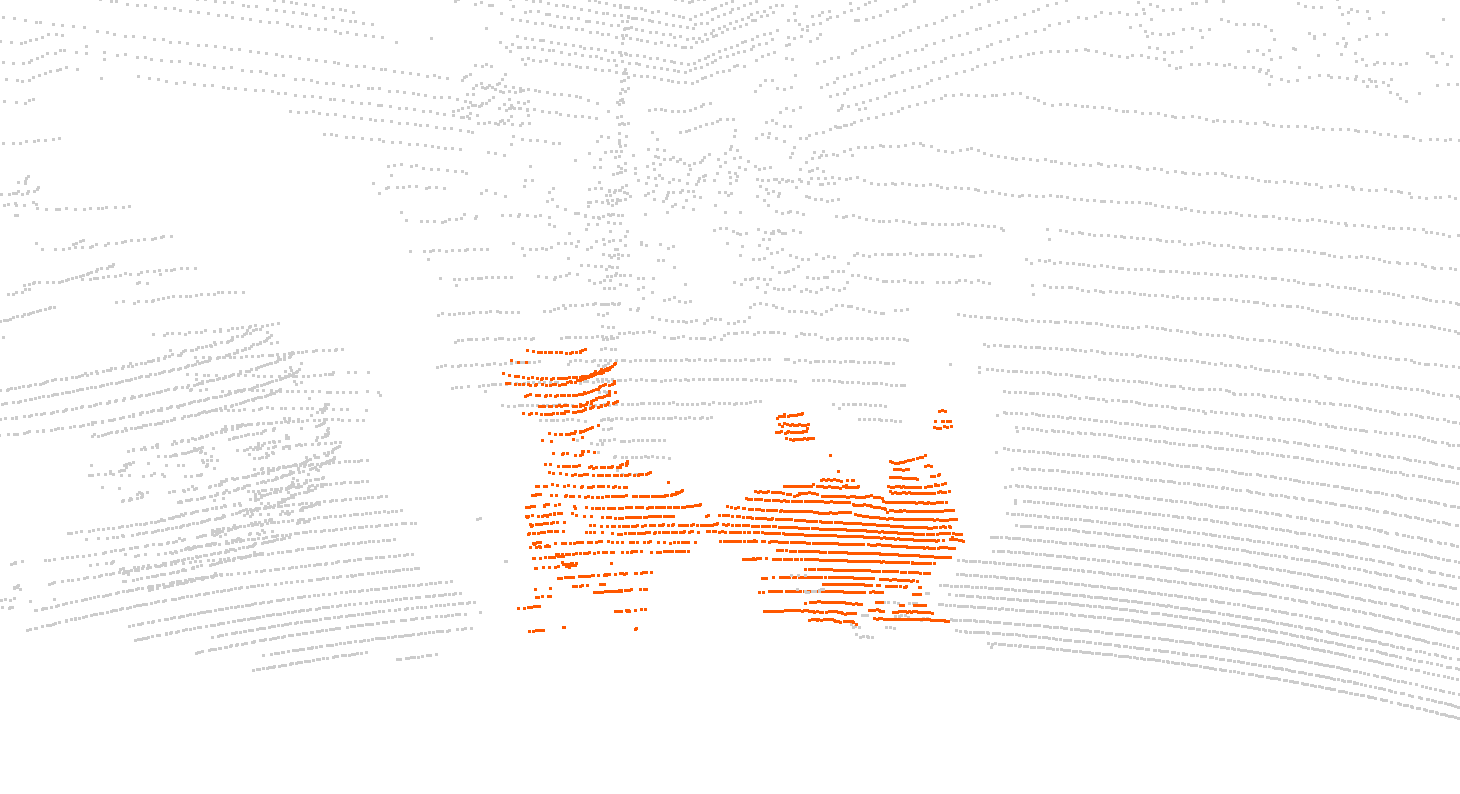}\\
Source-only & \method{} & GT\\
(non-motorcycle points are grey) & (non-motorcycle points are grey) & (non-motorcycle points are grey)\\

\begin{tikzpicture}[baseline=-21mm]
\node(a){\includegraphics[width=0.3\linewidth]{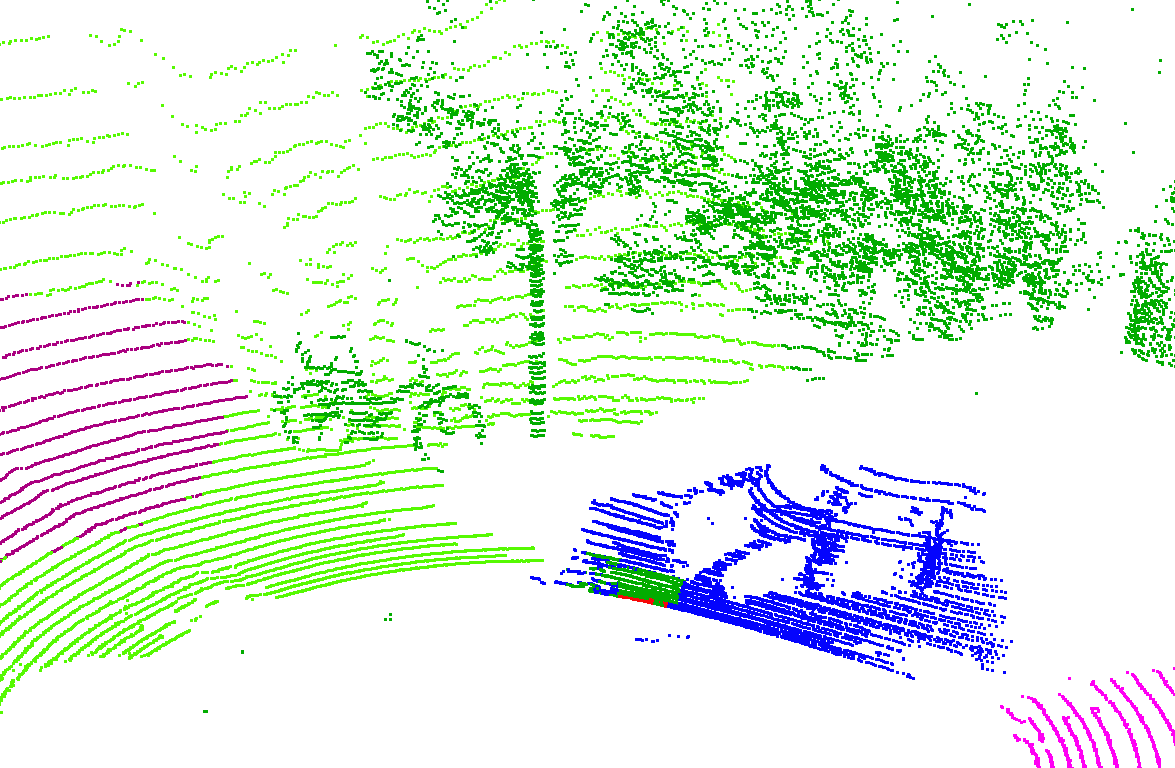}};
\node at(a.center)[draw, red,line width=1pt,ellipse, minimum width=30pt, minimum height=15pt,rotate=-5,xshift=-25pt, yshift=-6pt]{};
\end{tikzpicture} & 
\begin{tikzpicture}[baseline=-19mm]
\node(a){\includegraphics[width=0.3\linewidth]{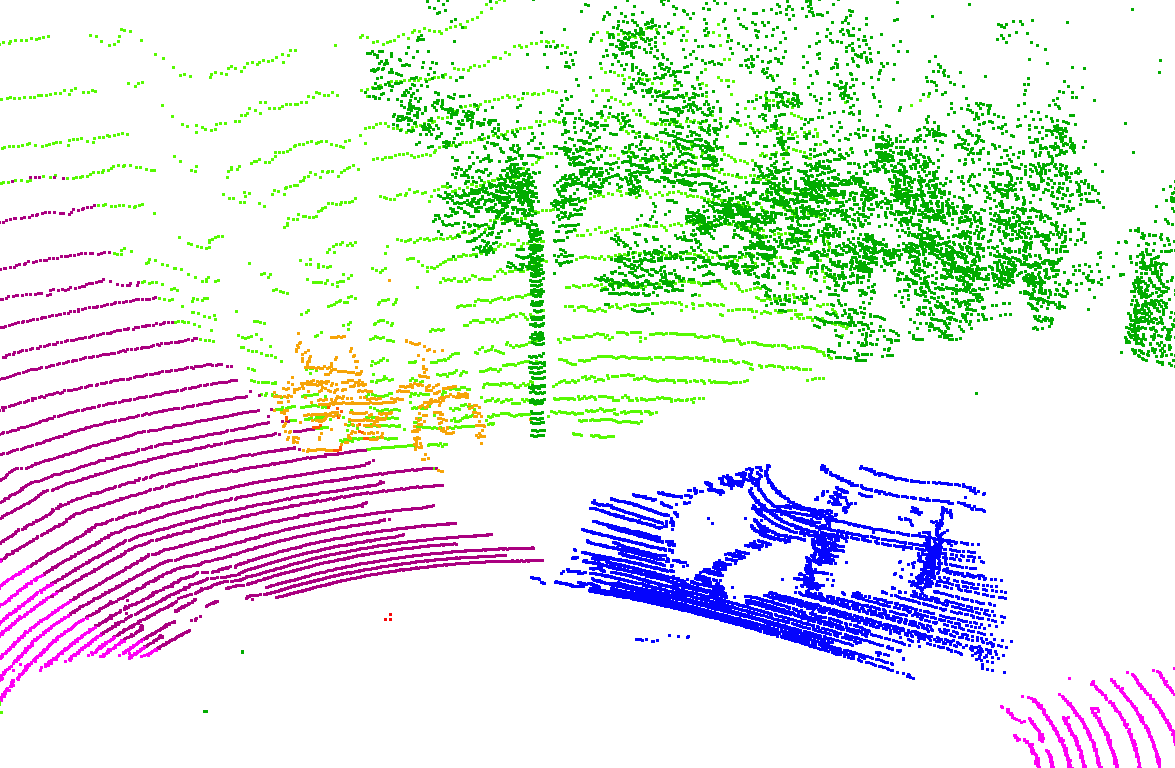}};
\end{tikzpicture}&
\includegraphics[width=0.3\linewidth]{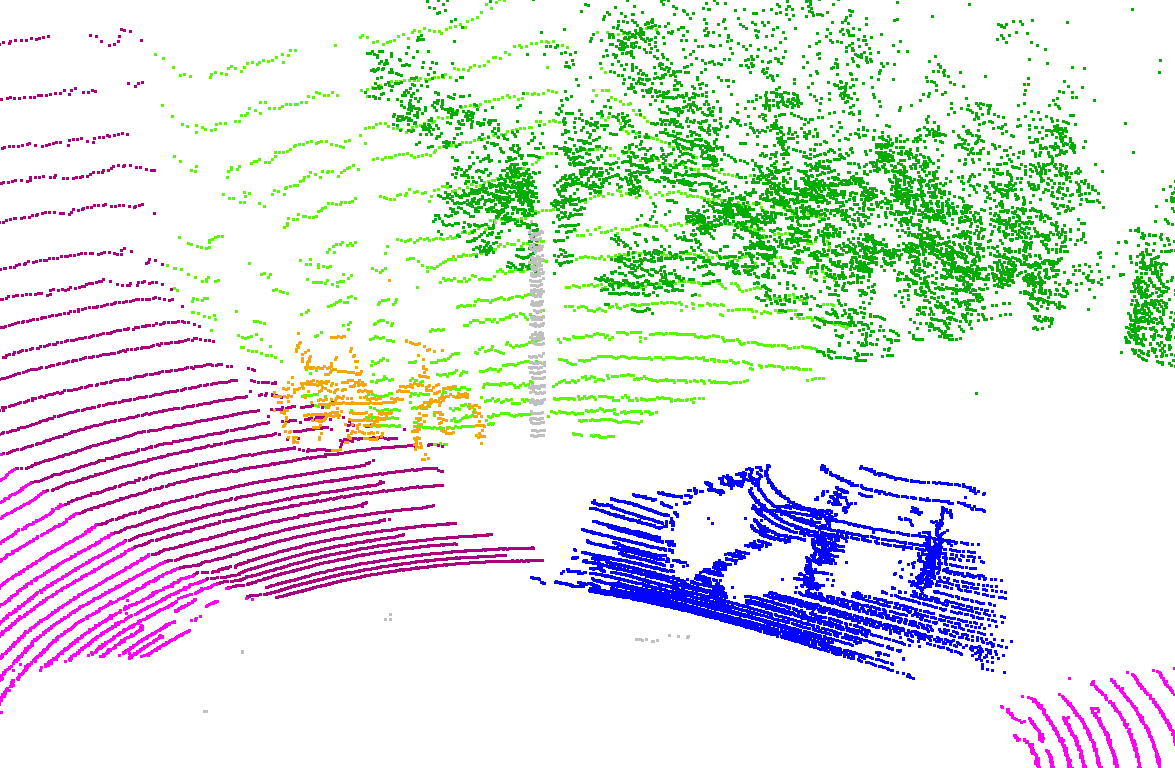}\\
Source-only & \method{} & GT\\

\begin{tikzpicture} 
\node(a){\includegraphics[width=0.3\linewidth]{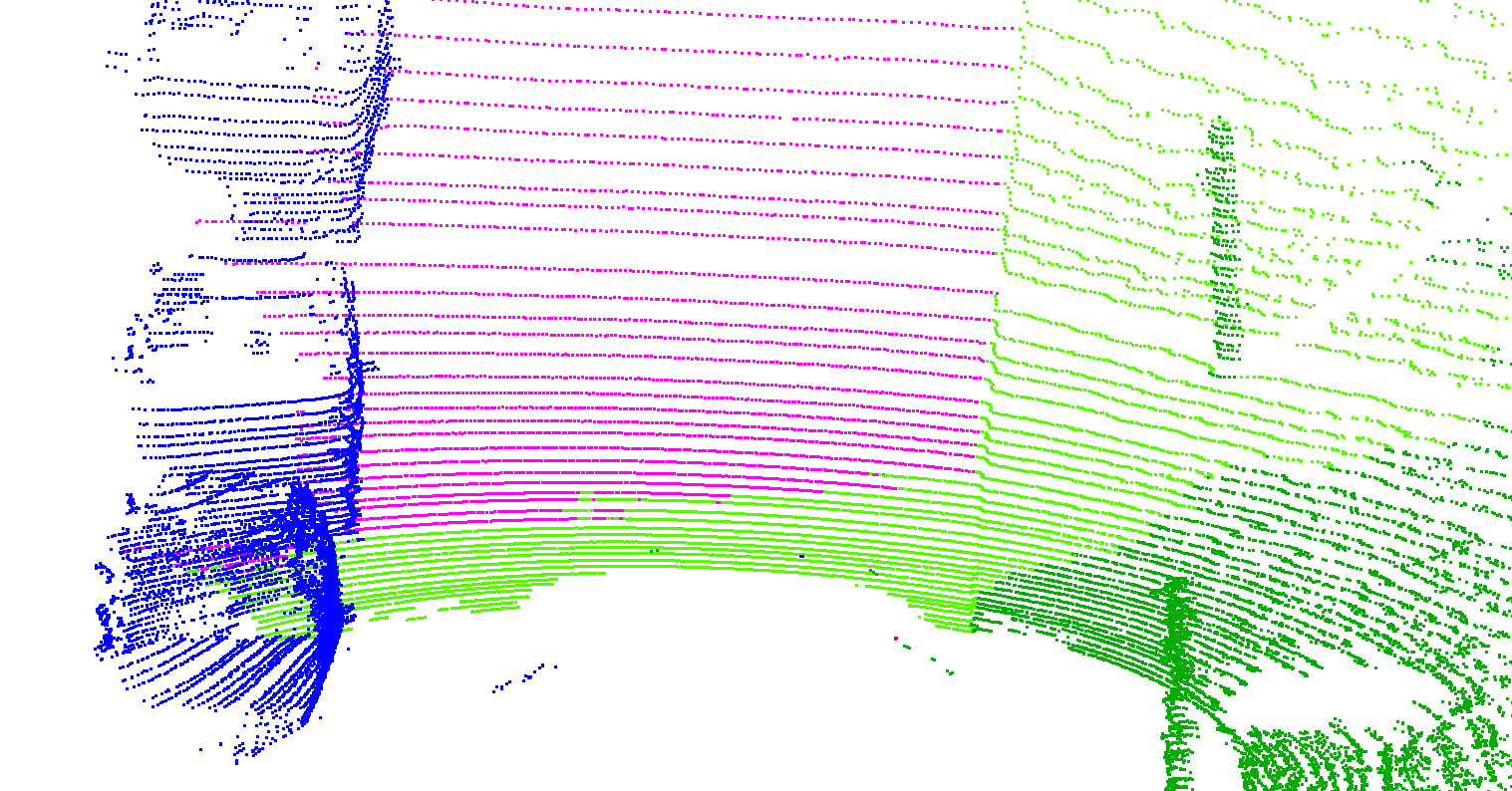}};
\node at(a.center)[draw, red,line width=1pt,ellipse, minimum width=60pt, minimum height=13pt,rotate=5,xshift=-10pt, yshift=-17pt]{};

\node at(a.center)[draw, red,line width=1pt,ellipse, minimum width=35pt, minimum height=15pt,rotate=-30,xshift=60pt, yshift=7pt]{};
\end{tikzpicture} & 
\includegraphics[width=0.3\linewidth]{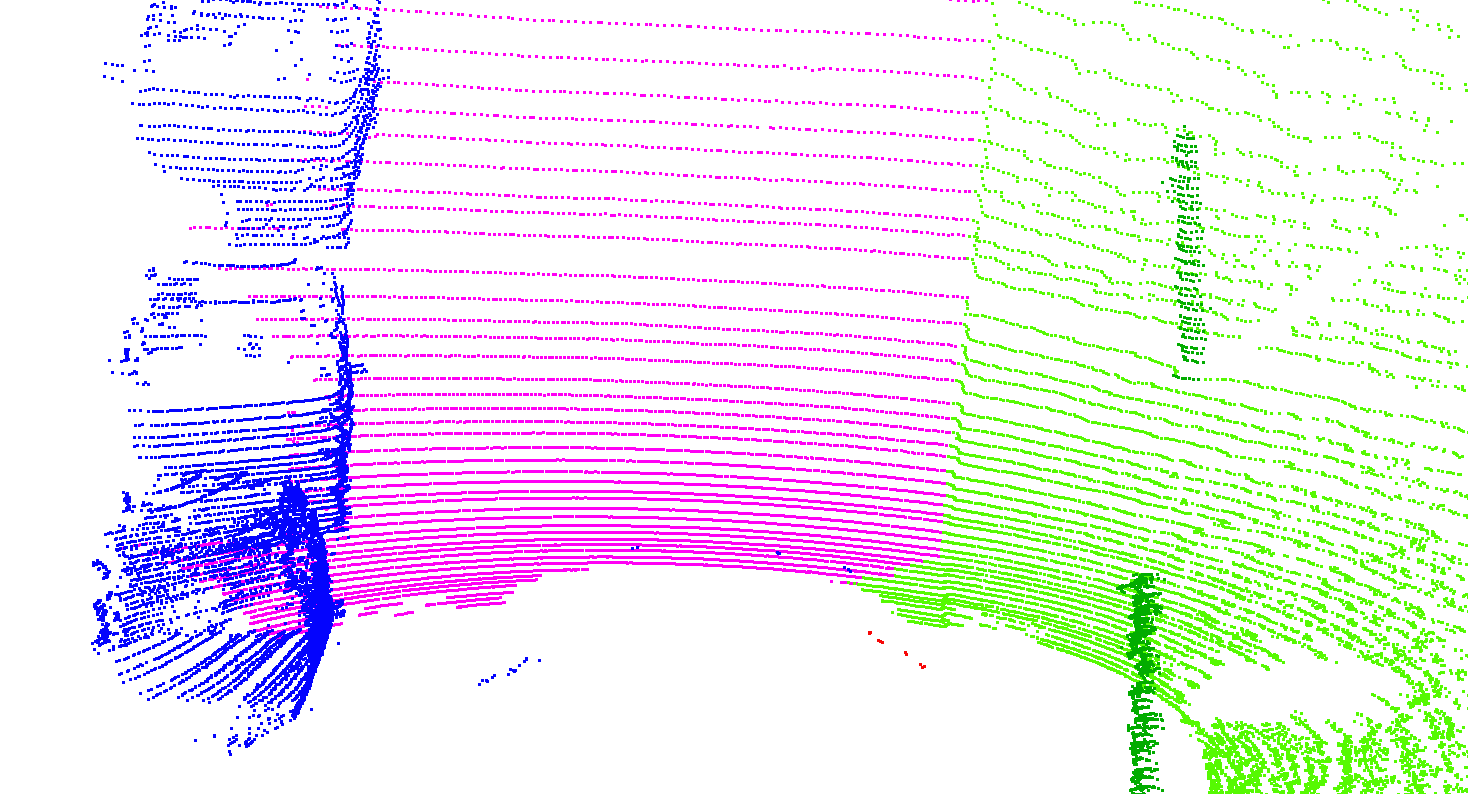}&
\includegraphics[width=0.3\linewidth]{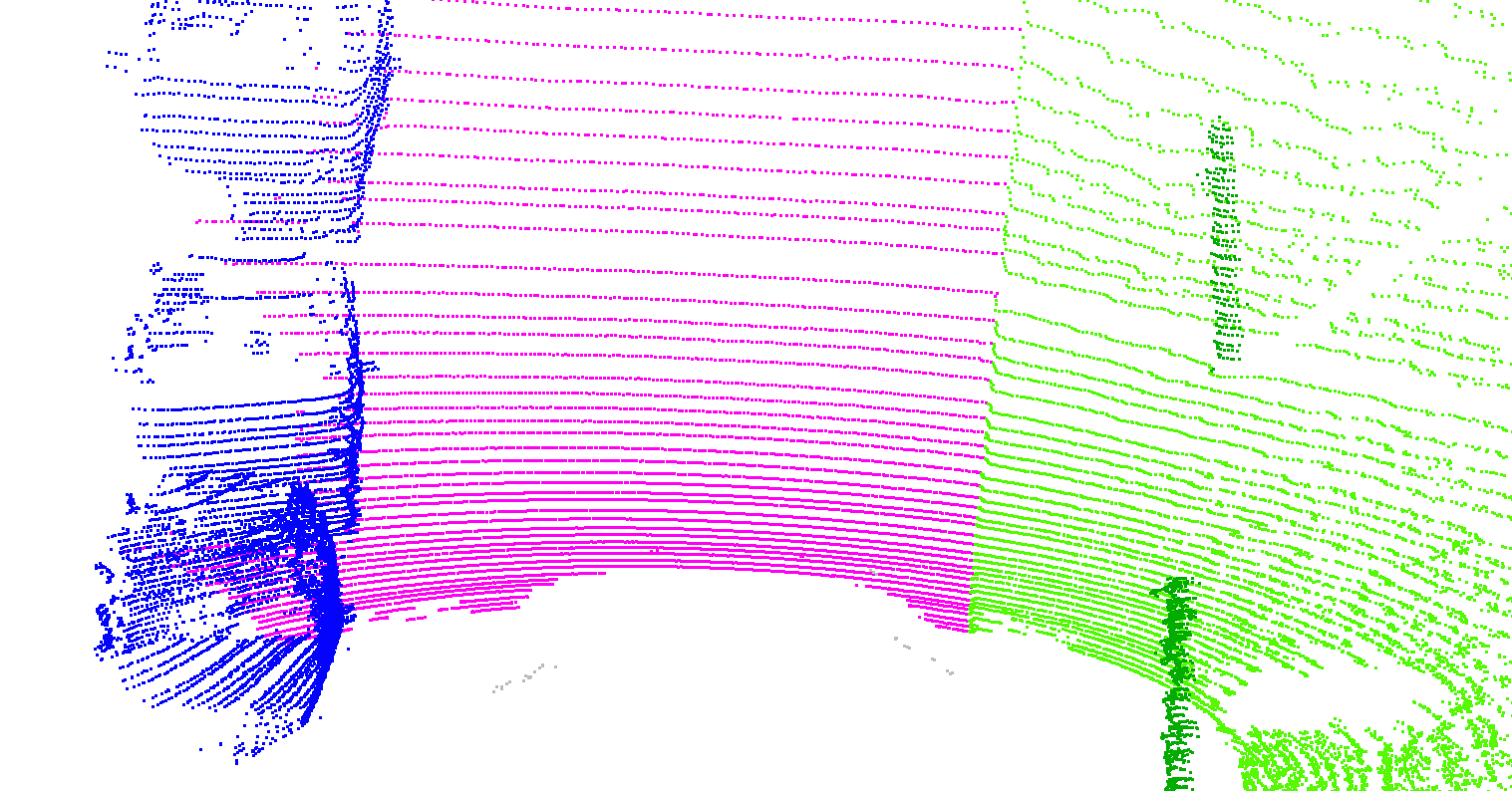}\\
Source-only & \method{} & GT\\

\end{tabular}
\caption{\textbf{Samples of semantic segmentation results in the \DAsetting{\ns}{\skns} setting}
for the Source-only method and for \method{}, to compare with the ground truth (GT). The red circles highlight wrong segmentations.}
\label{fig:app:qualitative_zoom_ns_sk}
\vspace{1cm}
\end{figure*}

In Figure~\ref{fig:app:qualitative_zoom_ns_sk}, 
we compare in the \nstosk{} setting predictions made by \method{} to predictions made by the source-only model. The most noticeable errors are marked with a red circle. 
The quantitative improvements observed earlier on classes such as \colorbox{skcar}{\color{white}"car"}, \colorbox{skbicycle}{\color{white}"bicycle"}, \colorbox{skmotorcycle}{\color{white}"motorcycle"} or \colorbox{skdrivable}{\color{white}"driveable surface"} obtained by \method over the source-only method can be noticed as well in these figures. For example, the car in the first row or the bicycle in the fourth row are not detected at all by the source-only model but are well segmented by \method. Nevertheless, \method is not flawless as, for example, parts of the front wheel of the bicycle are wrongly classified as vegetation (row 4).

\begin{figure*}
\newcommand{\rotext}[1]{{\begin{turn}{90}{#1}\end{turn}}}
\setlength{\tabcolsep}{1pt}
\centering
\begin{tabular}{ccc}

\includegraphics[trim=40 25 40 0,clip,width=0.30\linewidth]{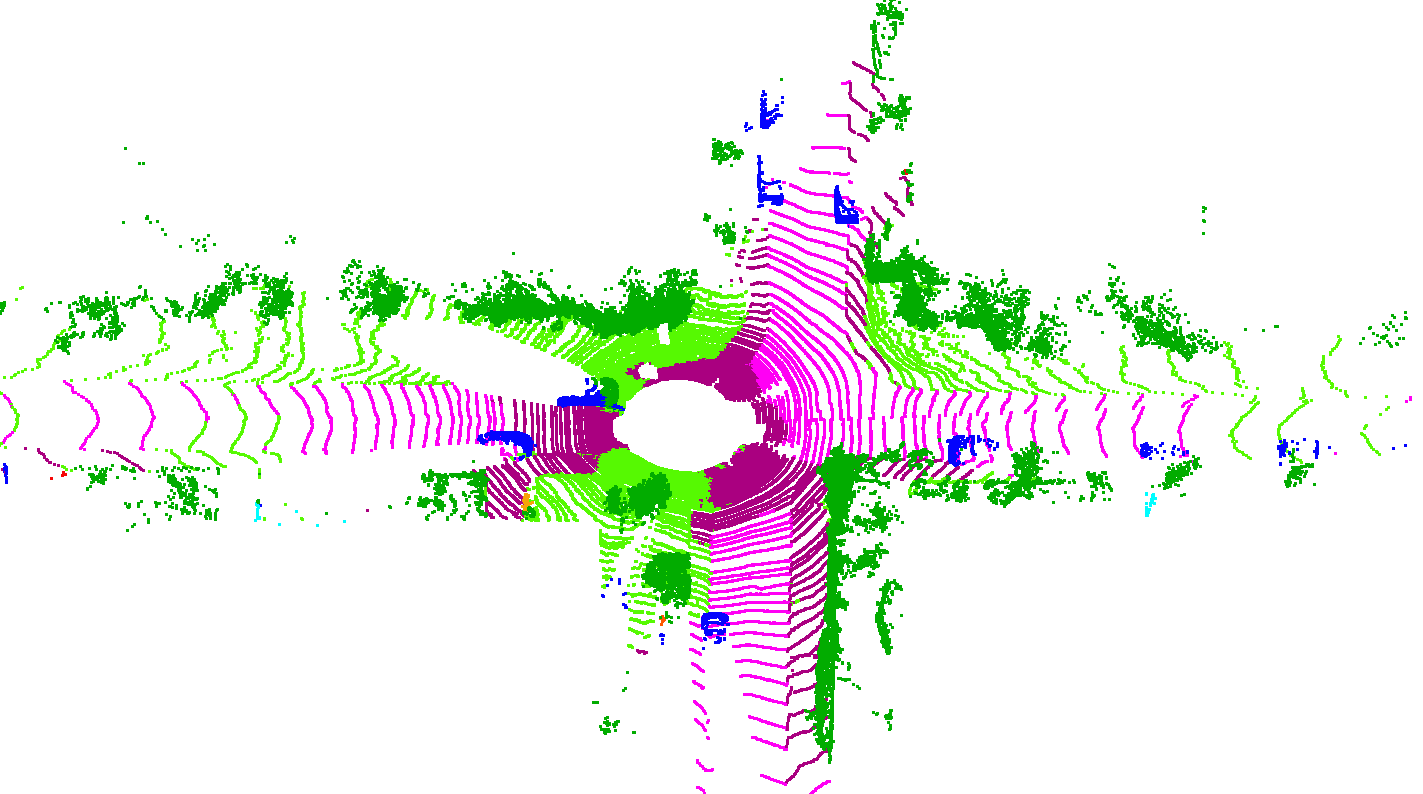}& 
\includegraphics[trim=40 25 40 0,clip,width=0.30\linewidth]{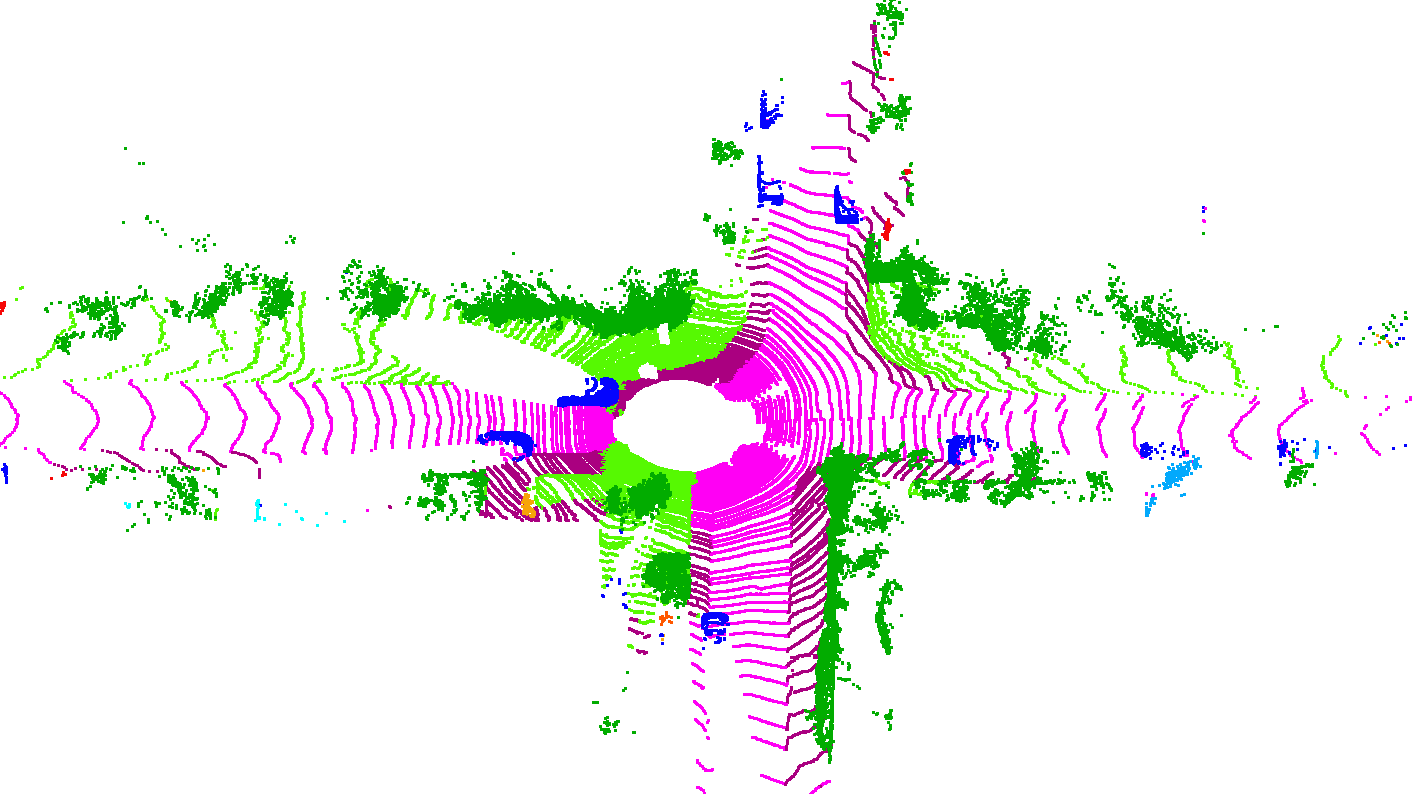}& 
\includegraphics[trim=40 25 40 0,clip,width=0.30\linewidth]{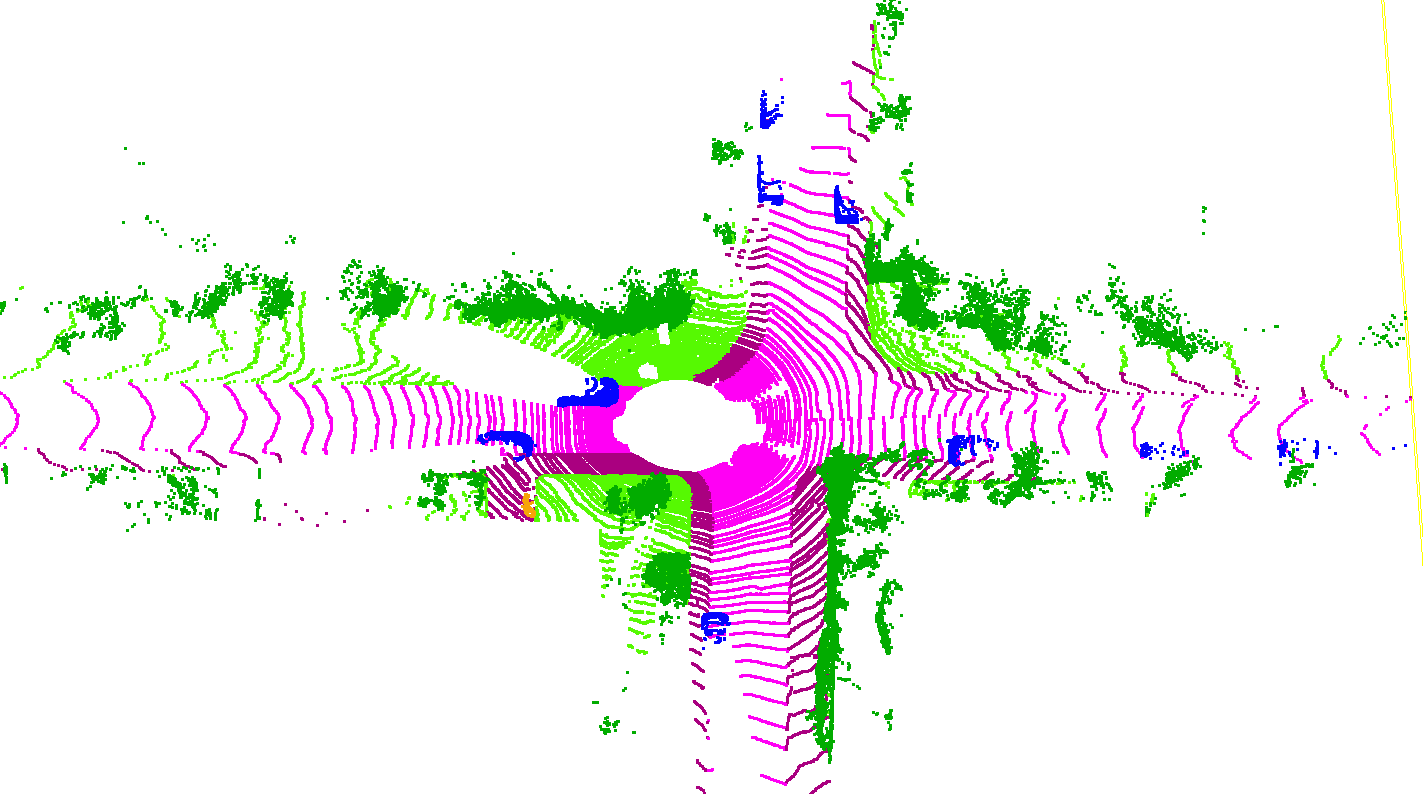}\\

 Source-only&  \method{} & GT\\
 
\includegraphics[trim=40 25 40 0,clip,width=0.3\linewidth]{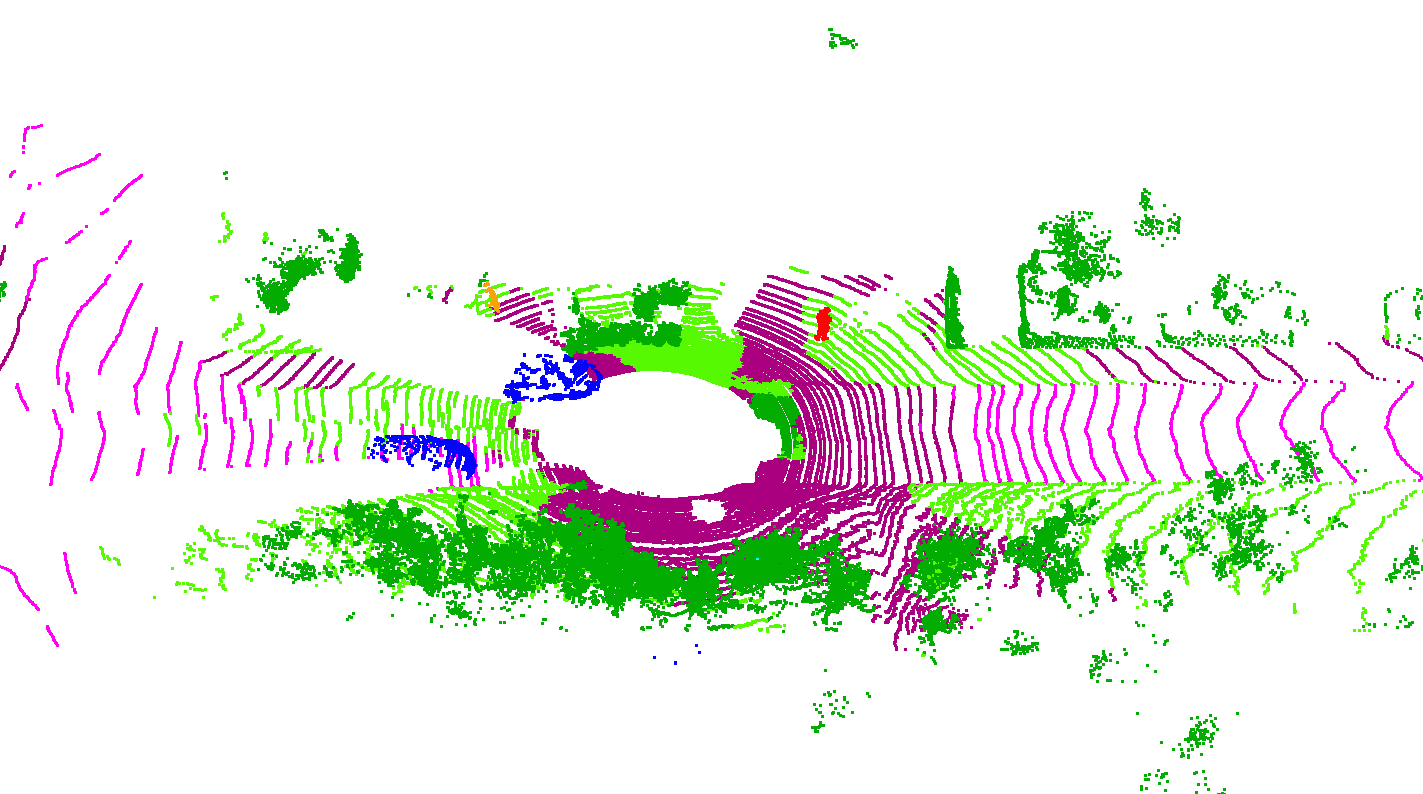}& 
\includegraphics[trim=40 25 40 0,clip,width=0.3\linewidth]{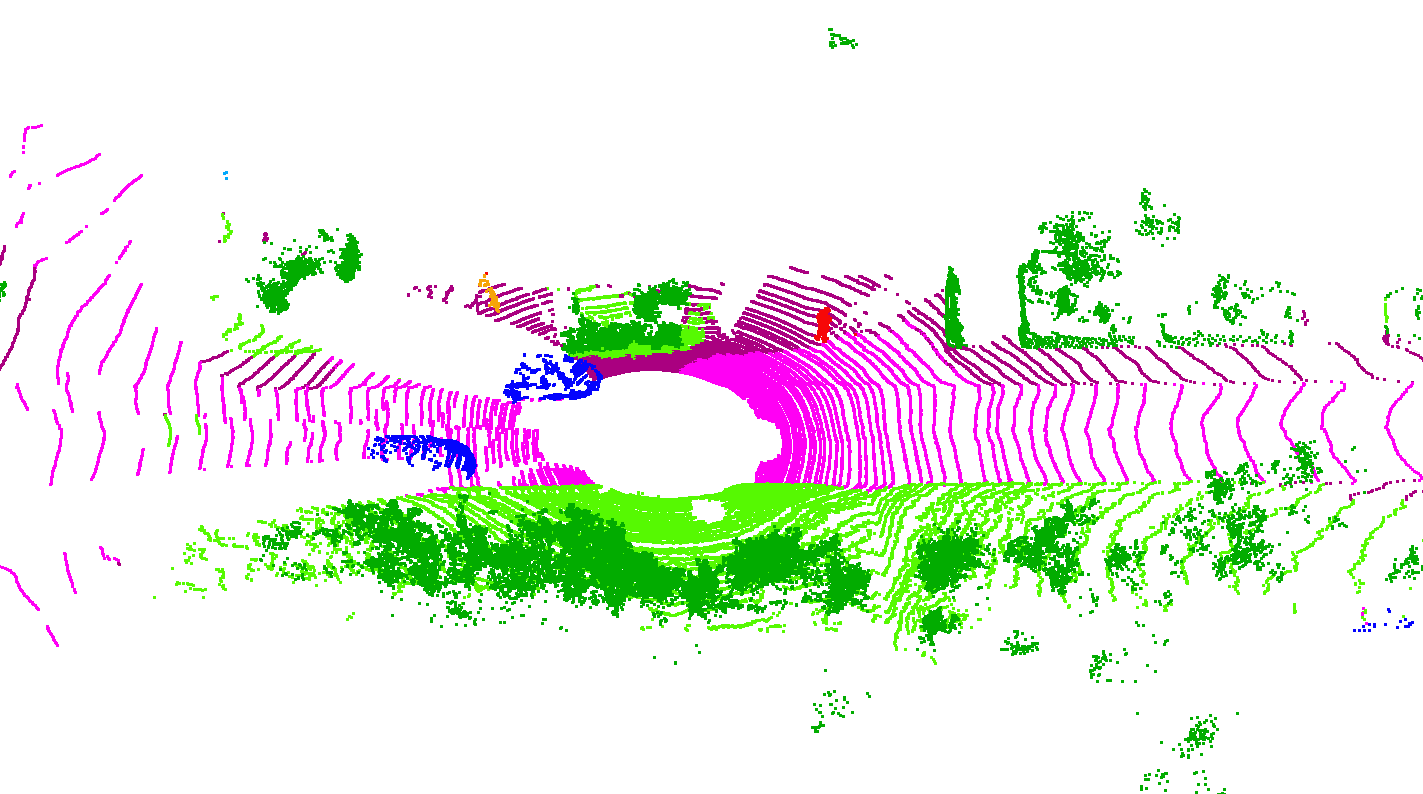}& 
\includegraphics[trim=40 25 40 0,clip,width=0.3\linewidth]{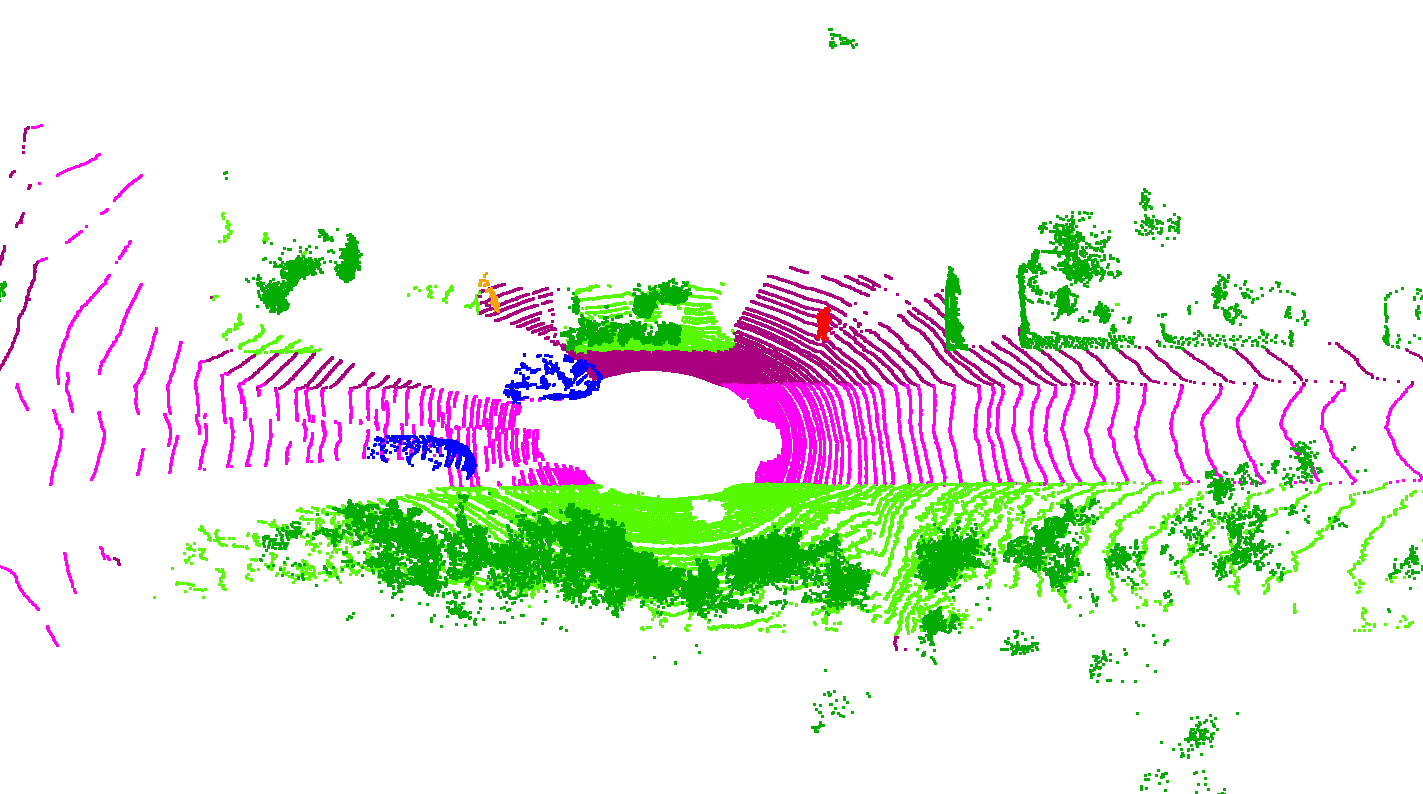}\\

 Source-only&  \method{} & GT\\

   \includegraphics[trim=40 25 40 0,clip,width=0.30\linewidth]{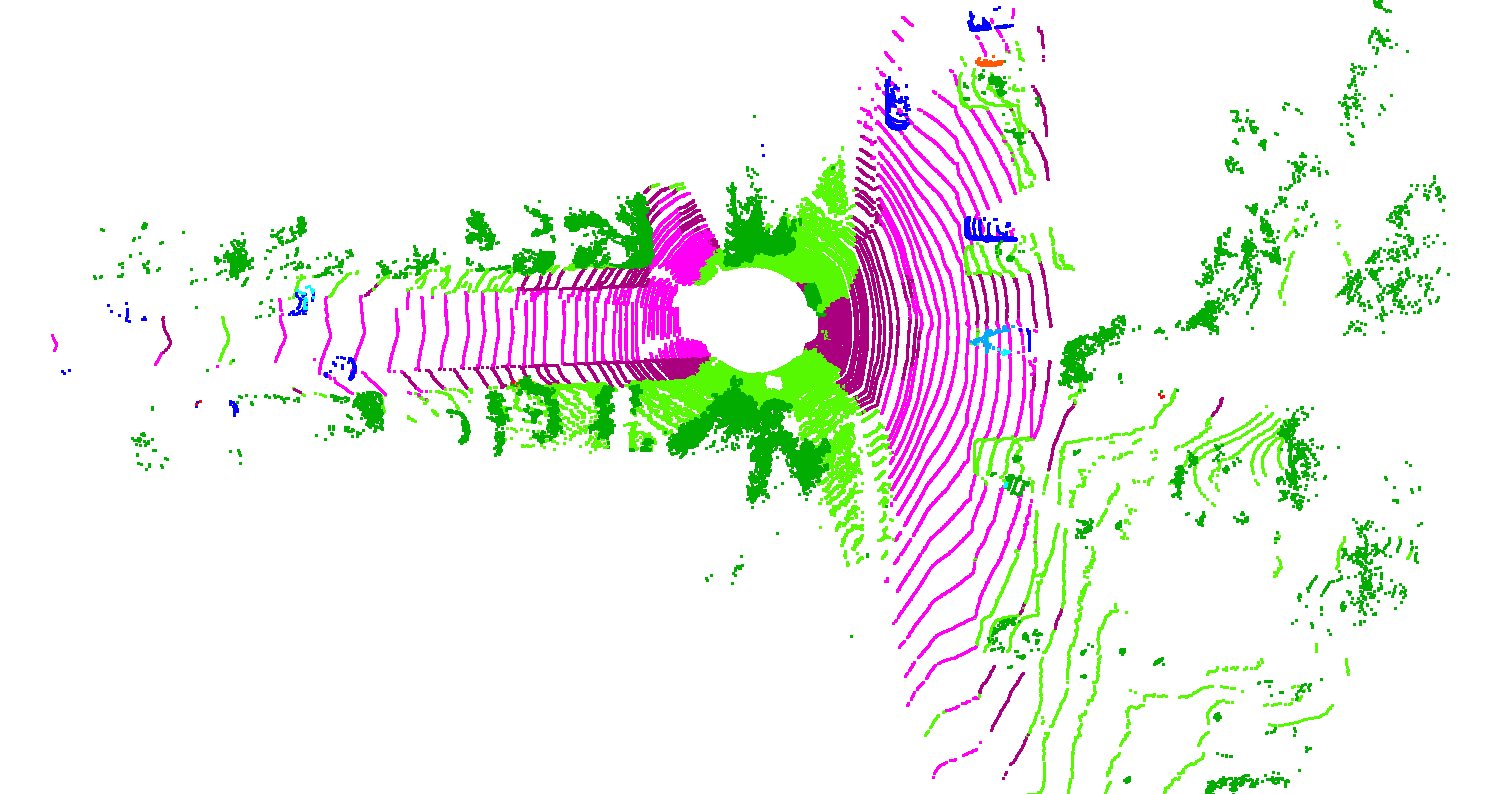}&
\includegraphics[trim=40 25 40 0,clip,width=0.30\linewidth]{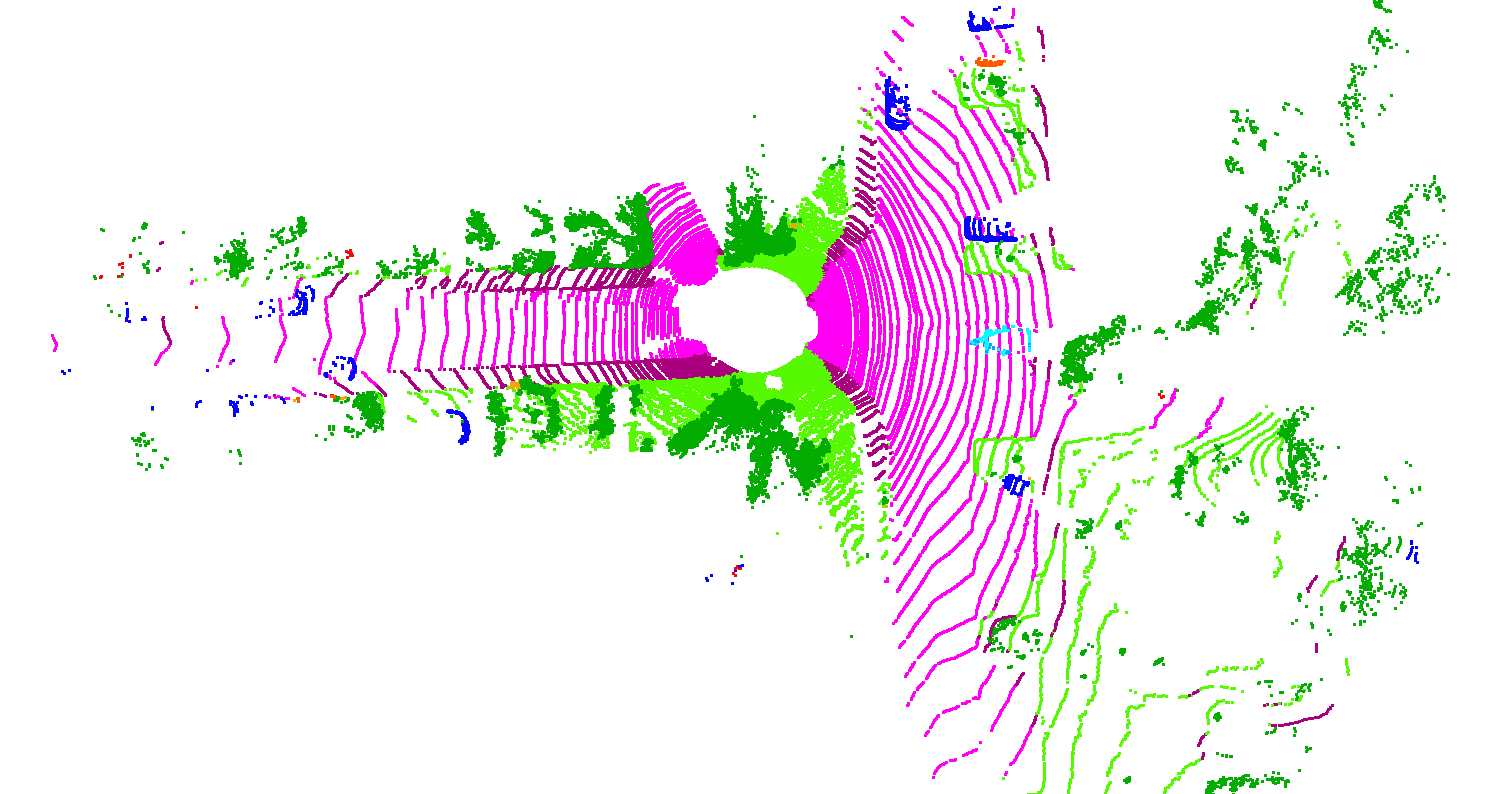}& 
\includegraphics[trim=40 25 40 0,clip,width=0.30\linewidth]{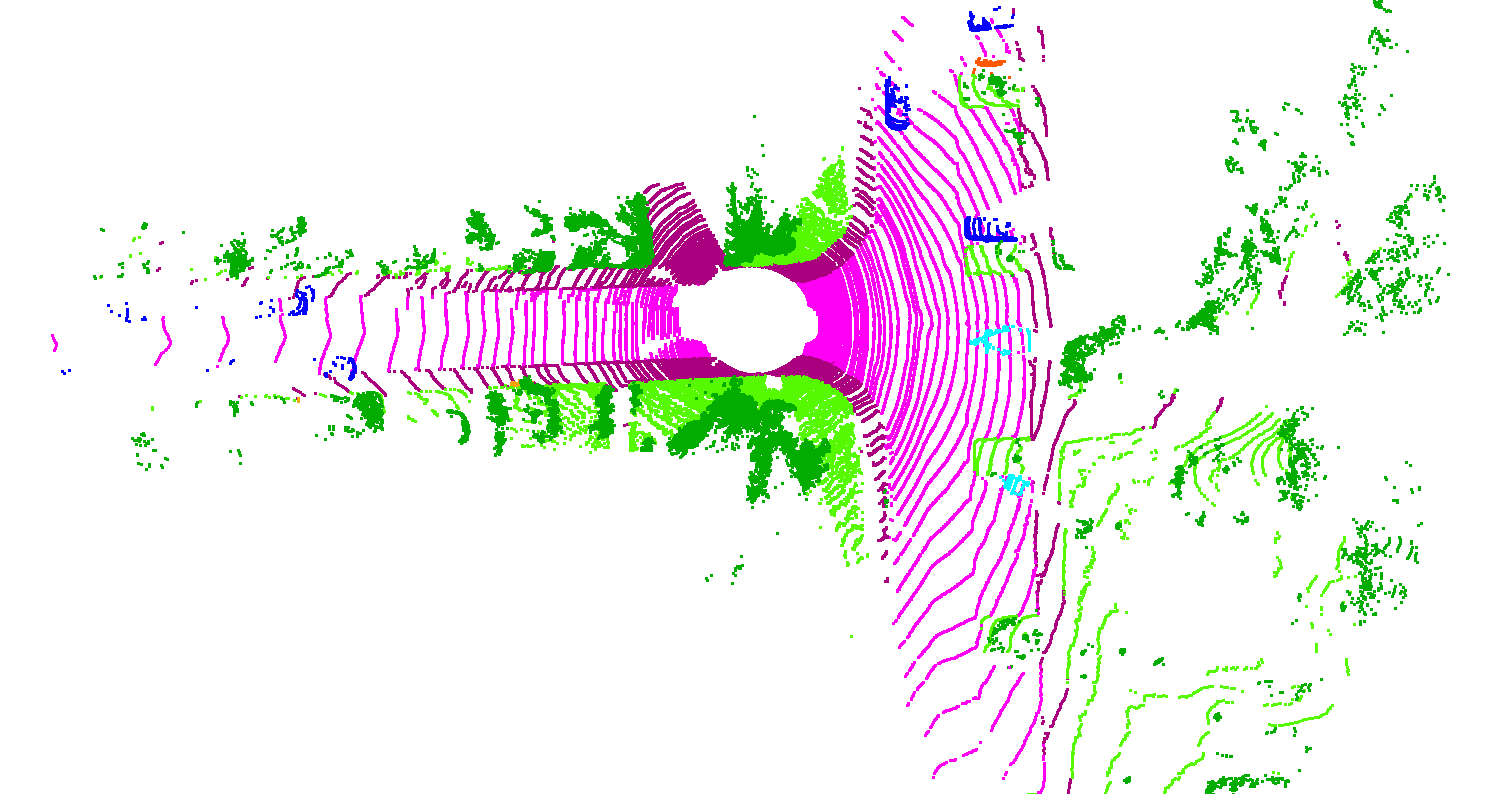}\\
  Source-only& \method{} & GT\\

 \end{tabular}

\caption{\textbf{Samples of semantic segmentation results of complete scenes in the \DAsetting{\ns}{\skns} setting} for the Source-only method and for \method{}, to compare with the ground truth (GT). The ``ignore'' class is removed for a better visualization.}
 \label{fig:app:qualitative_complete_ns_sk}
\vspace{10cm}
 \end{figure*}

For completeness, we present entire scenes segmented by \method and by the source-only method in  Figure~\ref{fig:app:qualitative_complete_ns_sk}. Even at this coarse scale, we notice that \method{} provides much better results than the source-only method, especially regarding the segmentation of the road.

\subsection{Scenes of \synthtosk}

\begin{figure*}
\newcommand{\rotext}[1]{{\begin{turn}{90}{#1}\end{turn}}}
\setlength{\tabcolsep}{1pt}
\centering
\begin{tabular}{ccc}

\begin{tikzpicture}[baseline=-16mm]
\node(a){\includegraphics[trim=40 25 40 0,clip,width=0.3\linewidth]{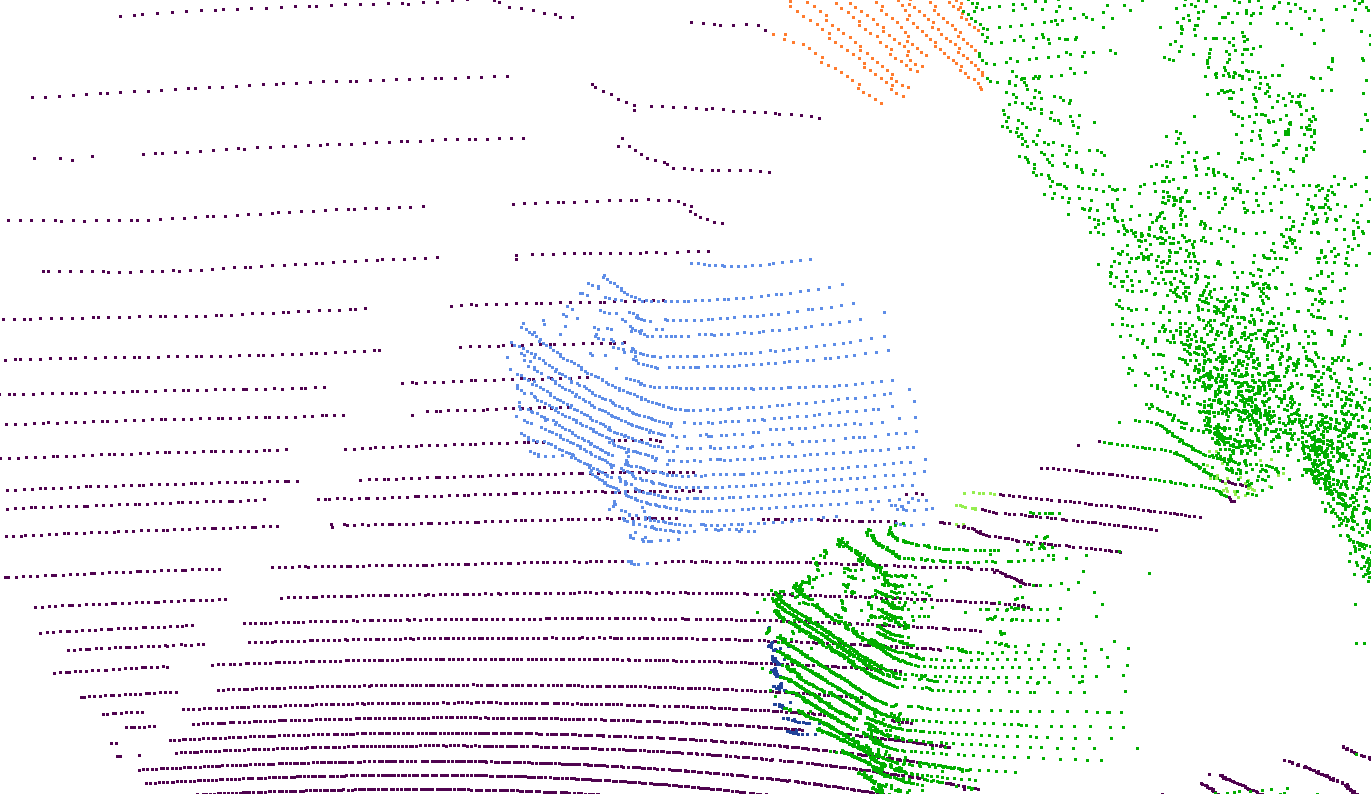}};
\node at(a.center)[draw, red,line width=1pt,ellipse, minimum width=45pt, minimum height=30pt,rotate=-7,xshift=35pt, yshift=-25pt]{}; 

\node at(a.center)[draw, red,line width=1pt,ellipse, minimum width=70pt, minimum height=25pt,xshift=-40pt, yshift=-10pt,rotate=110]{};
\end{tikzpicture} & 
\begin{tikzpicture}[baseline=-16mm]

\node(a){\includegraphics[trim=40 25 40 0,clip,width=0.3\linewidth]{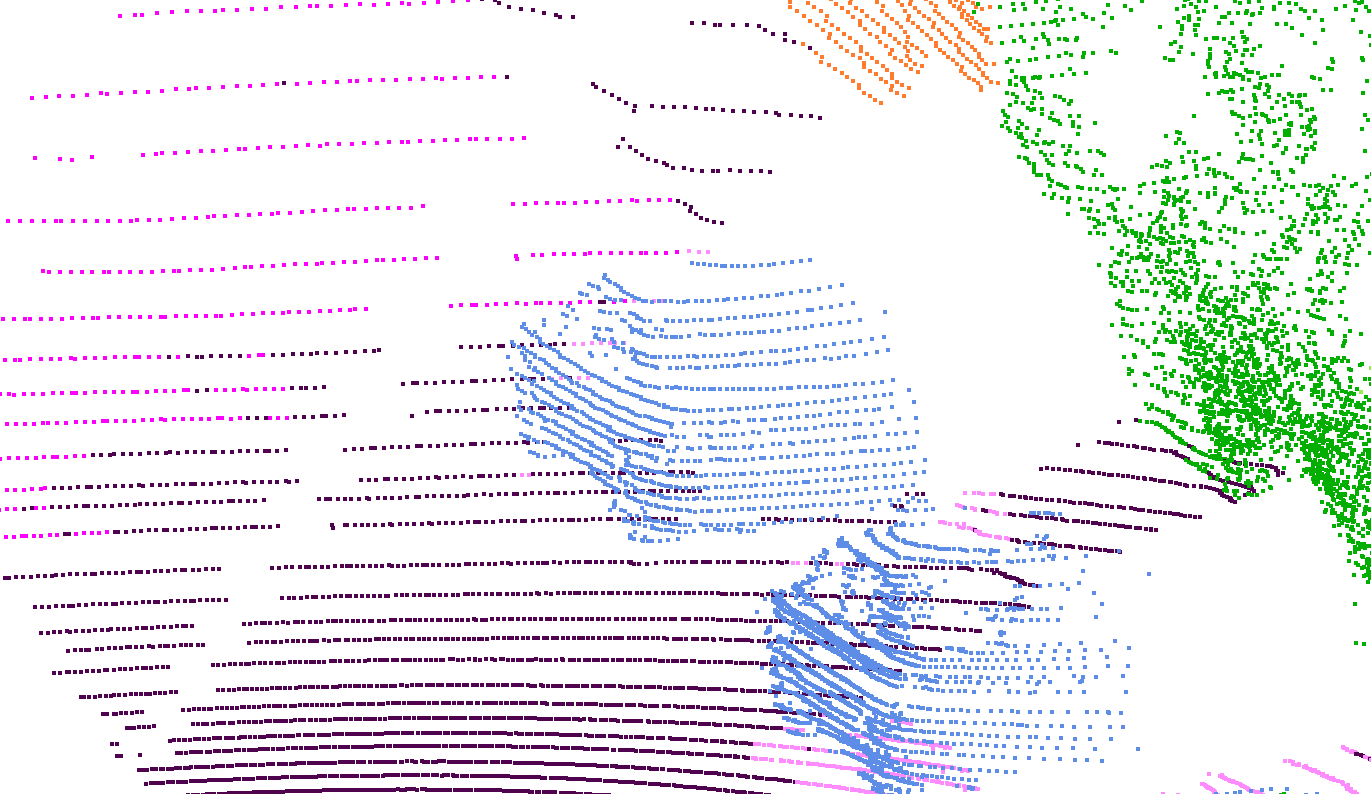}};
\node at(a.center)[draw, red,line width=1pt,ellipse, minimum width=70pt, minimum height=20pt,rotate=0,xshift=-35pt, yshift=-35pt]{}; 

\end{tikzpicture}&
\includegraphics[trim=40 25 40 0,clip,width=0.3\linewidth]{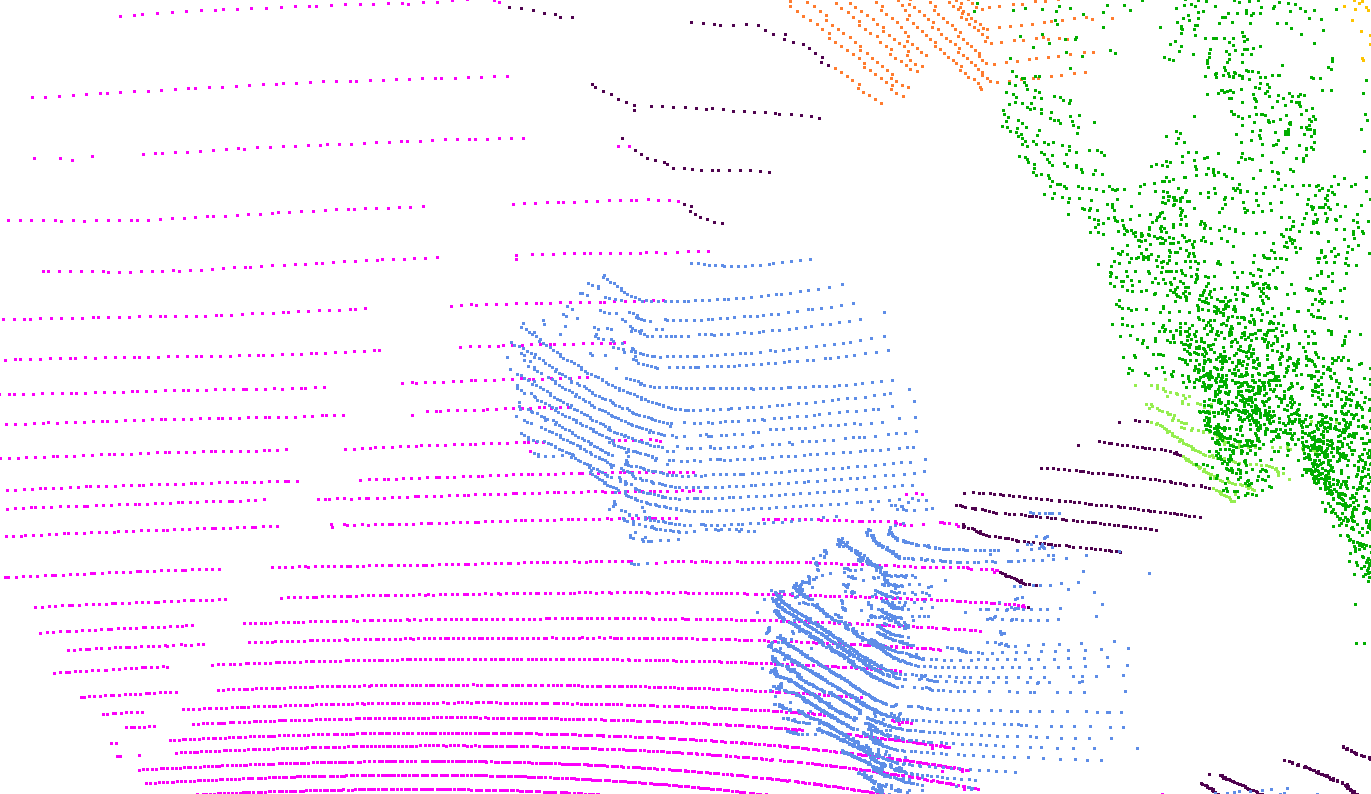}\\
Source-only & \method{} & GT\\

\begin{tikzpicture}[baseline=-14mm]
\node(a){\includegraphics[trim=40 25 40 0,clip,width=0.3\linewidth]{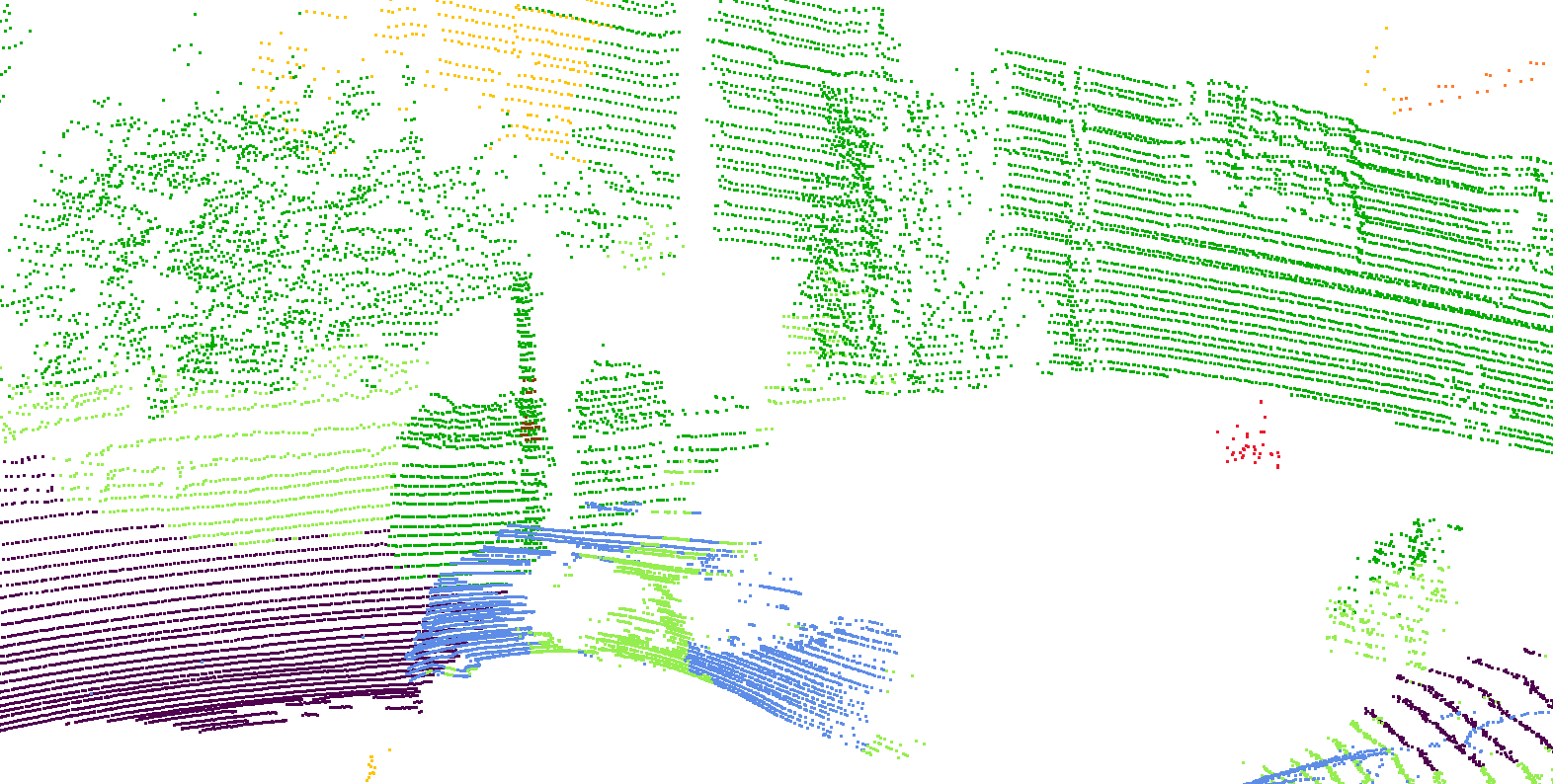}};
\node at(a.center)[draw, red,line width=1pt,ellipse, minimum width=30pt, minimum height=20pt,rotate=-7,xshift=-12pt, yshift=-28pt]{};

\node at(a.center)[draw, red,line width=1pt,ellipse, minimum width=40pt, minimum height=25pt,xshift=52pt, yshift=10pt,rotate=-5]{};

\end{tikzpicture} & 
\includegraphics[trim=40 25 40 0,clip,width=0.3\linewidth]{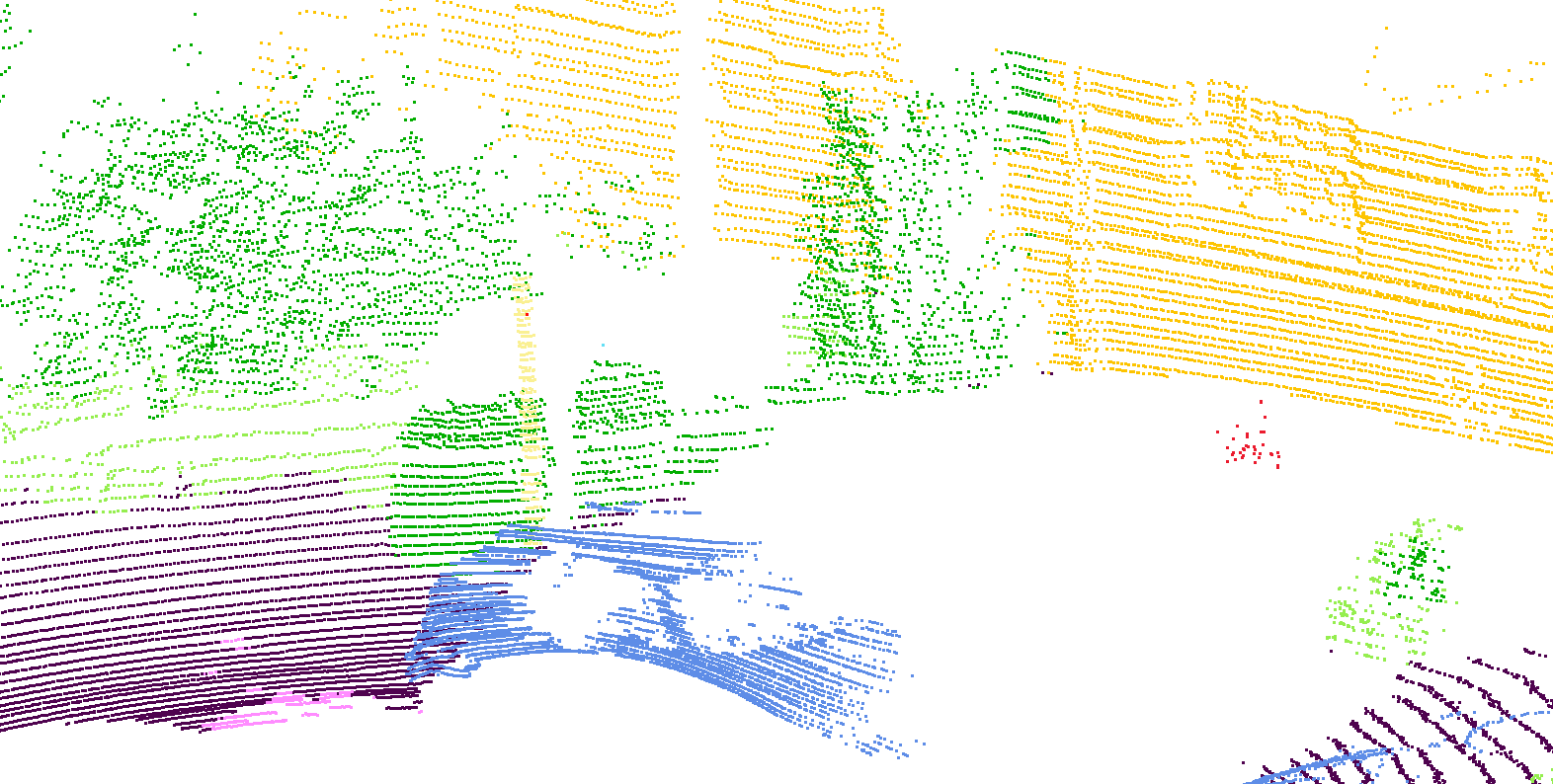}&
\includegraphics[trim=40 25 40 0,clip,width=0.3\linewidth]{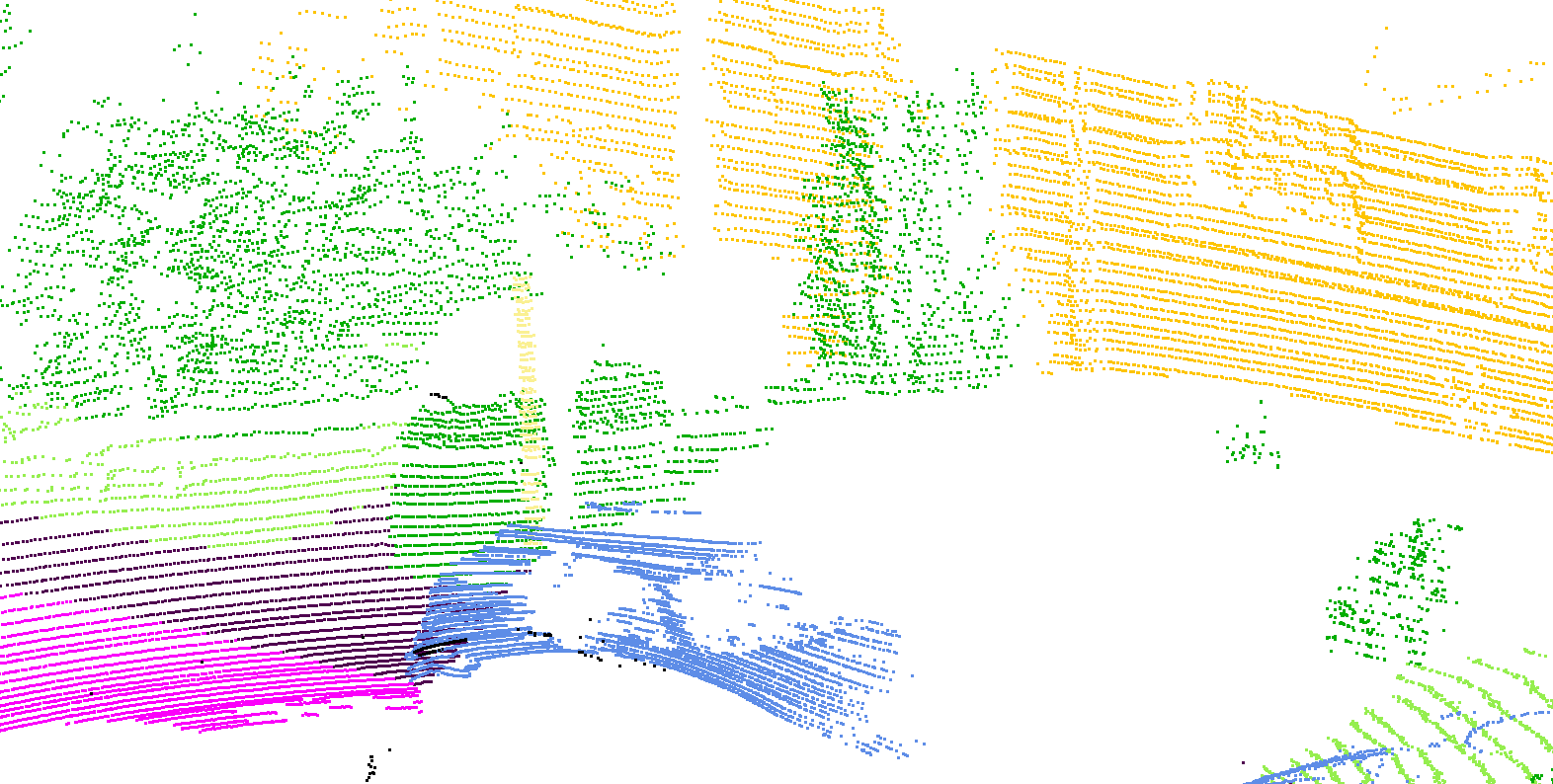}\\
Source-only & \method{} & GT\\

\begin{tikzpicture}
\node(a){\includegraphics[trim=40 25 40 0,clip,width=0.3\linewidth]{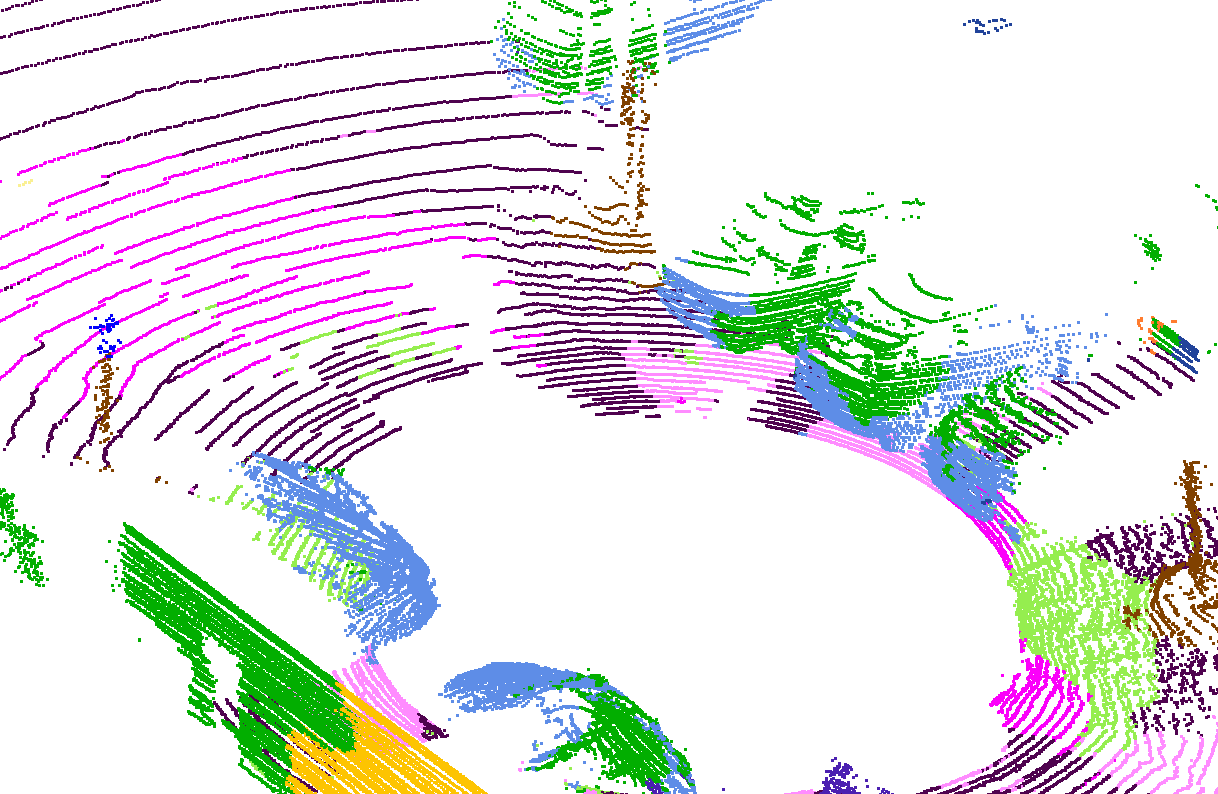}};

\node at(a.center)[draw, red,line width=1pt,ellipse, minimum width=55pt, minimum height=20pt,rotate=10,xshift=-17pt, yshift=2pt]{};

\node at(a.center)[draw, red,line width=1pt,ellipse, minimum width=45pt, minimum height=15pt,rotate=-30, xshift=23pt, yshift=21pt]{};
\end{tikzpicture} & 
\begin{tikzpicture}
\node(a){
\includegraphics[trim=40 25 40 0,clip,width=0.3\linewidth]{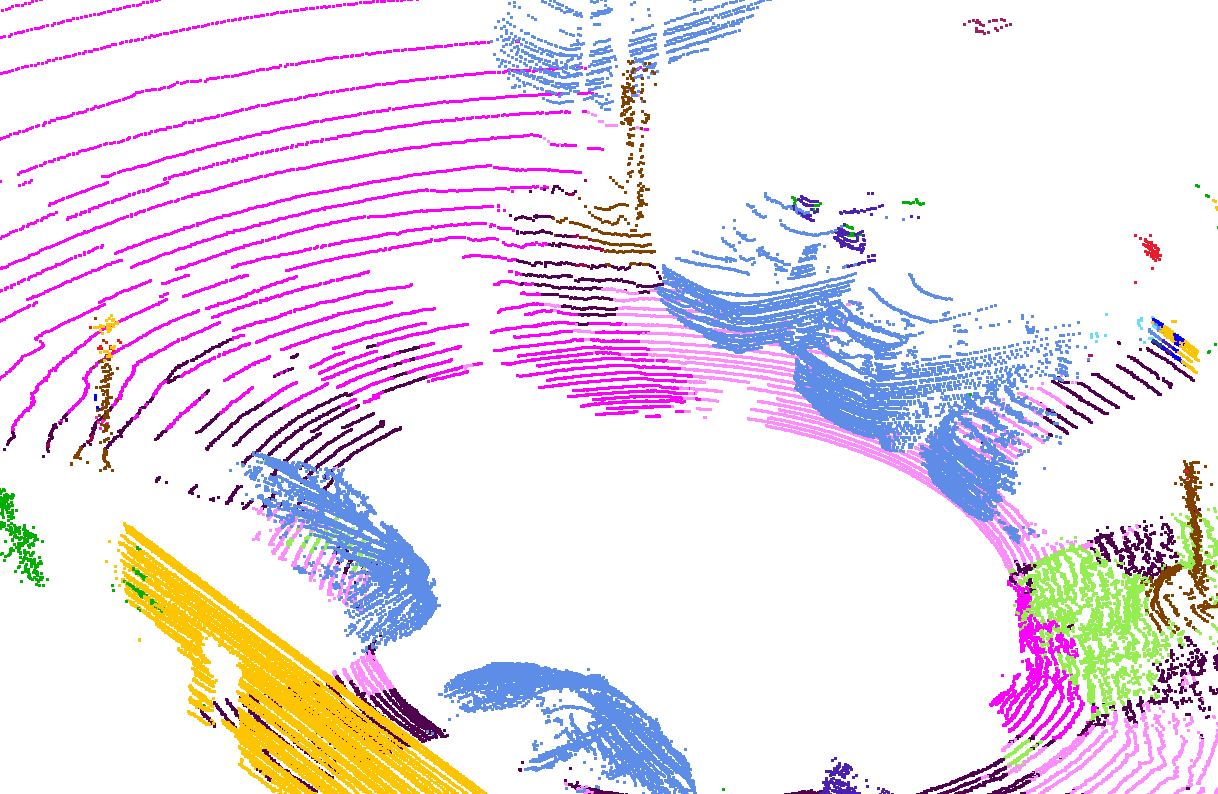}};

\end{tikzpicture} &
\includegraphics[trim=40 25 40 0,clip,width=0.3\linewidth]{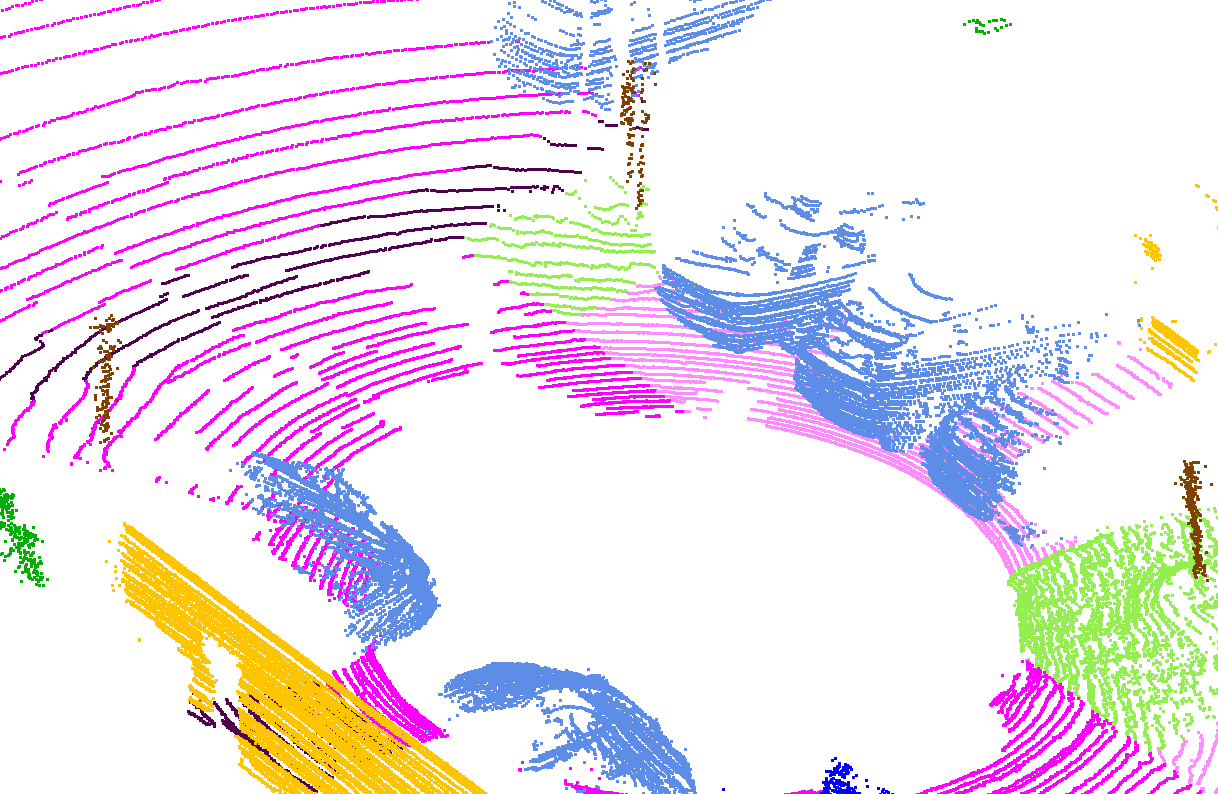}\\
Source-only & \method{} & GT\\

\begin{tikzpicture}
\node(a){\includegraphics[trim=40 25 40 0,clip,width=0.3\linewidth]{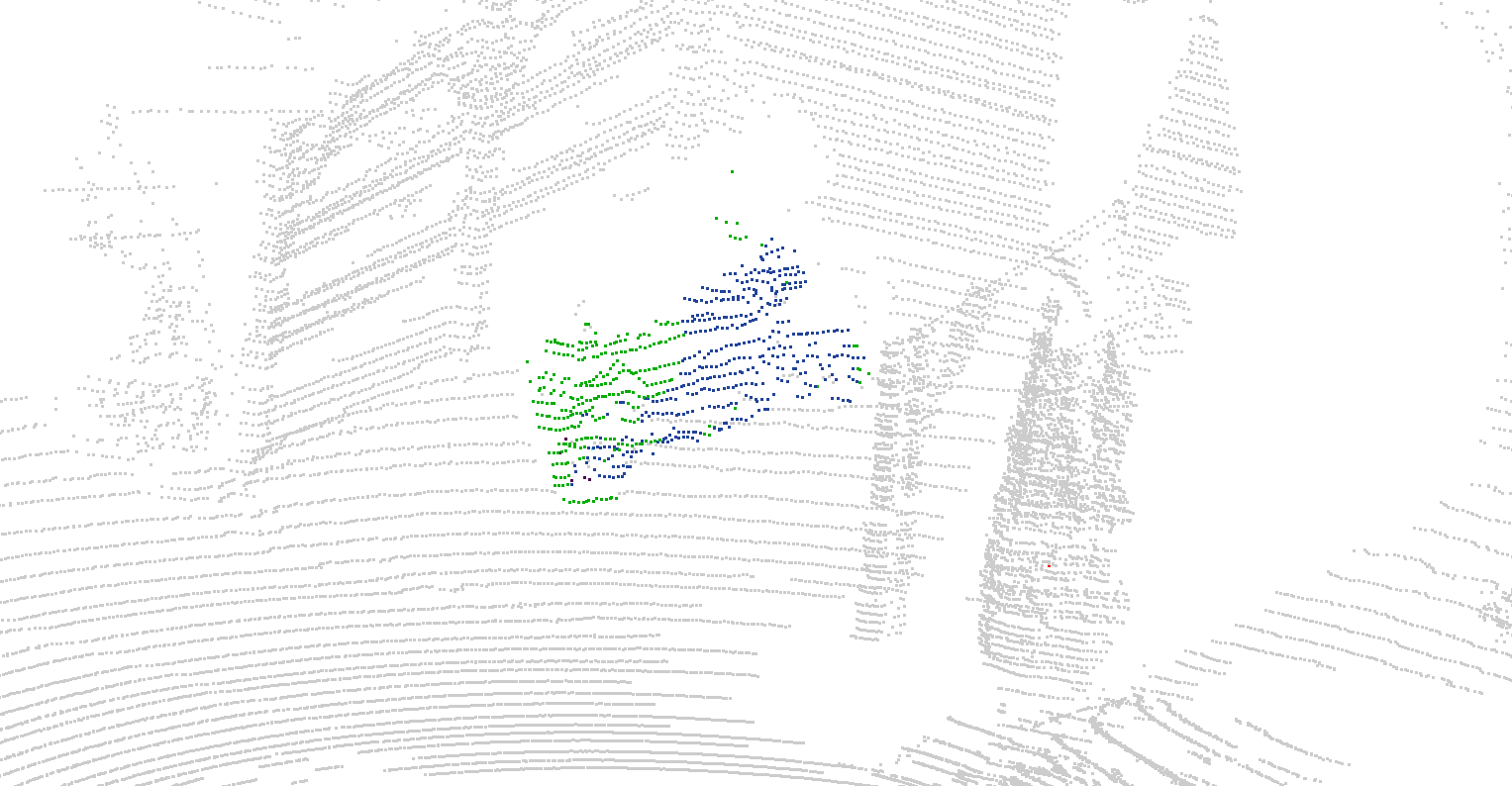}};
\node at(a.center)[draw, red,line width=1pt,ellipse, minimum width=20pt, minimum height=25pt, rotate=-13, xshift=-15pt, yshift=-9pt]{};
\end{tikzpicture} &

\begin{tikzpicture}
\node(a){\includegraphics[trim=40 25 40 0,clip,width=0.3\linewidth]{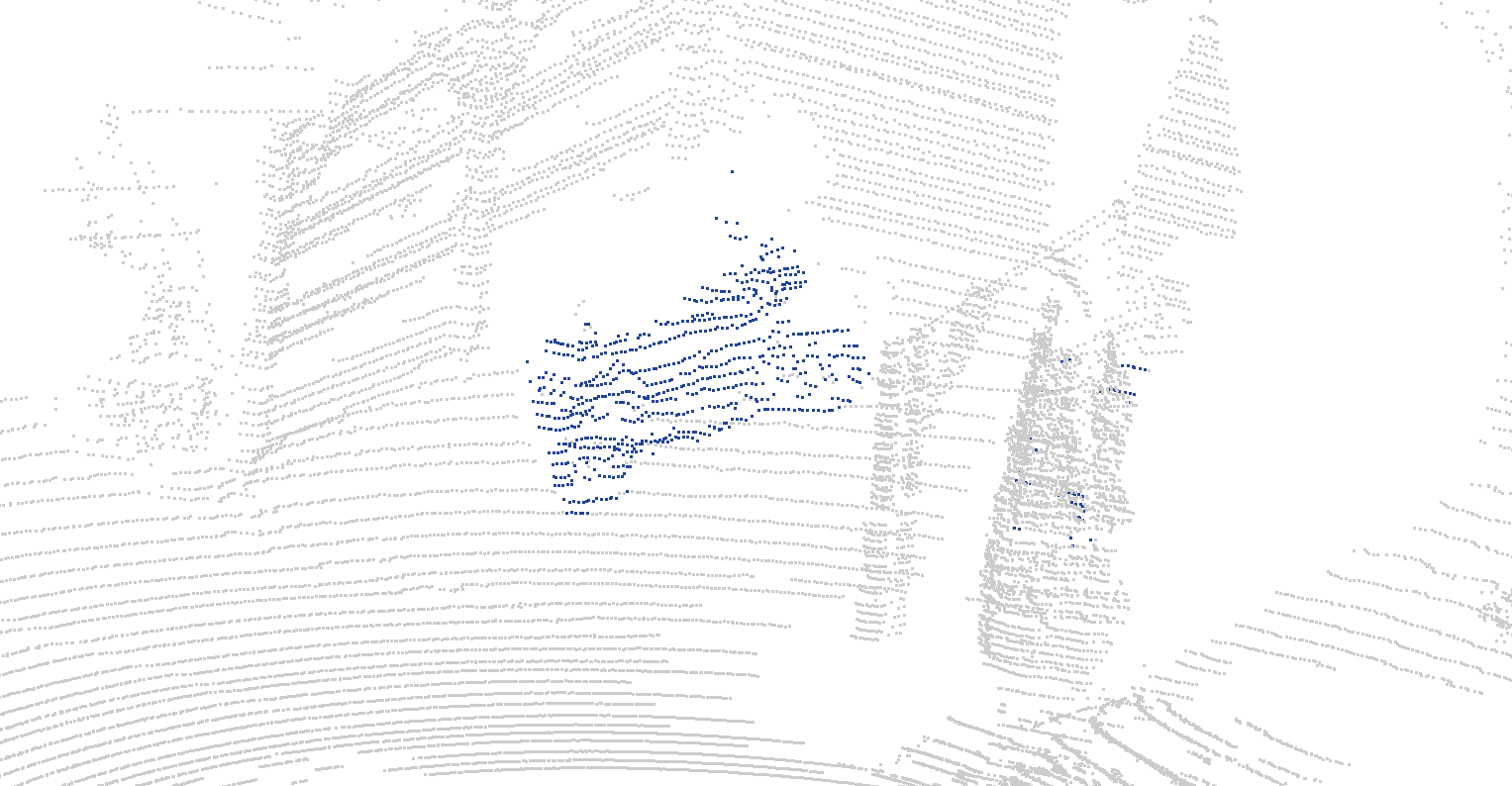}};
\end{tikzpicture}
&
\includegraphics[trim=40 25 40 0,clip,width=0.3\linewidth]{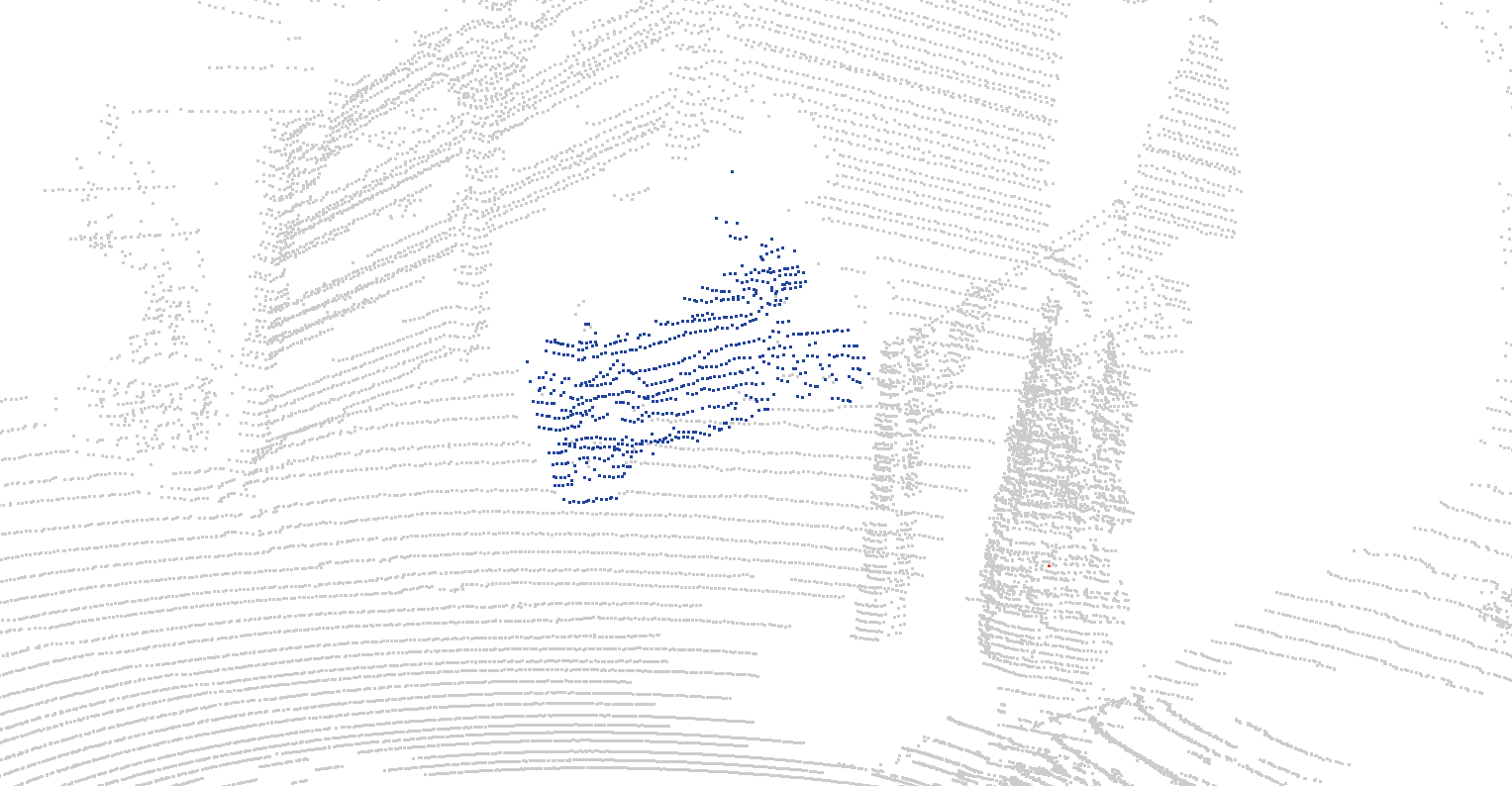}\\
Source-only & \method{} & GT\\
(non-motorcycle points are grey) & (non-motorcycle points are grey) & (non-motorcycle points are grey)\\

\begin{tikzpicture}[baseline=-16mm]
\node(a){\includegraphics[trim=40 25 40 0,clip,width=0.3\linewidth]{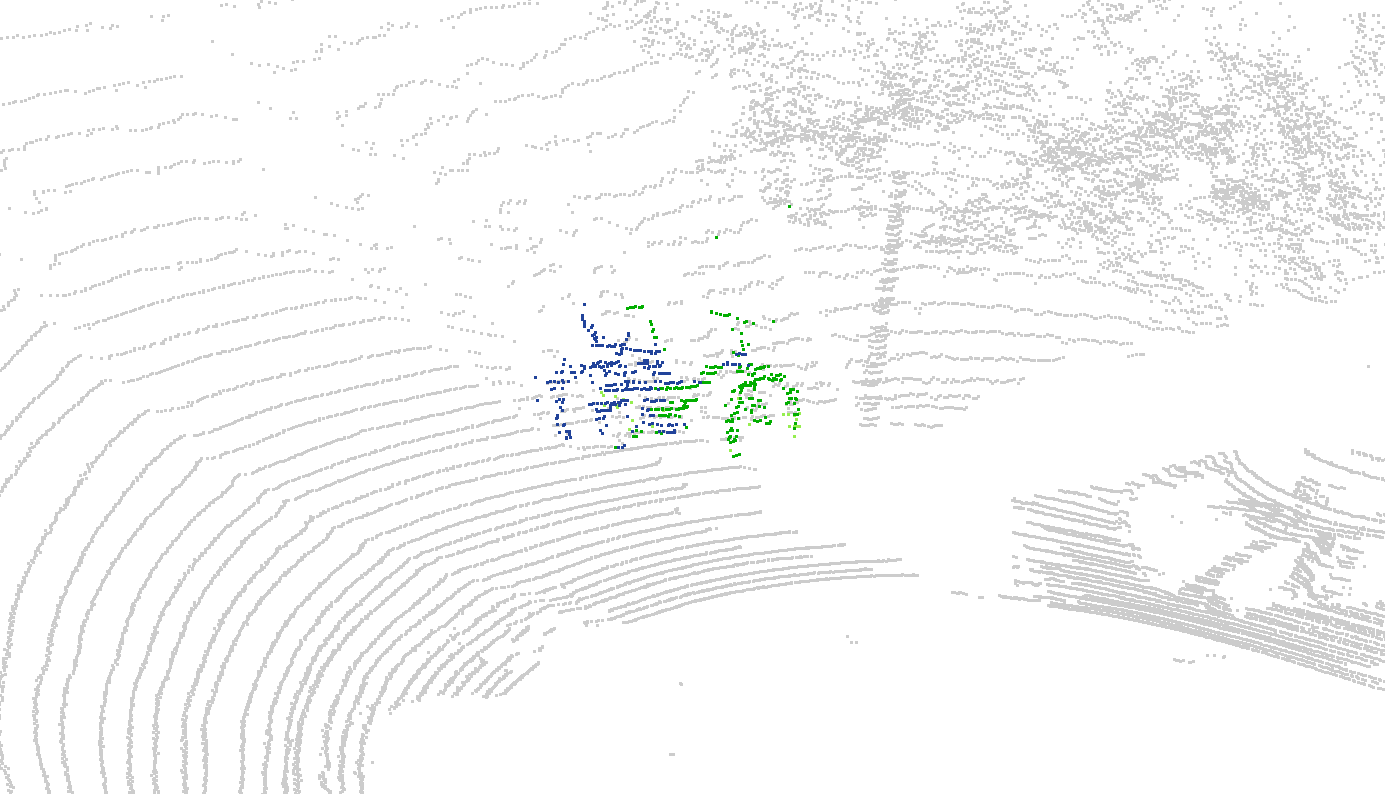}};
\node at(a.center)[draw, red,line width=1pt,ellipse, minimum width=35pt, minimum height=22pt,rotate=-13, xshift=-1pt, yshift=-2pt]{};
\end{tikzpicture} & 

\begin{tikzpicture}[baseline=-16mm]
\node(a){\includegraphics[trim=40 25 40 0,clip,width=0.3\linewidth]{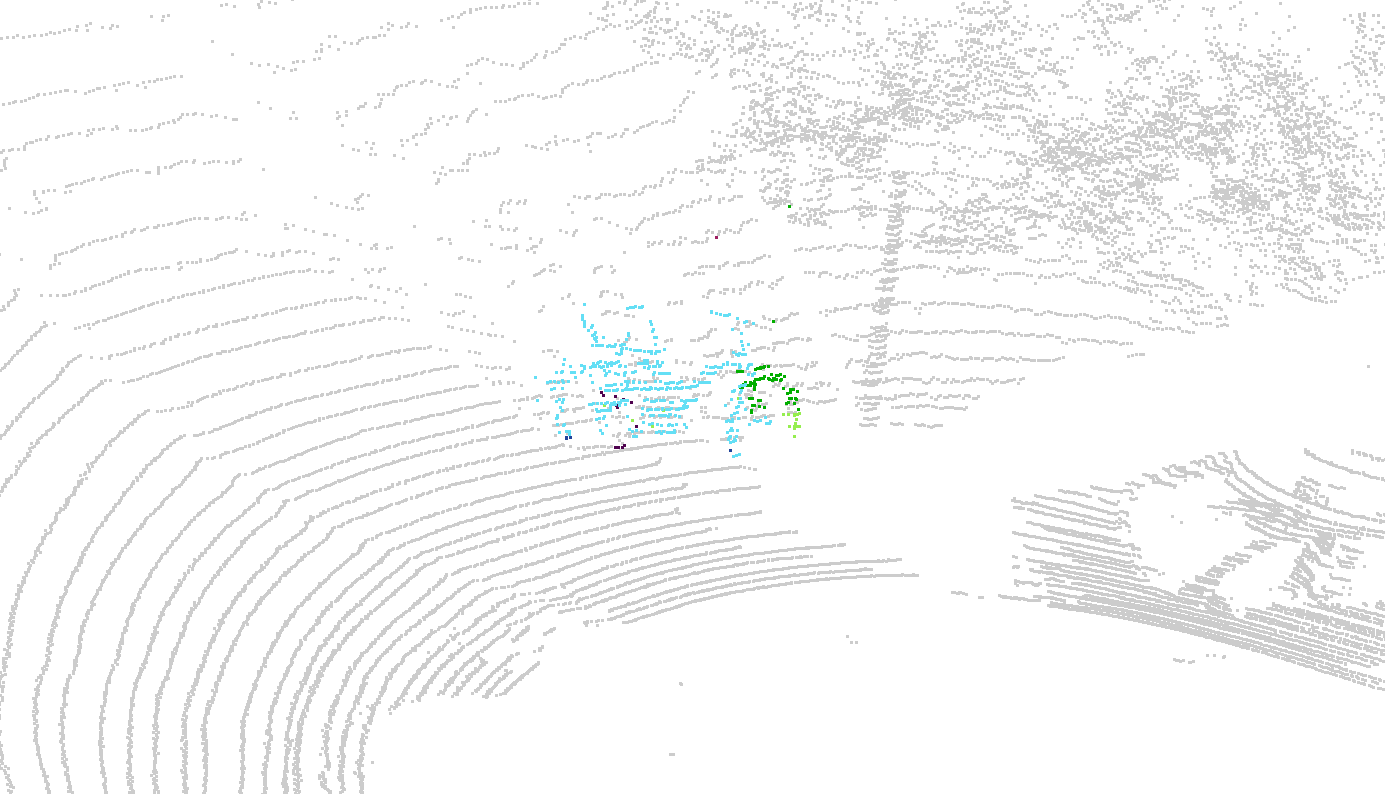}};
\node at(a.center)[draw, red,line width=1pt, ellipse, minimum width=9pt, minimum height=4pt,rotate=-70,xshift=9pt, yshift=9pt, inner sep=0pt]{};
\end{tikzpicture}
&
\includegraphics[trim=40 25 40 0,clip,width=0.3\linewidth]{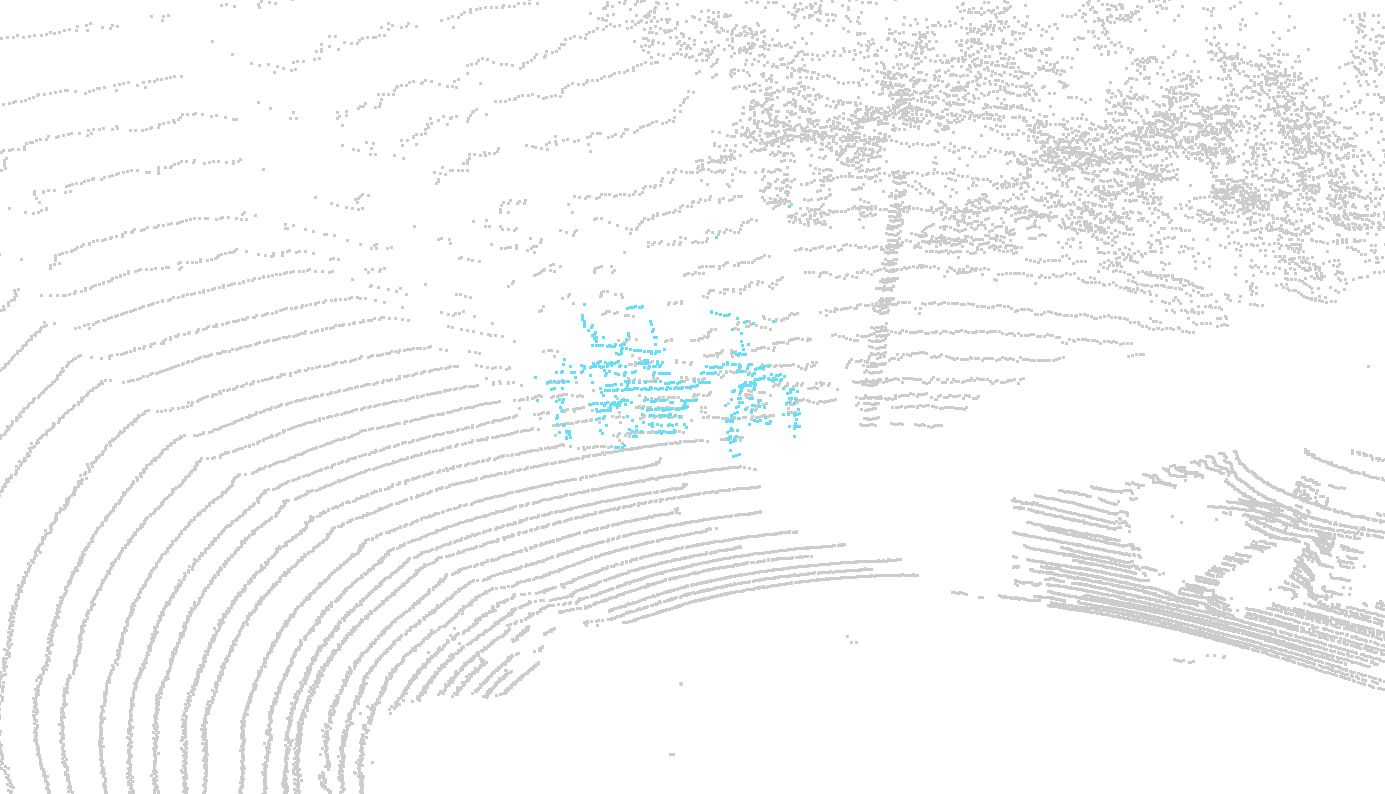}

\\
Source-only & \method{} & GT \\
(non-bicycle points are grey) & (non-bicycle points are grey) & (non-bicycle points are grey)\\

\end{tabular}
\caption{
\textbf{Samples of semantic segmentation results in the \DAsetting{SynL}{\sksyn} setting} for the Source-only method and for \method{}, to compare with the ground truth (GT). The red circles highlight wrong segmentations.}
\label{fig:app:qualitative_zoom_syn_sk}
\vspace{0.5cm}
\end{figure*}

In Figure~\ref{fig:app:qualitative_zoom_syn_sk}, we compare in the \DAsetting{\synth}{\sksyn} setting results obtained with \method{} with results obtained with the source-only model. In the examples on the first three rows, we notice that \method segments correctly the cars that are the closest to the sensor, while the source-only model can miss them even entirely (first row). The example on the last two rows show cases where \method{} is able to segment correctly the motorcycle and the bicycle despite their small size.

\begin{figure*}
\newcommand{\rotext}[1]{{\begin{turn}{90}{#1}\end{turn}}}
\setlength{\tabcolsep}{1pt}
\centering
\begin{tabular}{ccc}

\begin{tikzpicture}
\node(a){\includegraphics[trim=40 25 40 10,clip,width=0.3\linewidth]{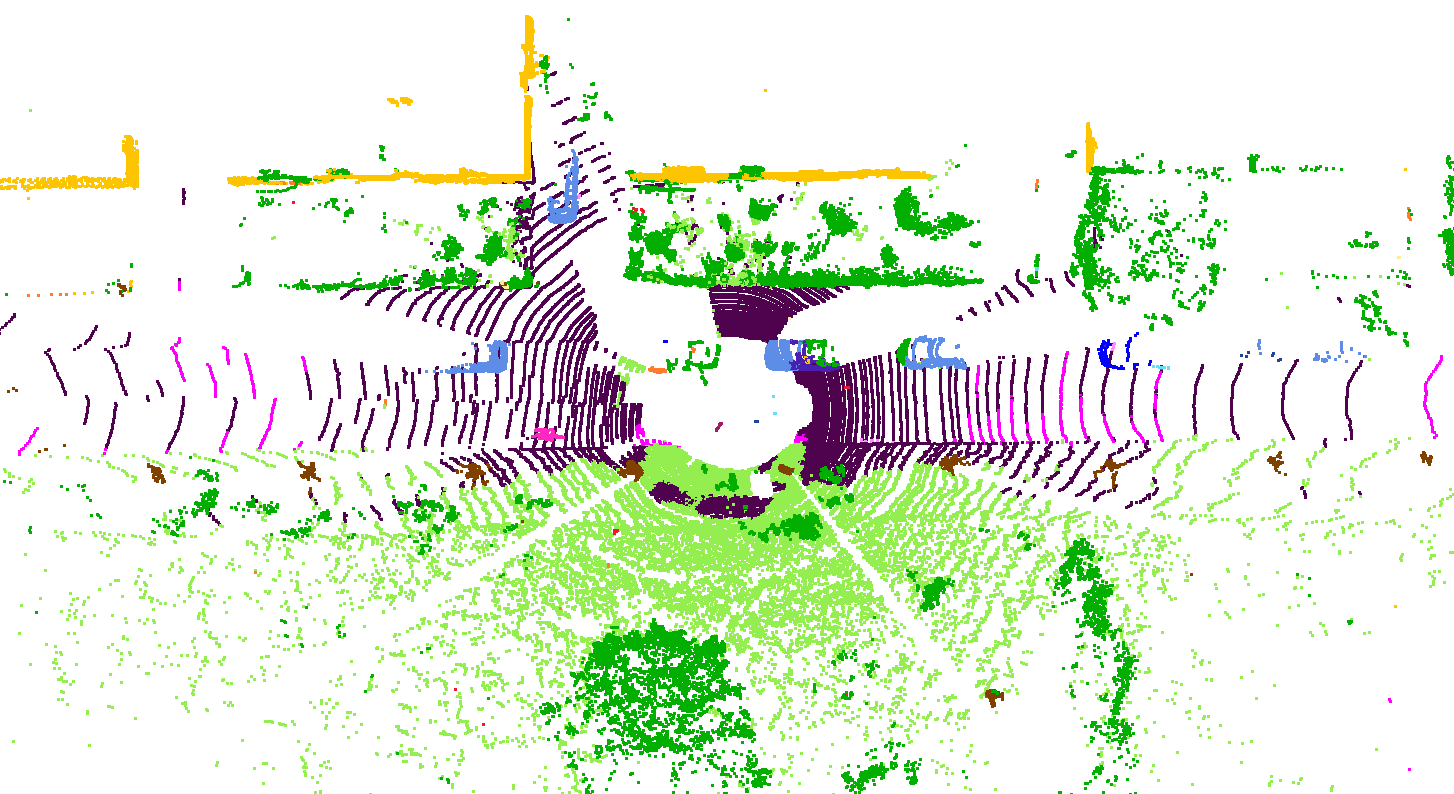}};
\end{tikzpicture} & 
\includegraphics[trim=40 25 40 10,clip,width=0.3\linewidth]{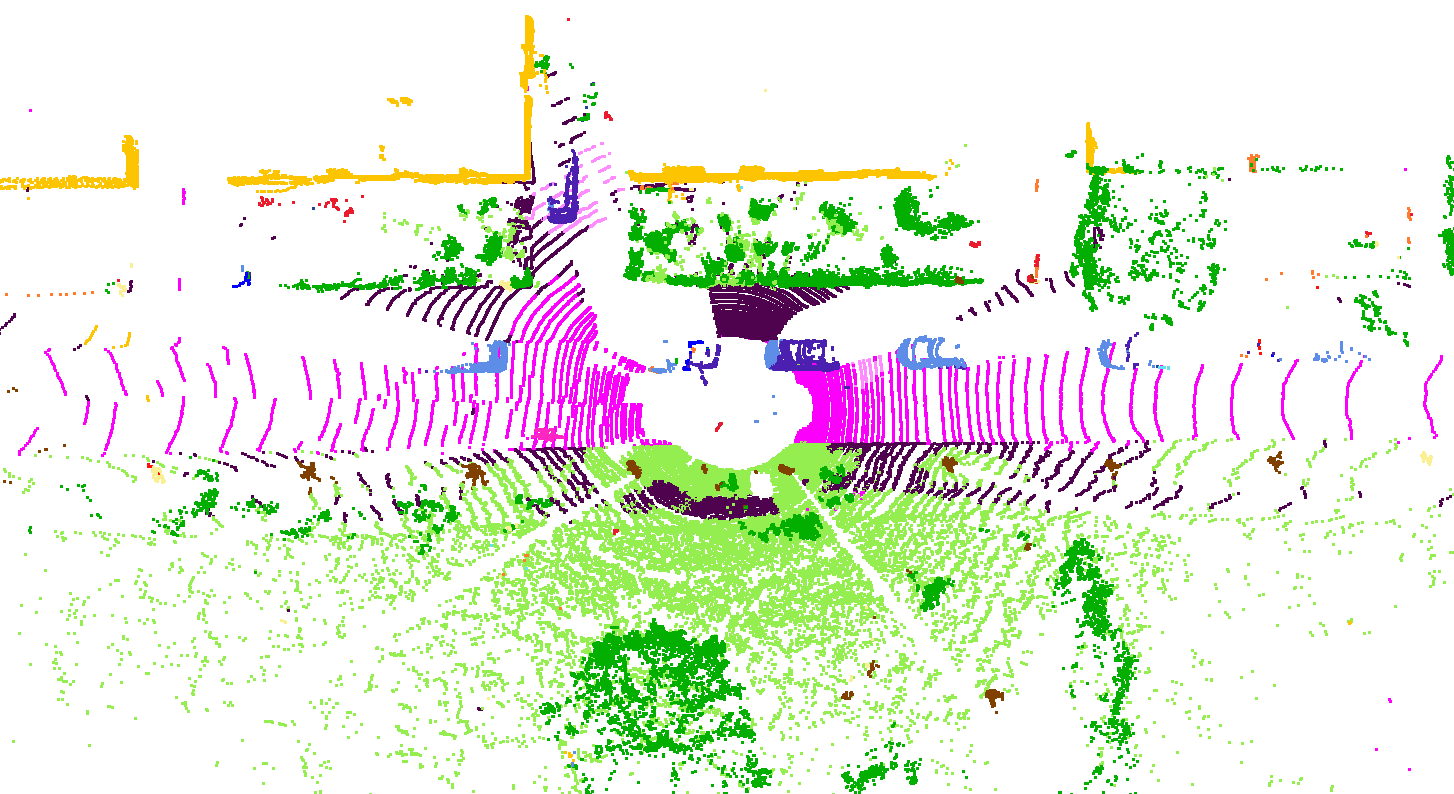}&
\includegraphics[trim=40 25 40 10,clip,width=0.3\linewidth]{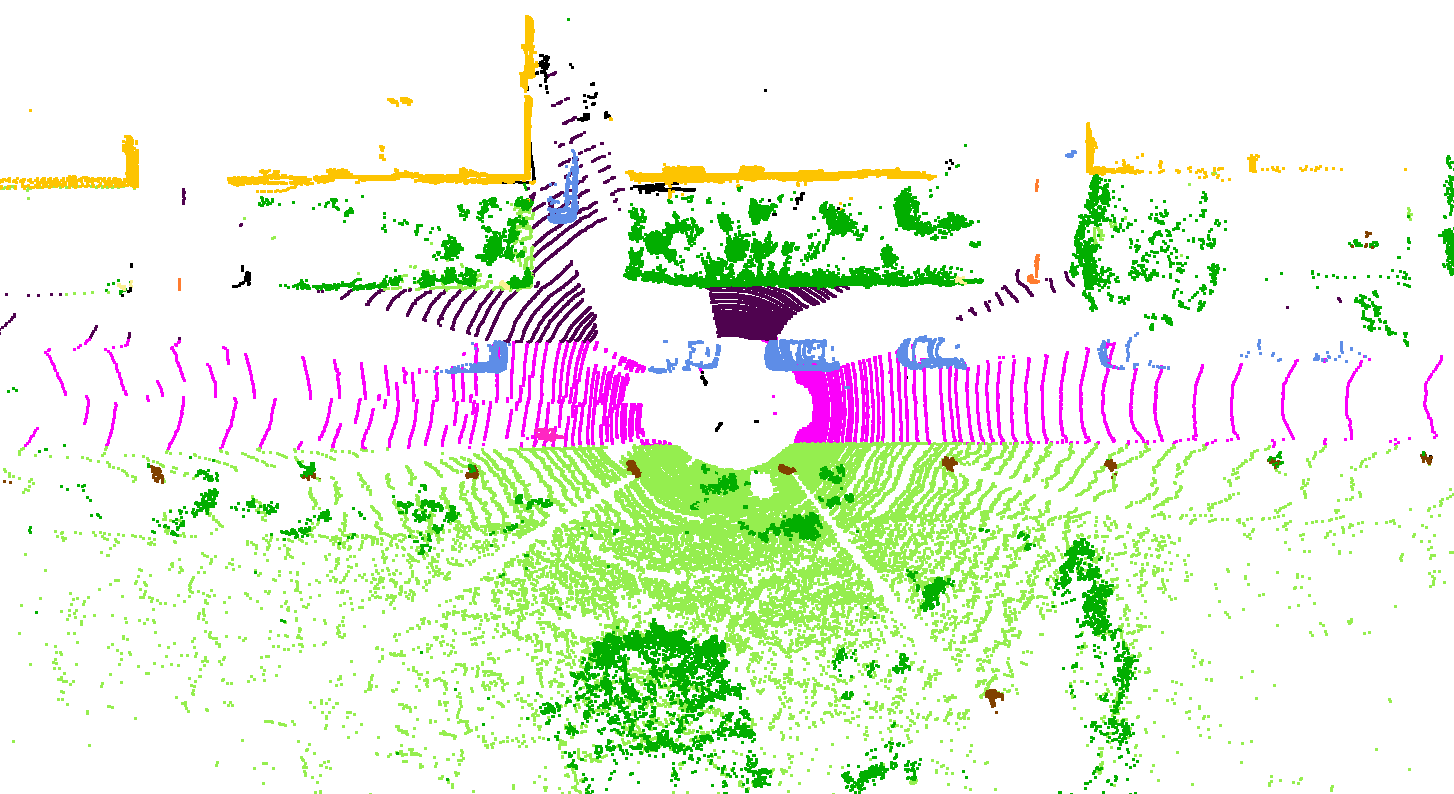}\\
Source-only & \method{} & GT\\

\begin{tikzpicture}
\node(a){\includegraphics[trim=40 25 40 10,clip,width=0.3\linewidth]{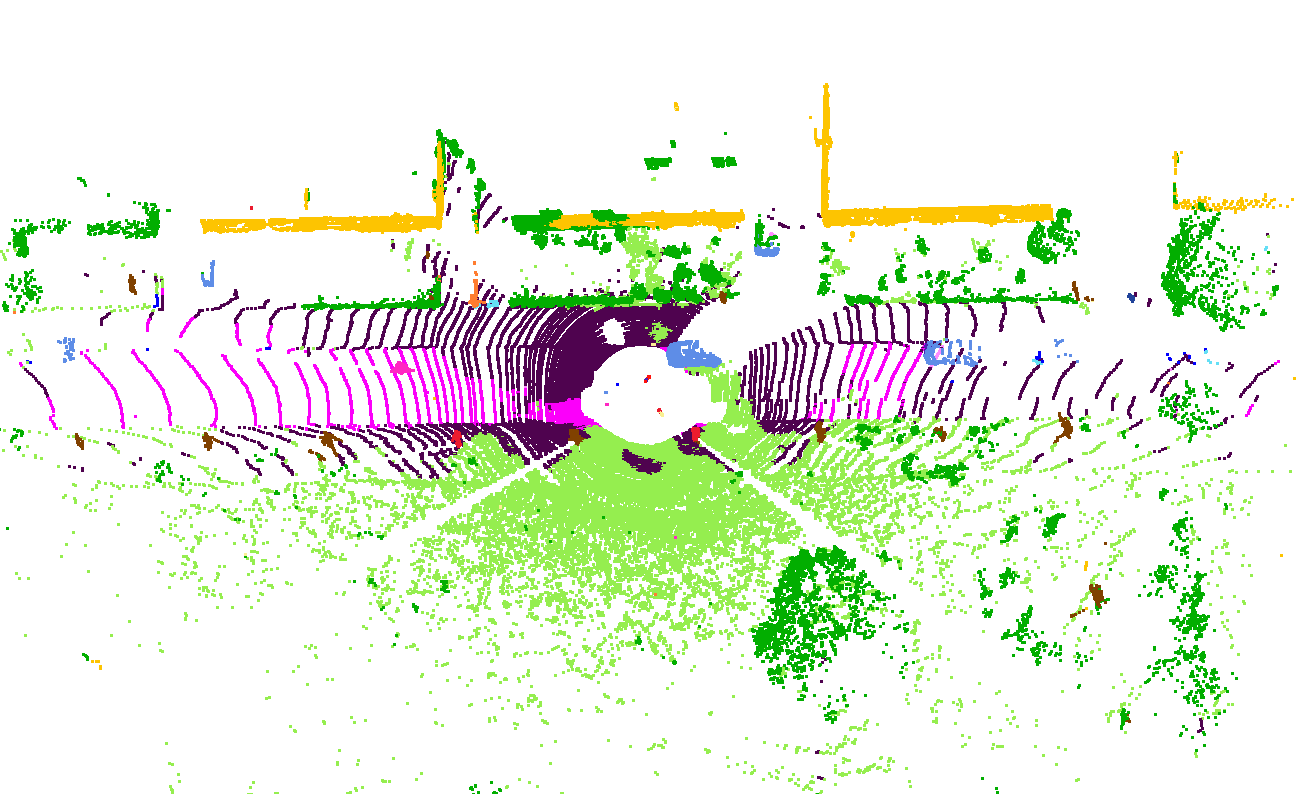}};
\end{tikzpicture} & 
\includegraphics[trim=40 25 40 10,clip,width=0.3\linewidth]{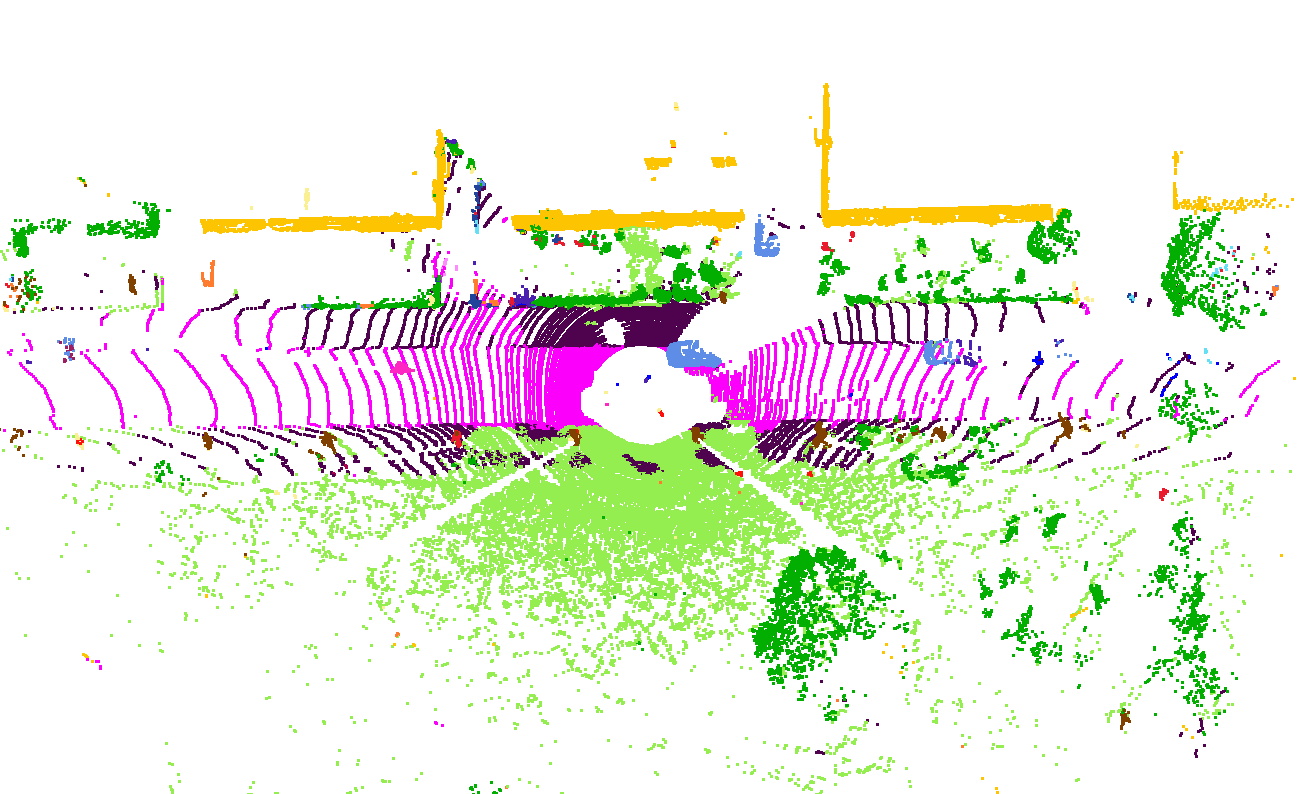}&
\includegraphics[trim=40 25 40 10,clip,width=0.3\linewidth]{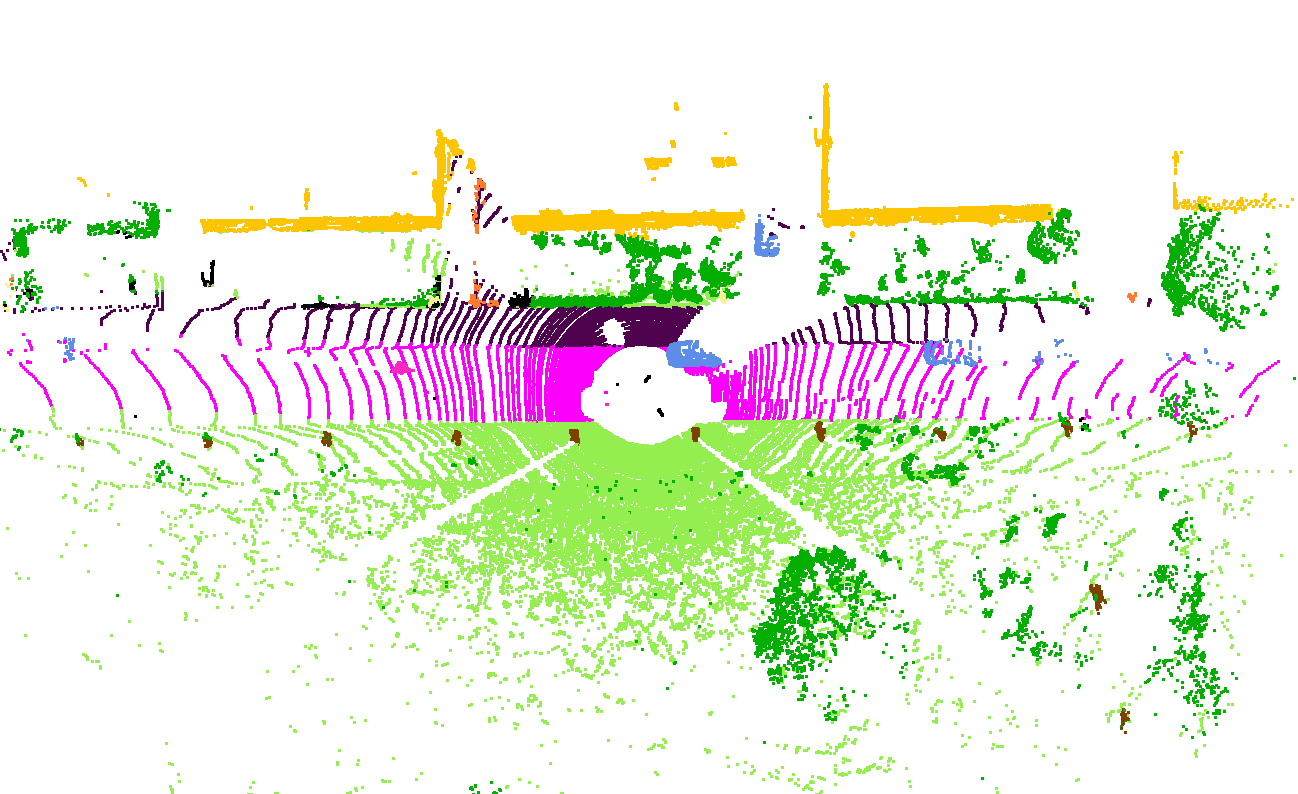}\\
Source-only & \method{} & GT\\

\begin{tikzpicture}
\node(a){\includegraphics[trim=40 25 40 10,clip,width=0.3\linewidth]{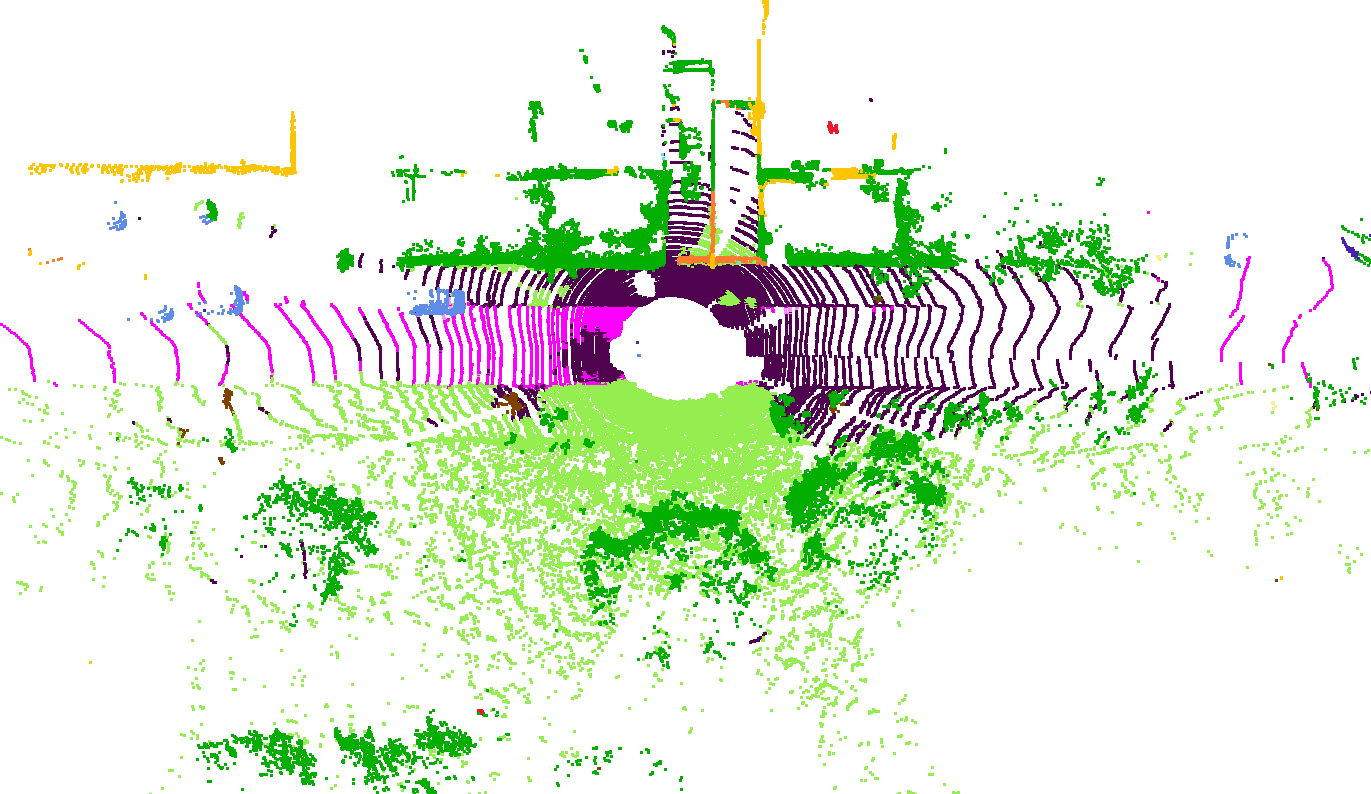}};
\end{tikzpicture} & 
\includegraphics[trim=40 25 40 10,clip,width=0.3\linewidth]{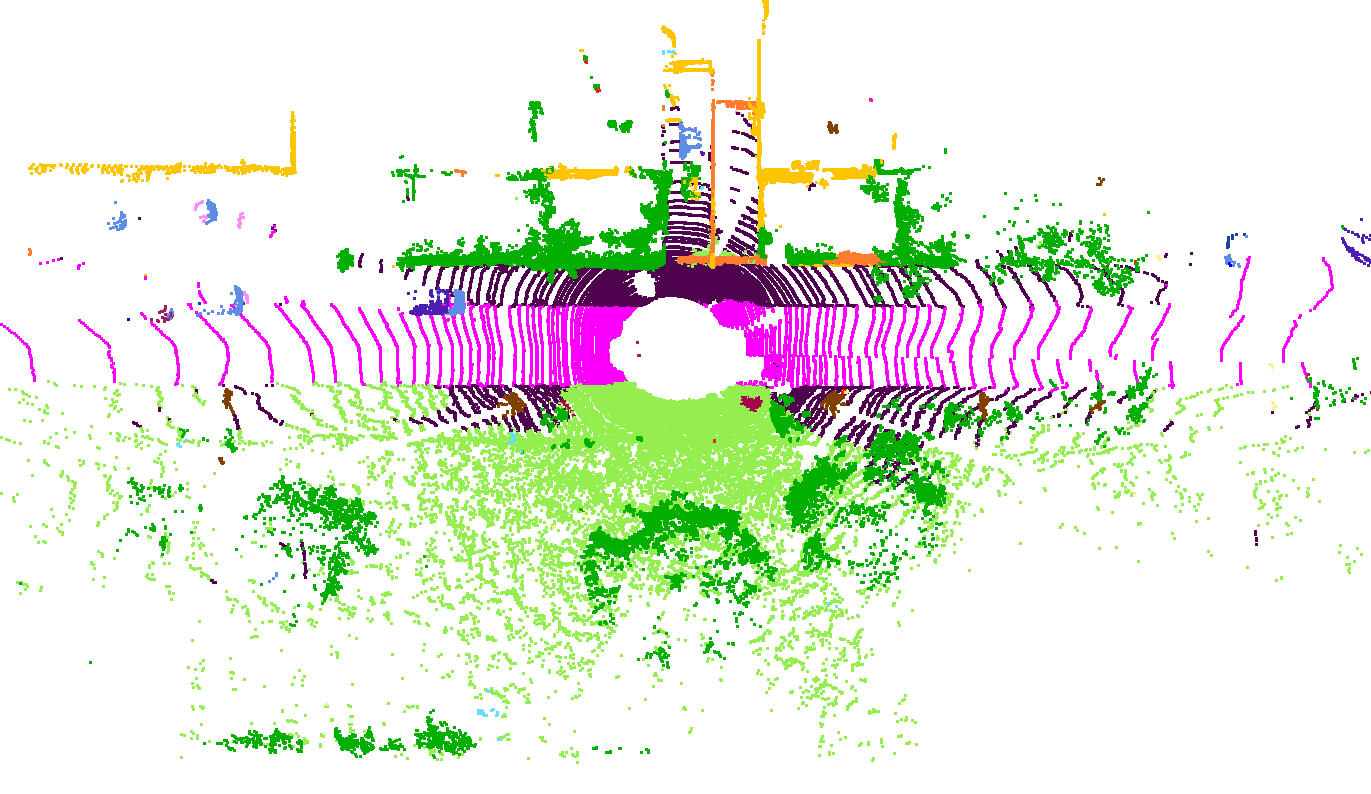}&
\includegraphics[trim=40 25 40 10,clip,width=0.3\linewidth]{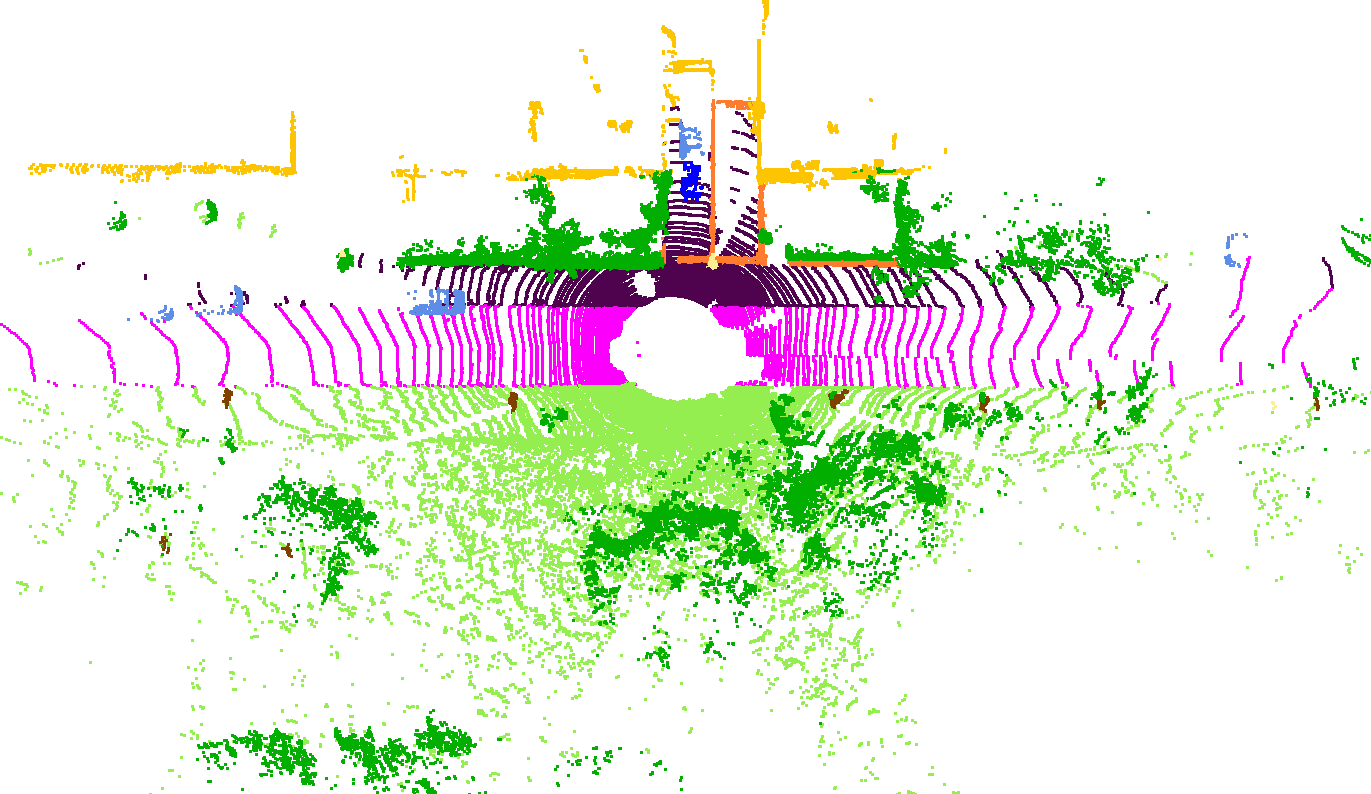}\\
Source-only & \method{} & GT\\

\end{tabular}

\caption{\textbf{Samples of semantic segmentation results of complete scenes in the \DAsetting{SynL}{\sksyn} setting} for the Source-only method and for \method{}, to compare with the ground truth (GT). The ``ignore'' class is removed for a better visualization.}
\label{fig:app:qualitative_complete_syn_sk}
\vspace{8.5cm}
\end{figure*}

As above, for completeness, we present in Figure~\ref{fig:app:qualitative_complete_syn_sk} entire scenes segmented by \method and by the source-only method. We notice that \method{} distinguishes \colorbox{sksidewalk}{\color{white}"sidewalk"} and \colorbox{skdrivable}{\color{white}"road"} better than the source-only model.

\begin{figure}[!t]
\newcommand{\rotext}[1]{{\begin{turn}{90}{#1}\end{turn}}}
\setlength{\tabcolsep}{1pt}
\centering
\begin{tabular}{ccc}

& Source-only & \method{} \\
\rotatebox{90}{\quad \enspace \DAsetting{NS}{\skns}} & 
\includegraphics[trim=40 25 40 0,clip,width=0.45\linewidth,]{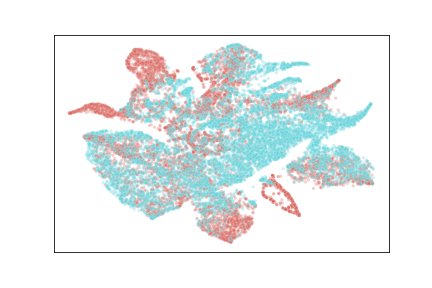}& 
\includegraphics[trim=40 25 40 0,clip,width=0.45\linewidth]{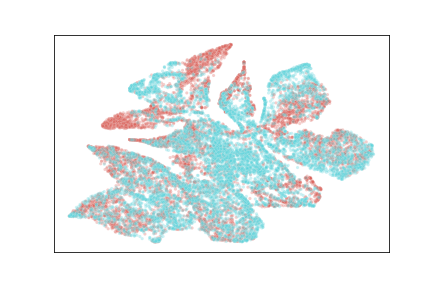}\\

\rotatebox{90}{\enspace\: \DAsetting{SynL}{\sksyn} } & 
\includegraphics[trim=40 25 40 0,clip,width=0.45\linewidth]{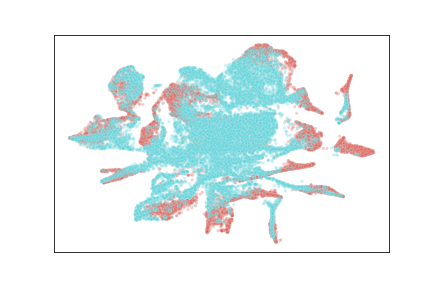}& 
\includegraphics[trim=40 25 40 0,clip,width=0.45\linewidth]{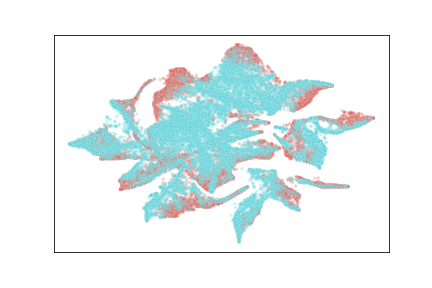}

\end{tabular}

\caption{
\textbf{\tsne\ visualisations of the latent space structure} for the source-only method and \method{}, in the  \DAsetting{NS}{\skns} and \DAsetting{SynL}{\sksyn} settings. Colors:\colorbox{tsne_source}{\color{white}Source points}, \colorbox{tsne_target}{\color{white}Target points}.
}
\label{fig:tsne_overlap}
\end{figure}

\subsection{\tsne\ representations}
Thanks to \tsne\cite{van2008visualizing} visualizations, we analyze qualitatively the structure of the learned latent space at the output of the backbone. In order to generate the \tsne\ visualizations, we select randomly 200 scenes of the validation set, sample randomly 1000 points for each of the classes, collect the corresponding $128$-dimensional latent vector $z$, and reduce their dimension to $2$ using \tsne. For the setting \DAsetting{\synth}{\sksyn}, we do not take into account the classes "parking" and "other ground" as they are too rare in the selected scenes to produce any useful visualizations.

\paragraph{Global source and target visualization.}
\tsne\ visualizations of the structure of the source and target latent spaces are presented in Figure~\ref{fig:tsne_overlap}, for the source-only method and \method{}, and for both \nstosk{} and \synthtosk. In all cases, there is a notable overlap between the structure of the source latent space and that of the target space, especially for \nstosk{}. For \synthtosk, the latent space for the target domain is however more spread than the latent space for the source domain, whose features are likely more concentrated thanks to the full supervision. It is consistent with the lower success of the unsupervised domain adaptation in this setting (cf.\ Table~2 of main paper), compared to \nstosk{} (cf.\ Table~1 of main paper). Nevertheless, {\method} features show a little more overlap than source-only features, which is reflected in quantitative results, where {\method} significantly outperforms source-only.

We continue below with a class-wise visualization to further analyze this latent space structure and search for differences between both methods.

\paragraph{Per-class visualization.}

\begin{figure*}
\small
\newcommand{\rotext}[1]{{\begin{turn}{90}{#1}\end{turn}}}
    \setlength{\tabcolsep}{1pt}
    \centering
    \begin{tabular}{ccc@{}c@{}c@{}c@{}c}
    \multirow{2}*{\rotext{ \textbf{Source-only}}} &
     \rotext{\quad \enspace Source}& 
       \includegraphics[trim=40 25 40 0,clip,width=0.19\linewidth]{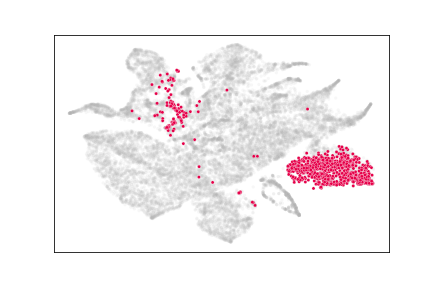}& 
        \includegraphics[trim=40 25 40 0,clip,width=0.19\linewidth]{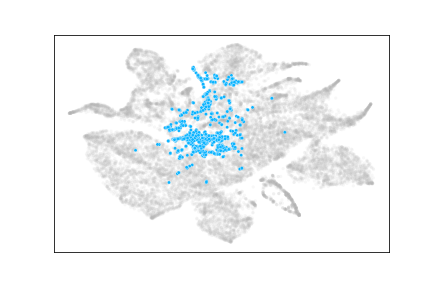}&
        \includegraphics[trim=40 25 40 0,clip,width=0.19\linewidth]{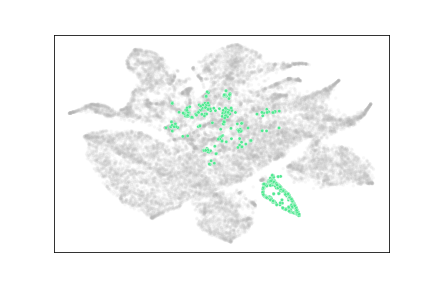}&
        \includegraphics[trim=40 25 40 0,clip,width=0.19\linewidth]{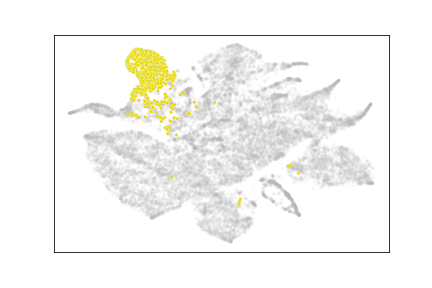}&
        \includegraphics[trim=40 25 40 0,clip,width=0.19\linewidth]{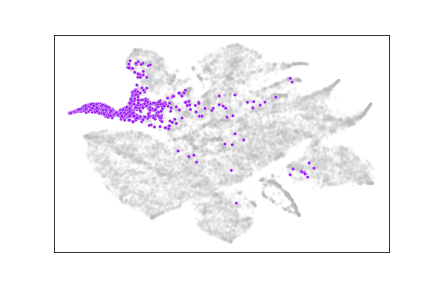}\\ &
        \rotext{\quad \enspace Target}&
        \includegraphics[trim=40 25 40 0,clip,width=0.19\linewidth]{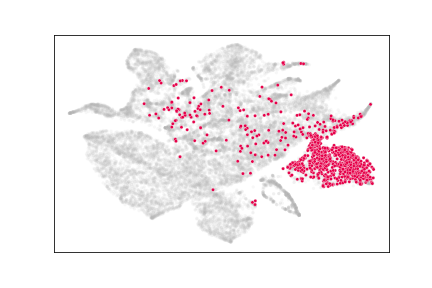}&
        \includegraphics[trim=40 25 40 0,clip,width=0.19\linewidth]{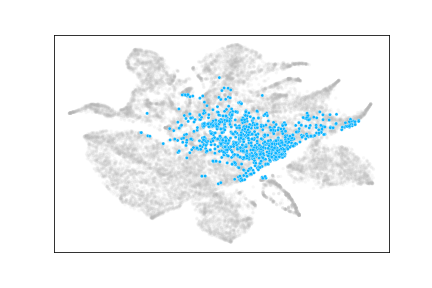}&
        \includegraphics[trim=40 25 40 0,clip,width=0.19\linewidth]{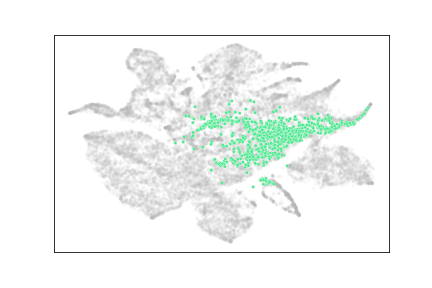}&
        \includegraphics[trim=40 25 40 0,clip,width=0.19\linewidth]{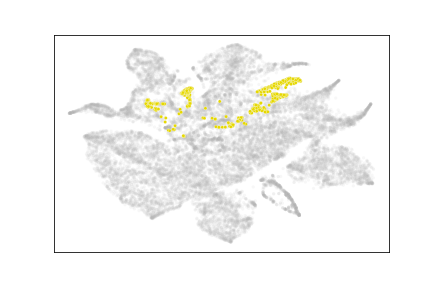}&
       \includegraphics[trim=40 25 40 0,clip,width=0.19\linewidth]{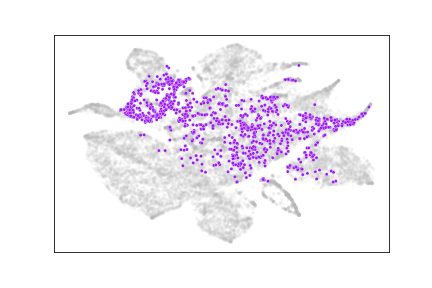}\\
       &  & Car & Bicycle  & Motorcycle   & Truck  & Other-vehicle   \\[-2mm]

     \multirow{2}*{\rotext{ \textbf{\method{}}}} &
     \rotext{\quad \enspace Source}& 
       \includegraphics[trim=40 25 40 0,clip,width=0.19\linewidth]{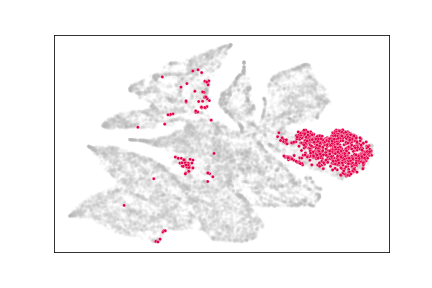}& 
        \includegraphics[trim=40 25 40 0,clip,width=0.19\linewidth]{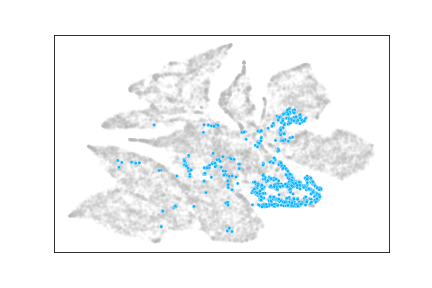}&
        \includegraphics[trim=40 25 40 0,clip,width=0.19\linewidth]{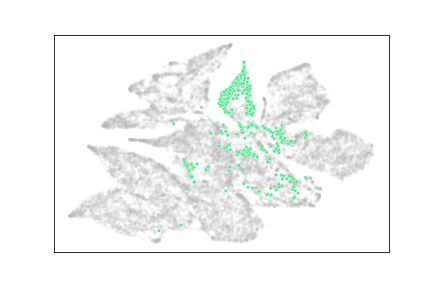}&
        \includegraphics[trim=40 25 40 0,clip,width=0.19\linewidth]{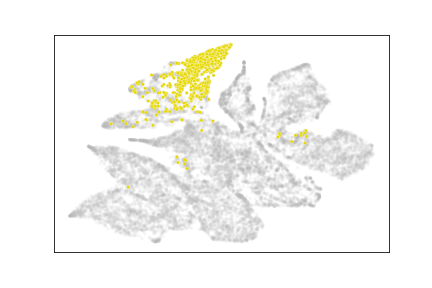}&
        \includegraphics[trim=40 25 40 0,clip,width=0.19\linewidth]{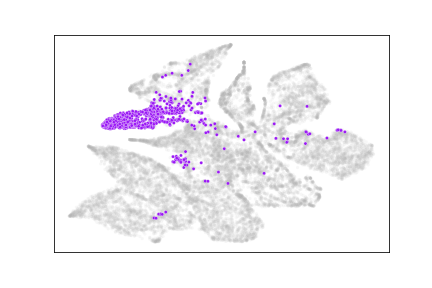}
        \\ & 
        \rotext{\quad \enspace Target}&
        \includegraphics[trim=40 25 40 0,clip,width=0.19\linewidth]{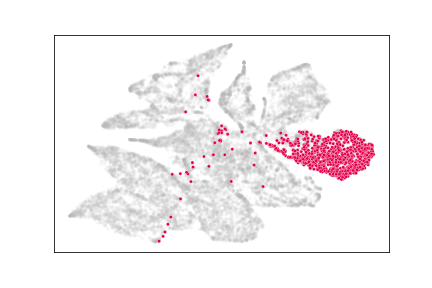}&
        \includegraphics[trim=40 25 40 0,clip,width=0.19\linewidth]{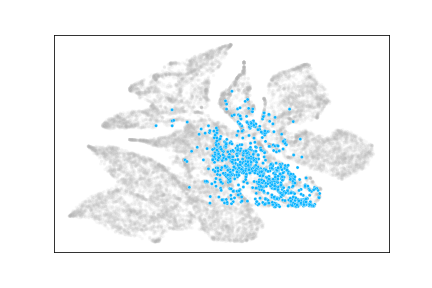}&
        \includegraphics[trim=40 25 40 0,clip,width=0.19\linewidth]{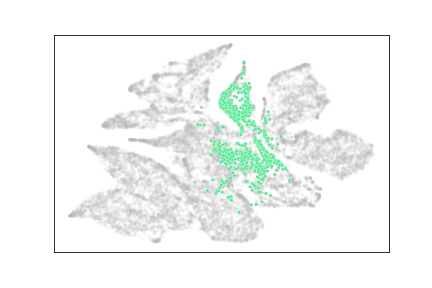}&
        \includegraphics[trim=40 25 40 0,clip,width=0.19\linewidth]{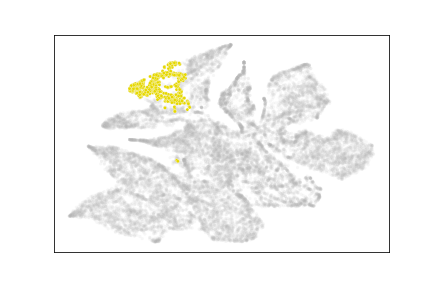}&
       \includegraphics[trim=40 25 40 0,clip,width=0.19\linewidth]{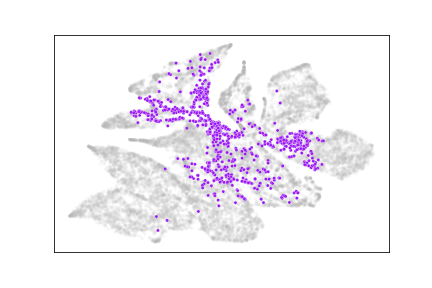}\\ 
        
        \midrule

    \multirow{2}*{\rotext{ \textbf{Source-only}}} &
     \rotext{\quad \enspace Source}& 
       \includegraphics[trim=40 25 40 0,clip,width=0.19\linewidth]{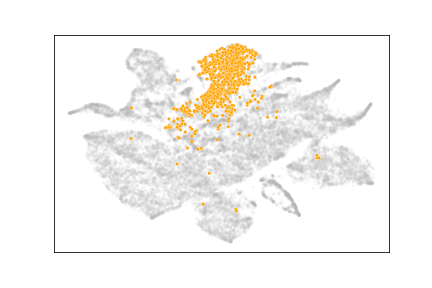}& 
        \includegraphics[trim=40 25 40 0,clip,width=0.19\linewidth]{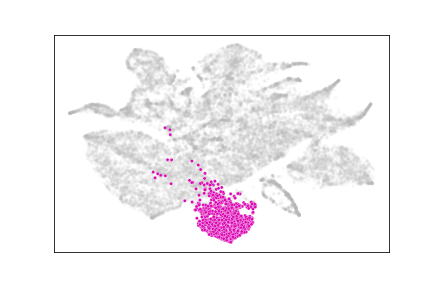}&
        \includegraphics[trim=40 25 40 0,clip,width=0.19\linewidth]{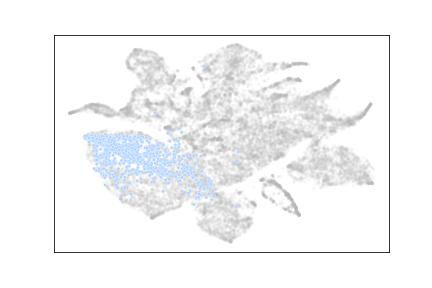}&
        \includegraphics[trim=40 25 40 0,clip,width=0.19\linewidth]{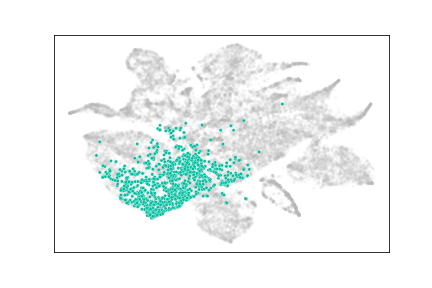}&
        \includegraphics[trim=40 25 40 0,clip,width=0.19\linewidth]{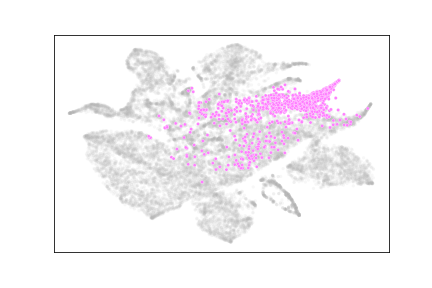}\\ &
        \rotext{\quad \enspace Target}&
        \includegraphics[trim=40 25 40 0,clip,width=0.19\linewidth]{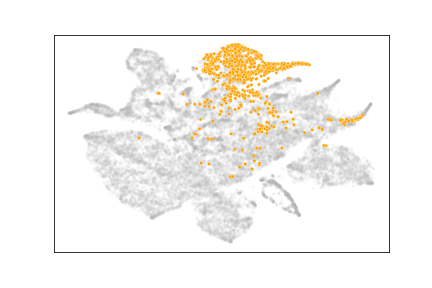}&
        \includegraphics[trim=40 25 40 0,clip,width=0.19\linewidth]{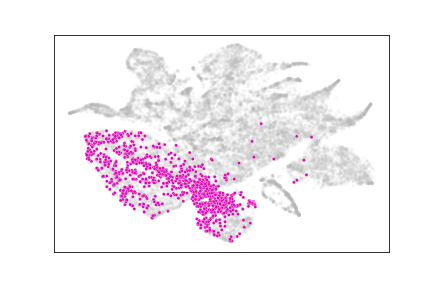}&
        \includegraphics[trim=40 25 40 0,clip,width=0.19\linewidth]{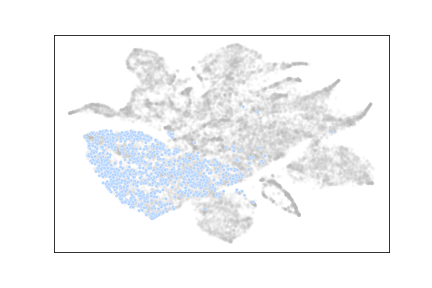}&
        \includegraphics[trim=40 25 40 0,clip,width=0.19\linewidth]{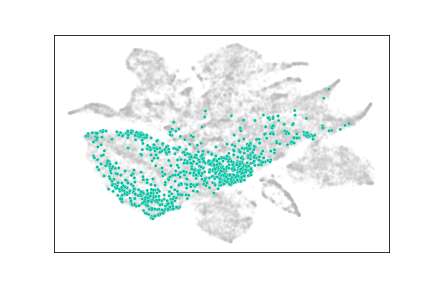}&
       \includegraphics[trim=40 25 40 0,clip,width=0.19\linewidth]{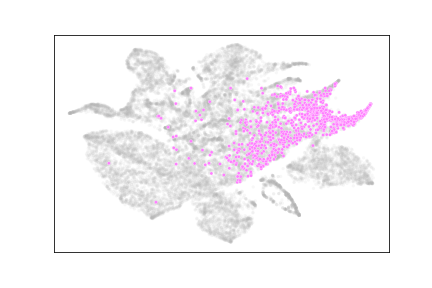}\\
       &  & Pedestrian & Driveable -surface  & Sidewalk   & Terrain  & Vegetation   \\[-2mm]

     \multirow{2}*{\rotext{ \textbf{\method{}}}} &
     \rotext{\quad \enspace Source}& 
       \includegraphics[trim=40 25 40 0,clip,width=0.19\linewidth]{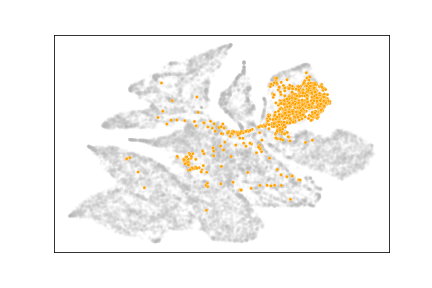}& 
        \includegraphics[trim=40 25 40 0,clip,width=0.19\linewidth]{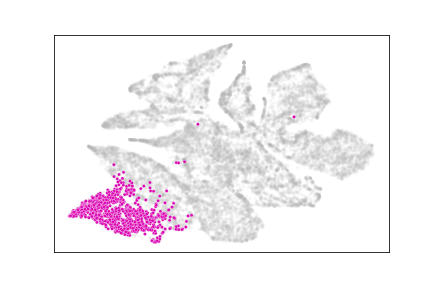}&
        \includegraphics[trim=40 25 40 0,clip,width=0.19\linewidth]{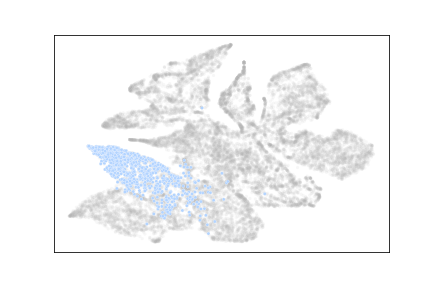}&
        \includegraphics[trim=40 25 40 0,clip,width=0.19\linewidth]{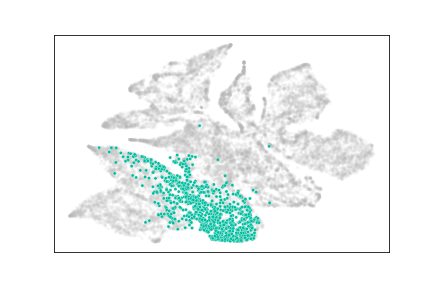}&
        \includegraphics[trim=40 25 40 0,clip,width=0.19\linewidth]{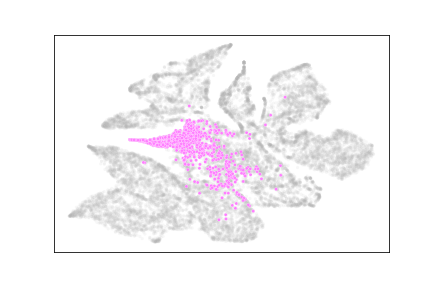}
        \\ & 
        \rotext{\quad \enspace Target}&
        \includegraphics[trim=40 25 40 0,clip,width=0.19\linewidth]{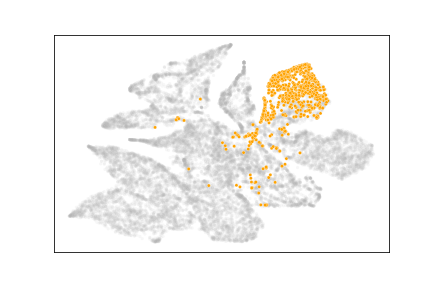}&
        \includegraphics[trim=40 25 40 0,clip,width=0.19\linewidth]{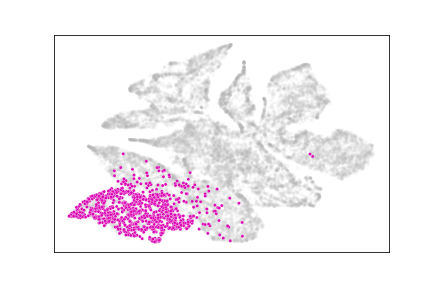}&
        \includegraphics[trim=40 25 40 0,clip,width=0.19\linewidth]{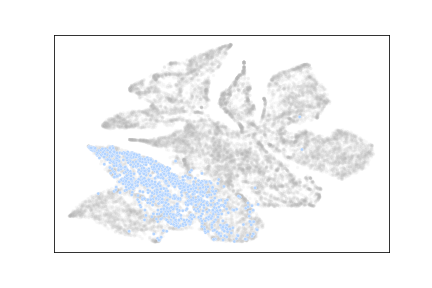}&
        \includegraphics[trim=40 25 40 0,clip,width=0.19\linewidth]{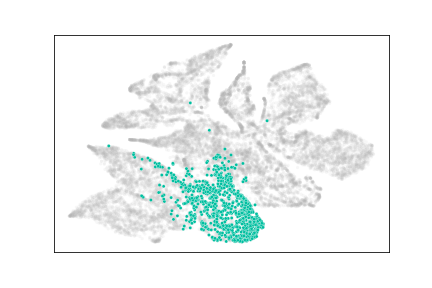}&
       \includegraphics[trim=40 25 40 0,clip,width=0.19\linewidth]{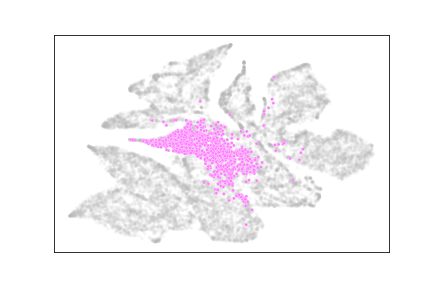}\\ 
    \end{tabular}
    
    \caption{\textbf{\tsne\ visualization of the structure of the source and target latent spaces in the \nstosk{} setting.}}
    \label{fig:tsne_ns_sk_baseline}
    
\end{figure*}

\begin{figure*}
\small
\newcommand{\rotext}[1]{{\begin{turn}{90}{#1}\end{turn}}}
    \setlength{\tabcolsep}{1pt}
    \centering
    \begin{tabular}{@{}ccc@{}c@{}c@{}c@{}c@{}}
    \multirow{2}*{\rotext{ \textbf{Source-only}}} &
     \rotext{\quad \enspace Source}& 
       \includegraphics[trim=40 25 40 0,clip,width=0.19\linewidth]{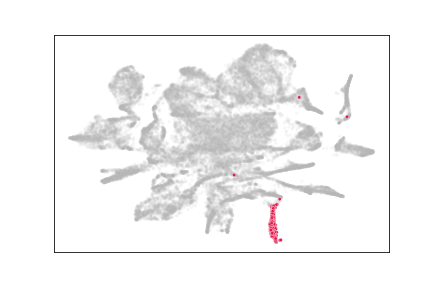}& 
        \includegraphics[trim=40 25 40 0,clip,width=0.19\linewidth]{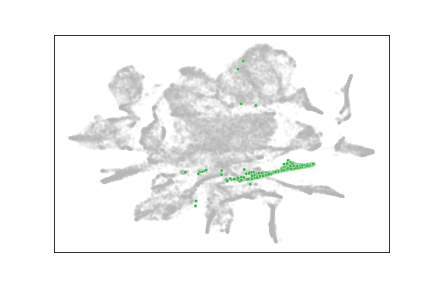}&
        \includegraphics[trim=40 25 40 0,clip,width=0.19\linewidth]{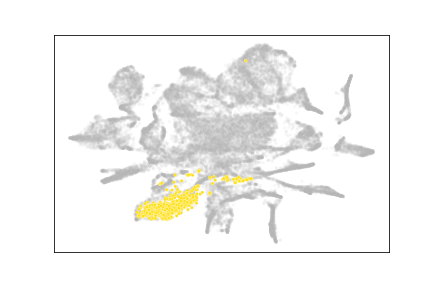}&
        \includegraphics[trim=40 25 40 0,clip,width=0.19\linewidth]{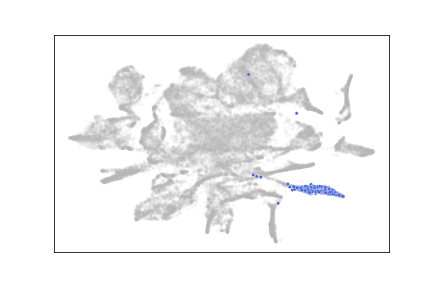}&
        \includegraphics[trim=40 25 40 0,clip,width=0.19\linewidth]{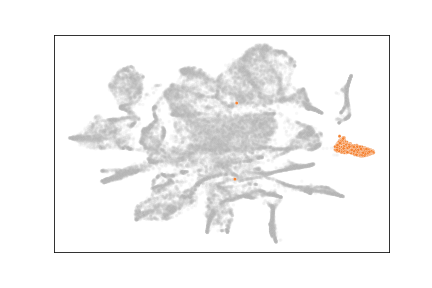}\\ &
        \rotext{\quad \enspace Target}&
        \includegraphics[trim=40 25 40 0,clip,width=0.19\linewidth]{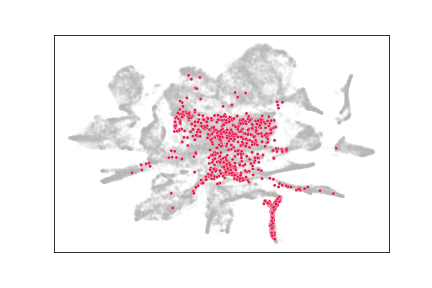}&
        \includegraphics[trim=40 25 40 0,clip,width=0.19\linewidth]{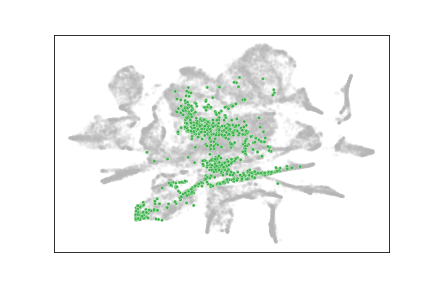}&
        \includegraphics[trim=40 25 40 0,clip,width=0.19\linewidth]{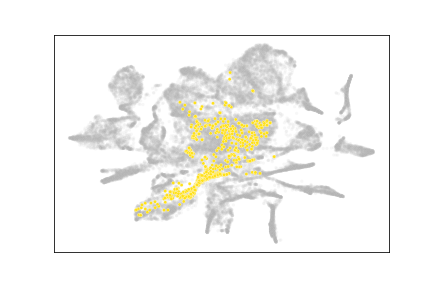}&
        \includegraphics[trim=40 25 40 0,clip,width=0.19\linewidth]{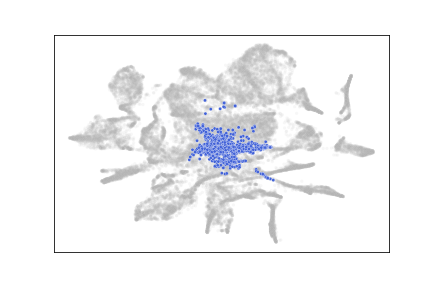}&
       \includegraphics[trim=40 25 40 0,clip,width=0.19\linewidth]{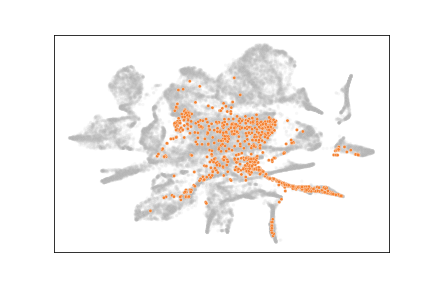}\\

       &  & Car & Bicycle  & Motorcycle   & Truck  & Other-vehicle   \\[-2mm]
      
      \multirow{2}*{\rotext{ \textbf{\method{}}}} &
     \rotext{\quad \enspace Source}& 
       \includegraphics[trim=40 25 40 0,clip,width=0.19\linewidth]{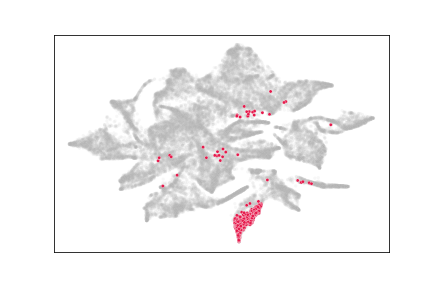}& 
        \includegraphics[trim=40 25 40 0,clip,width=0.19\linewidth]{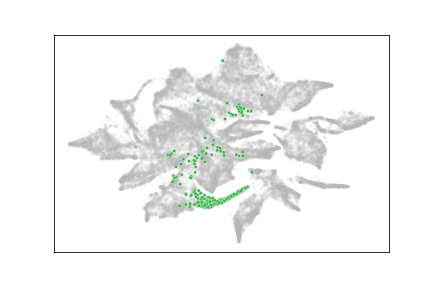}&
        \includegraphics[trim=40 25 40 0,clip,width=0.19\linewidth]{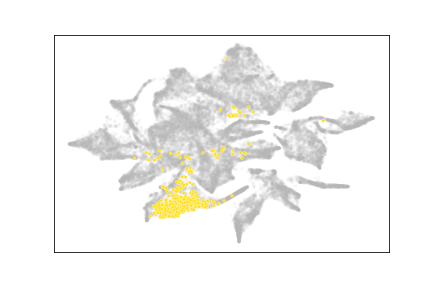}&
        \includegraphics[trim=40 25 40 0,clip,width=0.19\linewidth]{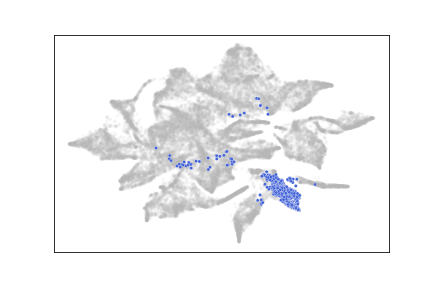}&
        \includegraphics[trim=40 25 40 0,clip,width=0.19\linewidth]{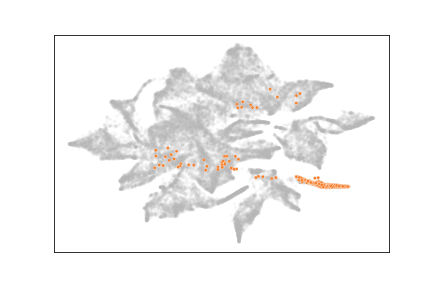}
        \\ & 
        \rotext{\quad \enspace Target}&
        \includegraphics[trim=40 25 40 0,clip,width=0.19\linewidth]{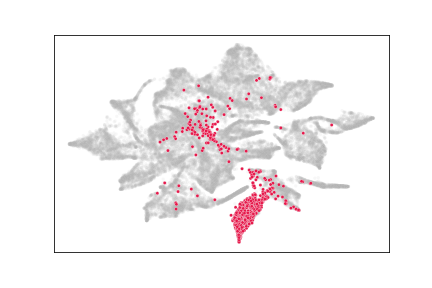}&
        \includegraphics[trim=40 25 40 0,clip,width=0.19\linewidth]{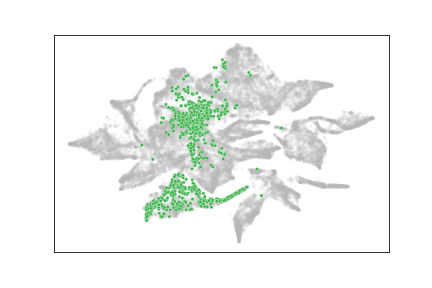}&
        \includegraphics[trim=40 25 40 0,clip,width=0.19\linewidth]{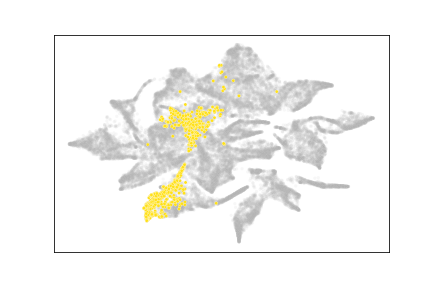}&
        \includegraphics[trim=40 25 40 0,clip,width=0.19\linewidth]{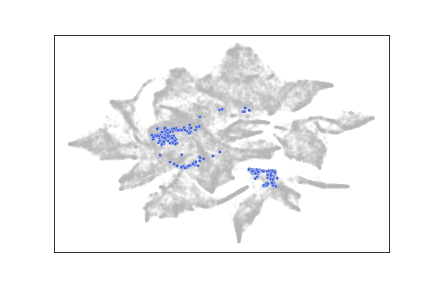}&
       \includegraphics[trim=40 25 40 0,clip,width=0.19\linewidth]{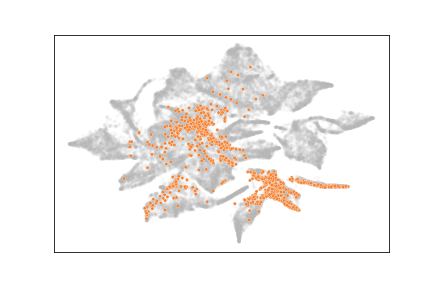}\\ 
        
    \midrule
    
    \multirow{2}*{\rotext{ \textbf{Source-only}}} &
     \rotext{\quad \enspace Source}& 
       \includegraphics[trim=40 25 40 0,clip,width=0.19\linewidth]{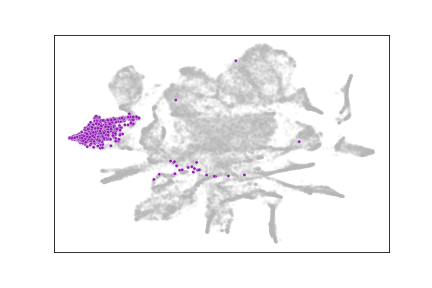}& 
        \includegraphics[trim=40 25 40 0,clip,width=0.19\linewidth]{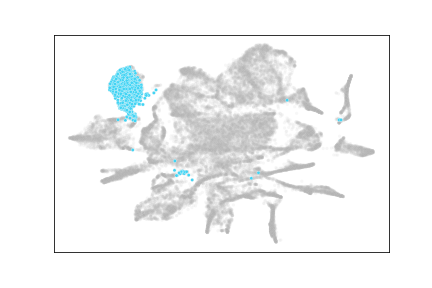}&
        \includegraphics[trim=40 25 40 0,clip,width=0.19\linewidth]{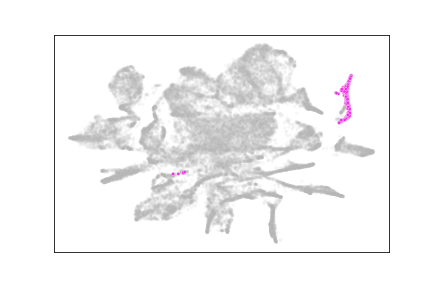}&
        \includegraphics[trim=40 25 40 0,clip,width=0.19\linewidth]{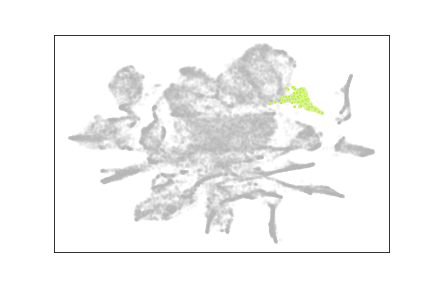}
        &\includegraphics[trim=40 25 40 0,clip,width=0.19\linewidth]{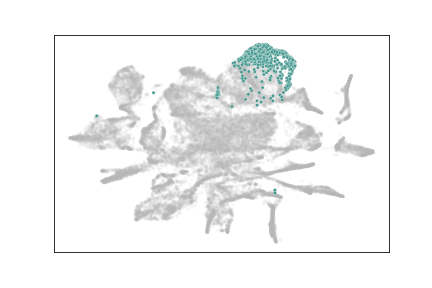}
        \\ &
        \rotext{\quad \enspace Target}&
        \includegraphics[trim=40 25 40 0,clip,width=0.19\linewidth]{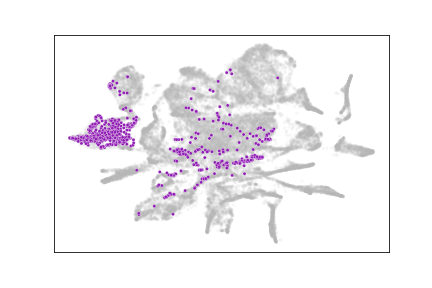}&
        \includegraphics[trim=40 25 40 0,clip,width=0.19\linewidth]{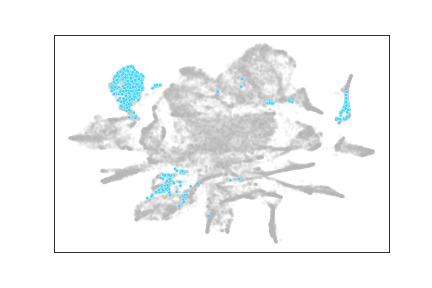}&
        \includegraphics[trim=40 25 40 0,clip,width=0.19\linewidth]{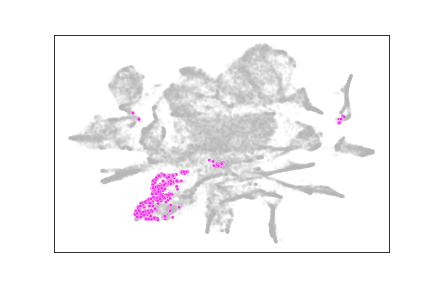}&
        \includegraphics[trim=40 25 40 0,clip,width=0.19\linewidth]{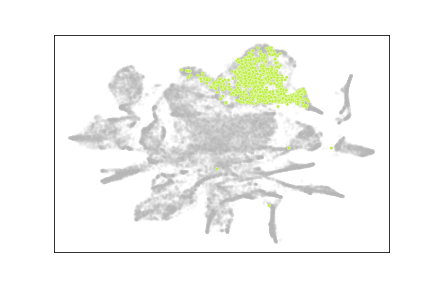}
        &\includegraphics[trim=40 25 40 0,clip,width=0.19\linewidth]{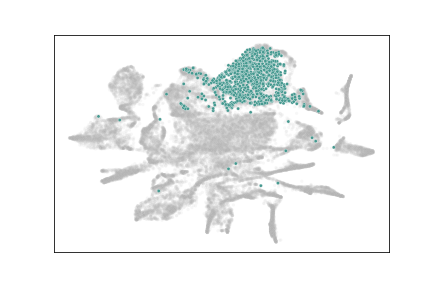}
        \\
       &  & Pedestrian & Bicyclist & Motorcyclist & Road  & Sidewalk  \\[-2mm] 
    
     \multirow{2}*{\rotext{ \textbf{\method{}}}} &
     \rotext{\quad \enspace Source}& 
       \includegraphics[trim=40 25 40 0,clip,width=0.19\linewidth]{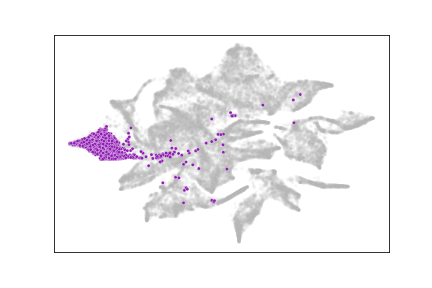}& 
        \includegraphics[trim=40 25 40 0,clip,width=0.19\linewidth]{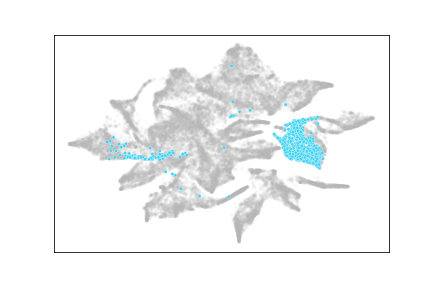}&
        \includegraphics[trim=40 25 40 0,clip,width=0.19\linewidth]{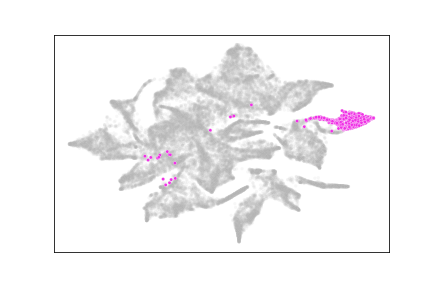}&
        \includegraphics[trim=40 25 40 0,clip,width=0.19\linewidth]{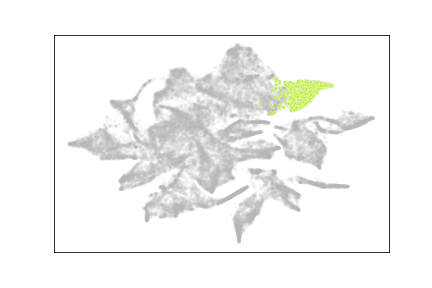}
        &\includegraphics[trim=40 25 40 0,clip,width=0.19\linewidth]{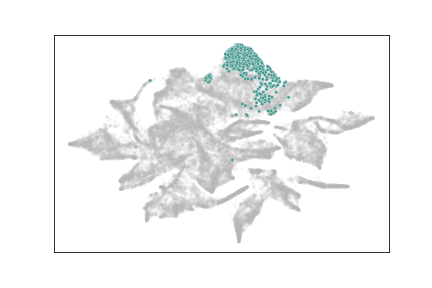}
        \\ & 
        \rotext{\quad \enspace Target}&
        \includegraphics[trim=40 25 40 0,clip,width=0.19\linewidth]{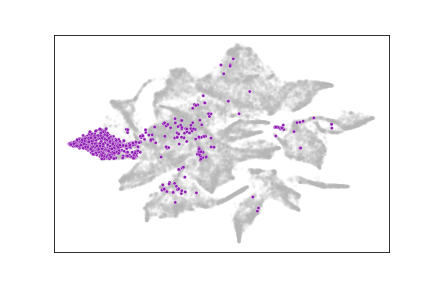}&
        \includegraphics[trim=40 25 40 0,clip,width=0.19\linewidth]{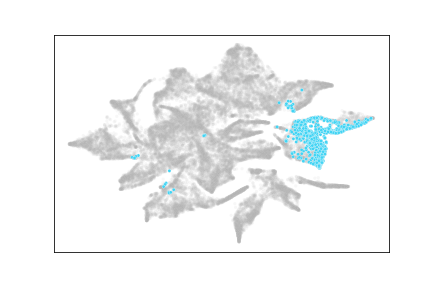}&
        \includegraphics[trim=40 25 40 0,clip,width=0.19\linewidth]{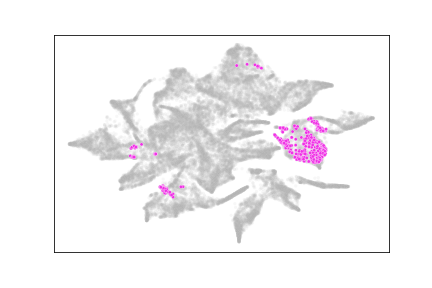}&
        \includegraphics[trim=40 25 40 0,clip,width=0.19\linewidth]{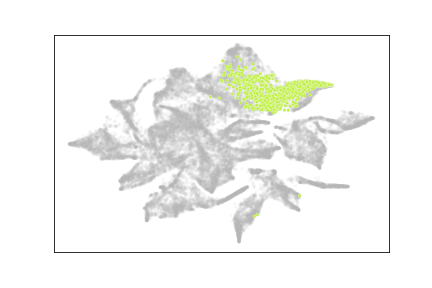}
        &\includegraphics[trim=40 25 40 0,clip,width=0.19\linewidth]{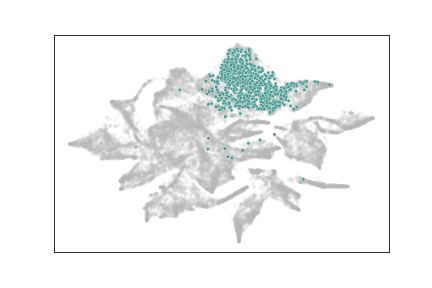}
       \\ 
    \end{tabular}
    
    \caption{\textbf{\tsne\ visualizations of the structure of source and target latent spaces in the \DAsetting{SynL}{\sksyn} setting}  (classes 1 to 10).}
    \label{fig:tsne_syn_sk_baseline1}
    
\end{figure*}

\begin{figure*}
\small
\newcommand{\rotext}[1]{{\begin{turn}{90}{#1}\end{turn}}}
    \setlength{\tabcolsep}{1pt}
    \centering
    \begin{tabular}{ccc@{}c@{}c@{}c@{}c}
    \multirow{2}*{\rotext{ \textbf{Source-only}}} &
     \rotext{\quad \enspace Source}& 
      
        \includegraphics[trim=40 25 40 0,clip,width=0.19\linewidth]{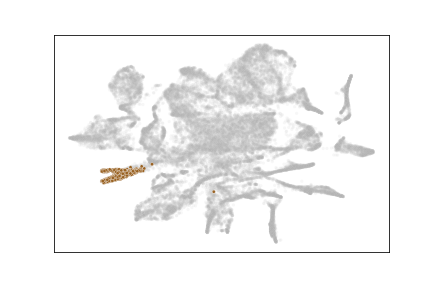}&
        \includegraphics[trim=40 25 40 0,clip,width=0.19\linewidth]{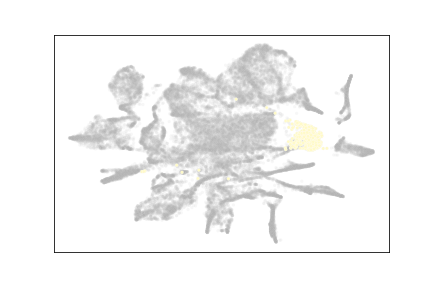}&
        \includegraphics[trim=40 25 40 0,clip,width=0.19\linewidth]{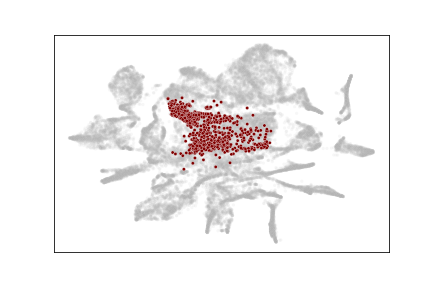}&
        \includegraphics[trim=40 25 40 0,clip,width=0.19\linewidth]{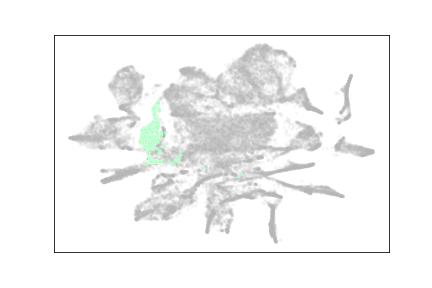}&
        \includegraphics[trim=40 25 40 0,clip,width=0.19\linewidth]{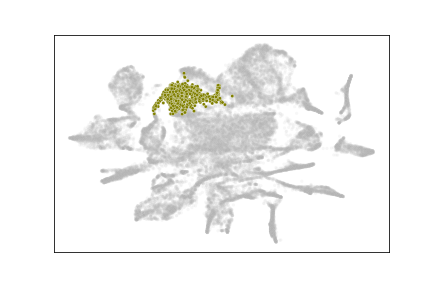}
        \\ &
        \rotext{\quad \enspace Target}&
        
        \includegraphics[trim=40 25 40 0,clip,width=0.19\linewidth]{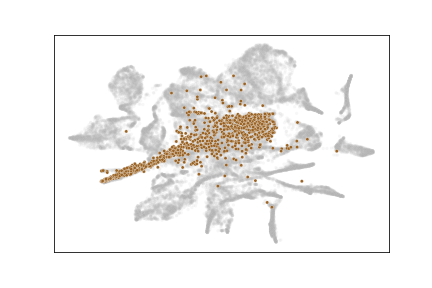}&
        \includegraphics[trim=40 25 40 0,clip,width=0.19\linewidth]{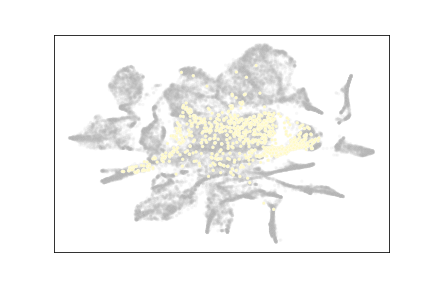}&
       \includegraphics[trim=40 25 40 0,clip,width=0.19\linewidth]{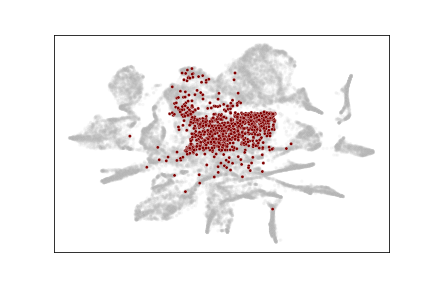}&
       \includegraphics[trim=40 25 40 0,clip,width=0.19\linewidth]{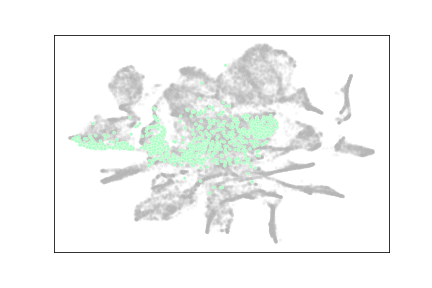}&
       \includegraphics[trim=40 25 40 0,clip,width=0.19\linewidth]{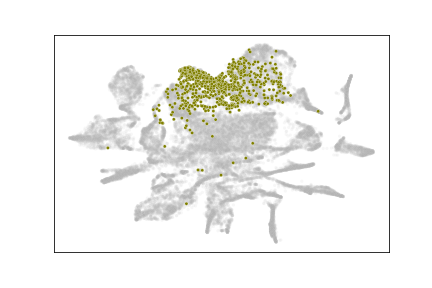}\\
       &  & Building & Fence & Vegetation  & Trunk & Terrain\\[-2mm] 
    
    \multirow{2}*{\rotext{ \textbf{\method{}}}} &
     \rotext{\quad \enspace Source}& 
        \includegraphics[trim=40 25 40 0,clip,width=0.19\linewidth]{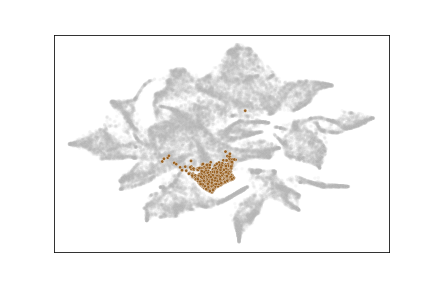}&
        \includegraphics[trim=40 25 40 0,clip,width=0.19\linewidth]{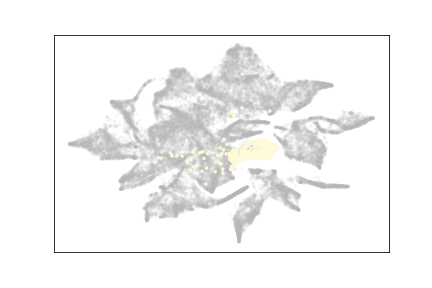}&
        \includegraphics[trim=40 25 40 0,clip,width=0.19\linewidth]{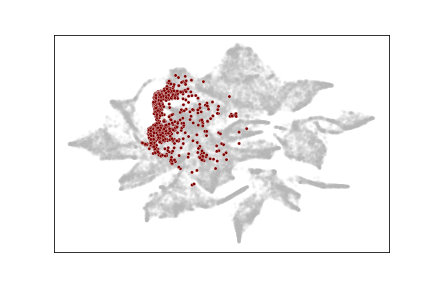}&
        \includegraphics[trim=40 25 40 0,clip,width=0.19\linewidth]{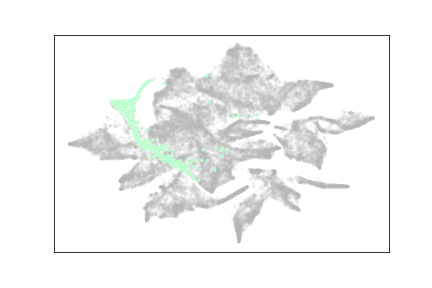}&
        \includegraphics[trim=40 25 40 0,clip,width=0.19\linewidth]{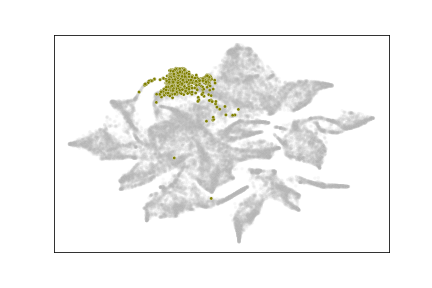}
        \\ & 
        \rotext{\quad \enspace Target}&
        \includegraphics[trim=40 25 40 0,clip,width=0.19\linewidth]{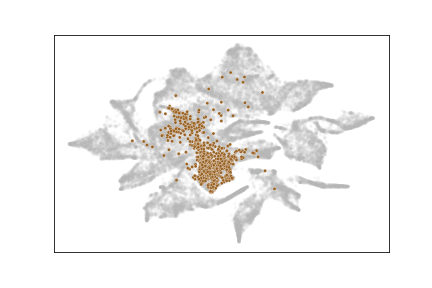}&
        \includegraphics[trim=40 25 40 0,clip,width=0.19\linewidth]{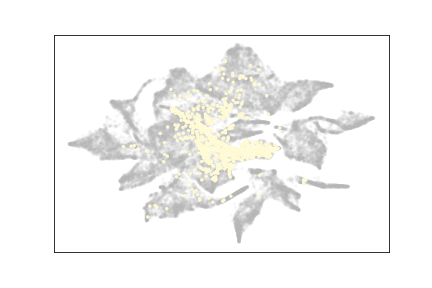}&
       \includegraphics[trim=40 25 40 0,clip,width=0.19\linewidth]{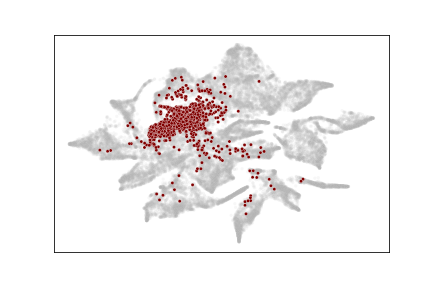}&
       \includegraphics[trim=40 25 40 0,clip,width=0.19\linewidth]{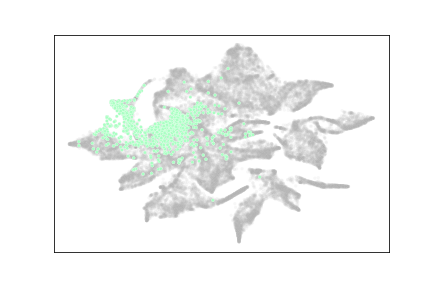}&
       \includegraphics[trim=40 25 40 0,clip,width=0.19\linewidth]{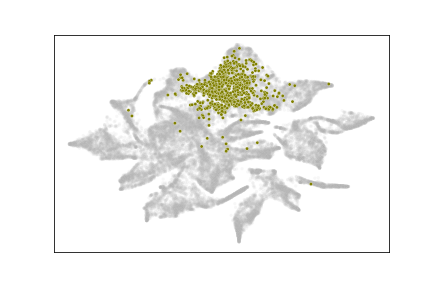}\\ 
    \midrule

    \multirow{2}*{\rotext{ \textbf{Source-only}}} &
     \rotext{\quad \enspace Source}& 
        \includegraphics[trim=40 25 40 0,clip,width=0.19\linewidth]{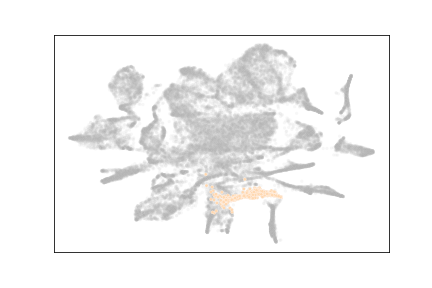}&
        \includegraphics[trim=40 25 40 0,clip,width=0.19\linewidth]{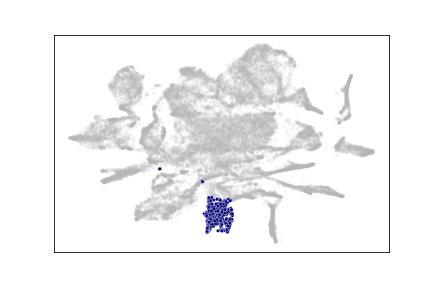}\\ &
        \rotext{\quad \enspace Target}&
        \includegraphics[trim=40 25 40 0,clip,width=0.19\linewidth]{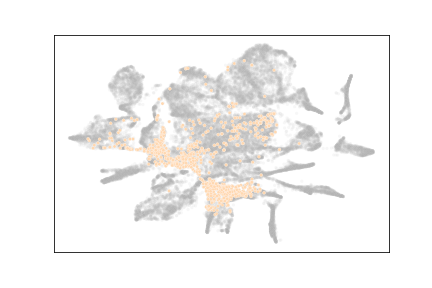}&
        \includegraphics[trim=40 25 40 0,clip,width=0.19\linewidth]{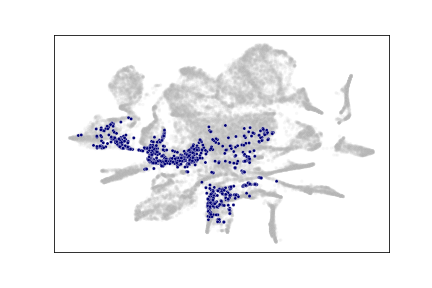}\\
       & &  Pole & Traffic sign  \\[-2mm] 
    
    \multirow{2}*{\rotext{ \textbf{\method{}}}} &
     \rotext{\quad \enspace Source}& 
        \includegraphics[trim=40 25 40 0,clip,width=0.19\linewidth]{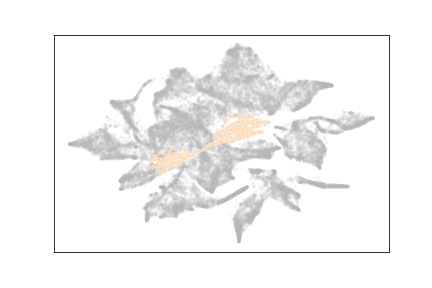}&
        \includegraphics[trim=40 25 40 0,clip,width=0.19\linewidth]{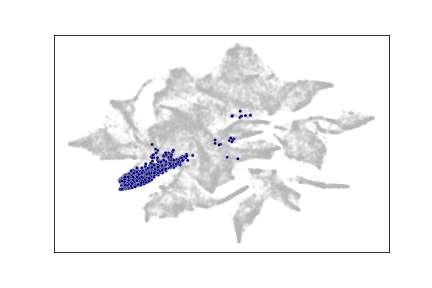}
        \\ & 
        \rotext{\quad \enspace Target}&
        \includegraphics[trim=40 25 40 0,clip,width=0.19\linewidth]{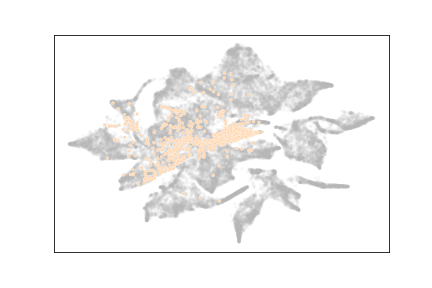}&
        \includegraphics[trim=40 25 40 0,clip,width=0.19\linewidth]{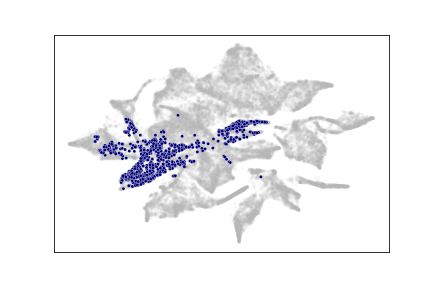}\\ 
    \end{tabular}
    
    \caption{\textbf{\tsne\ visualizations of the structure of source and target latent spaces in the \DAsetting{\synth}{\sksyn} setting}  (classes 11 to 17). Classes "parking" and "other ground" are ignored because they are too rare in the selected scenes to produce any useful visualizations.}
    \label{fig:tsne_syn_sk_baseline2}
    
\end{figure*}

We highlight, in Figure~\ref{fig:tsne_ns_sk_baseline} for \nstosk{} and in Figures~\ref{fig:tsne_syn_sk_baseline1},~\ref{fig:tsne_syn_sk_baseline2} for \DAsetting{\synth}{\sksyn}, the points belonging to different classes in the \tsne\ representations. Overall, we notice that each class is well clustered in these representations, but with slightly bigger clusters in the target latent space, probably because, here as well, the structure of the source latent space is guided by full supervision unlike the target latent space.

\emph{\nstosk{} setting.}
We want to draw the reader's attention on the visualizations corresponding to the classes "car" and "driveable surface". We already know that \method{} performs better than the source-only method on these classes (cf.\ Table~\ref{tab:experiments_per_class_ns_sk}). In Figure~\ref{fig:tsne_ns_sk_baseline}, for these classes, we also notice a better overlap of the corresponding clusters between the source and target domain in \method{}, compared to the case of the source-only method.

\emph{\DAsetting{\synth}{\sksyn}.}
We remark in Figures~\ref{fig:tsne_syn_sk_baseline1},~\ref{fig:tsne_syn_sk_baseline2} that the class clusters in the synthetic source domain have very small sizes. This maybe due to the fact that the synthetic data are less diverse than real data, leading to a less scattered representation in the latent space.


\end{document}